\documentclass[letterpaper]{article} % DO NOT CHANGE THIS
\usepackage{aaai2026}  % DO NOT CHANGE THIS
\usepackage{times}  % DO NOT CHANGE THIS
\usepackage{helvet}  % DO NOT CHANGE THIS
\usepackage{courier}  % DO NOT CHANGE THIS
\usepackage[hyphens]{url}  % DO NOT CHANGE THIS
\usepackage{graphicx} % DO NOT CHANGE THIS
\usepackage{natbib}  % DO NOT CHANGE THIS AND DO NOT ADD ANY OPTIONS TO IT
\usepackage{caption} % DO NOT CHANGE THIS AND DO NOT ADD ANY OPTIONS TO IT
\usepackage{algorithm}
\usepackage{algorithmic}

\usepackage{amsmath,amsfonts,bm}

\def\eqref#1{eq.~(\ref{#1})}
\def\Eqref#1{Eq.~(\ref{#1})}
\def\1{\bm{1}}

\def\rd{{\textnormal{d}}}

\def\vx{{\bm{x}}}

\def\vz{{\bm{z}}}

\DeclareMathAlphabet{\mathsfit}{\encodingdefault}{\sfdefault}{m}{sl}
\SetMathAlphabet{\mathsfit}{bold}{\encodingdefault}{\sfdefault}{bx}{n}

\def\gP{{\mathcal{P}}}

\newcommand{\R}{\mathbb{R}}

\usepackage{amsmath,amsthm,amssymb,stmaryrd}
\newtheorem{theorem}{Theorem}

\usepackage{bm}
\usepackage{array}
\usepackage{comment}
\usepackage{booktabs}
\usepackage{multirow}
\usepackage{arydshln}
\usepackage{subfig}

\newcommand{\PreserveBackslash}[1]{\let\temp=\\#1\let\\=\temp}
\newcolumntype{C}[1]{>{\PreserveBackslash\centering}m{#1}}
\newcolumntype{R}[1]{>{\PreserveBackslash\raggedleft}m{#1}}
\newcolumntype{L}[1]{>{\PreserveBackslash\raggedright}m{#1}}

\newcommand{\std}[1]{\scriptsize{$\pm$#1}}
\newcommand{\avgstd}[1]{\scriptsize{\; (#1)}}

\usepackage{soul}
\let\ul\underline

\usepackage{xcolor}

\usepackage{newfloat}
\usepackage{listings}
\DeclareCaptionStyle{ruled}{labelfont=normalfont,labelsep=colon,strut=off} % DO NOT CHANGE THIS
\floatstyle{ruled}
\newfloat{listing}{tb}{lst}{}
\floatname{listing}{Listing}
\title{FlowPath: Learning Data-Driven Manifolds with Invertible Flows for Robust Irregularly-sampled Time Series Classification}
\author{
    YongKyung Oh\textsuperscript{\rm 1}, 
    Dong-Young Lim\textsuperscript{\rm 2,3}\footnotemark[1], 
    Sungil Kim\textsuperscript{\rm 2,3}\thanks{Corresponding Authors}
}
\affiliations{
    \textsuperscript{\rm 1}Medical \& Imaging Informatics (MII) Group, % Department of Radiological Sciences, 
    University of California, Los Angeles (UCLA), CA, USA \\
    \textsuperscript{\rm 2}Department of Industrial Engineering, Ulsan National Institute of Science and Technology (UNIST), Republic of Korea \\
    \textsuperscript{\rm 3}Artificial Intelligence Graduate School, Ulsan National Institute of Science and Technology (UNIST), Republic of Korea \\
    yongkyungoh@mednet.ucla.edu, \{dlim, sungil.kim\}@unist.ac.kr
}

\begin{document}

\maketitle

\begin{abstract}
    Modeling continuous-time dynamics from sparse and irregularly-sampled time series remains a fundamental challenge. Neural controlled differential equations provide a principled framework for such tasks, yet their performance is highly sensitive to the choice of control path constructed from discrete observations. Existing methods commonly employ fixed interpolation schemes, which impose simplistic geometric assumptions that often misrepresent the underlying data manifold, particularly under high missingness. We propose \textit{FlowPath}, a novel approach that learns the geometry of the control path via an invertible neural flow. Rather than merely connecting observations, FlowPath constructs a continuous and data-adaptive manifold, guided by invertibility constraints that enforce information-preserving and well-behaved transformations. This inductive bias distinguishes FlowPath from prior unconstrained learnable path models. Empirical evaluations on 18 benchmark datasets and a real-world case study demonstrate that FlowPath consistently achieves statistically significant improvements in classification accuracy over baselines using fixed interpolants or non-invertible architectures. These results highlight the importance of modeling not only the dynamics along the path but also the geometry of the path itself, offering a robust and generalizable solution for learning from irregular time series.
\end{abstract}

\begin{links}
    \link{Code}{https://github.com/yongkyung-oh/FlowPath}
    % \link{Datasets}{https://github.com/yongkyung-oh/FlowPath}
    \link{Extended version}{https://arxiv.org/abs/2511.10841}
\end{links}

%-----------------------------------------------------------------------
\section{Introduction}
Many real-world systems evolve continuously over time. In contrast, observations of these systems are typically sparse, irregular, and incomplete \citep{choi_doctor_2016, hossain_method_2020, oh_stable_2024}. Such data are referred to as irregularly-sampled time series (ISTS), characterized by non-uniform sampling intervals and missing observations over time.
While discrete-time models such as GRUs \citep{chung_empirical_2014} have been adapted to handle irregularity using decay mechanisms \citep{che_recurrent_2018}, they remain fundamentally discrete approximations. Neural Differential Equations (NDEs) address this limitation by modeling time as a continuous variable, with Neural Ordinary Differential Equations (Neural ODEs) as a foundational case \citep{chen_neural_2018, lu_beyond_2018}.

For ISTS, Neural Controlled Differential Equations (Neural CDEs) extend this idea by modeling system dynamics as driven by a continuous control path derived from the input \citep{kidger_neural_2020}. Unlike Neural ODEs that evolve from an initial state \citep{rubanova_latent_2019}, Neural CDEs respond directly to the full input stream. However, the control path is typically constructed via fixed interpolation methods, which impose predefined geometric assumptions. This choice has been shown to affect performance substantially and remains a key limitation of the Neural CDE framework \citep{morrill_choice_2022}.

In this paper, we propose \textit{FlowPath}, a framework that models the control path using an invertible neural flow \citep{dinh_density_2017,papamakarios_normalizing_2021}. This replaces the fixed interpolant with a learned, data-adaptive transformation that preserves invertibility. Such flows serve as universal approximators of diffeomorphisms \citep{teshima_coupling-based_2020, ishikawa_universal_2023}, allowing the learned path to be smooth and information-preserving. We formalize the FlowPath architecture, analyze its theoretical properties, and evaluate it on 18 benchmark datasets and real-world datasets. 
Our results demonstrate that learning a structurally-constrained, data-driven manifold leads to statistically significant improvements in classification accuracy over baselines using either fixed interpolants or non-invertible learned paths.

%-----------------------------------------------------------------------
\section{Related Work}\label{sec:related_work}

\paragraph{RNN-style Discretizations.} 
Traditional Recurrent Neural Networks (RNNs) assume fixed step sizes \citep{goodfellow_deep_2016}, and adaptations such as GRU-D \citep{che_recurrent_2018} incorporate missingness indicators or time-lag decay to handle irregular sampling, but remain discrete-time approximations that can still struggle with large sampling gaps or nonuniform intervals.
Thus, a conventional RNN or GRU-D processes data at sampled time points \(\{t_k\}\). An RNN state \(\bm{h}_k\in\mathbb{R}^{d_h}\) typically evolves via:
\begin{equation*}
\bm{h}_{k+1}
=
\Psi\Bigl(\bm{h}_k,\;\bm{x}_{k+1},\;\Delta t_k\Bigr),
\end{equation*}
where \(\bm{x}_{k+1}\) is the observation at time \(t_{k+1}\) and \(\Delta t_k=t_{k+1}-t_k\). The function \(\Psi\) may incorporate gating or decay terms to handle missingness and so remains a discrete approximation of an underlying continuous process.

\paragraph{NDE-based Models.}
To overcome the limitations of purely discrete-time methods, \citet{chen_neural_2018} introduced Neural ODEs, casting hidden state evolution as a continuous function of time, 
\begin{equation*}
\frac{\mathrm{d}\bm{z}(t)}{\mathrm{d}t}
=
f\bigl(t,\bm{z}(t)\bigr),\quad
\bm{z}(0)=h\bigl(\bm{x}(0)\bigr).
\end{equation*}
The state \(\bm{z}(t)\in\mathbb{R}^{d_z}\) in a Neural ODE is obtained by integrating an initial value \(\bm{z}(0)\) from \(0\) to \(T\). While this continuous viewpoint elegantly handles irregular time points, its strict reliance on the initial value problem makes it difficult to incorporate new observations arriving at later times without restarting or extending the integration \citep{rubanova_latent_2019,kidger_neural_2022}.

Neural CDEs~\citep{kidger_neural_2020} overcome this limitation by representing the input time series as a control path \(X(t)\in \mathbb{R}^{d_x}\), thereby enabling the hidden state to evolve continuously in response to incoming observations. The hidden state dynamics are governed by a Riemann--Stieltjes integral with respect to the control path:
\begin{equation*}
\bm{z}(t)
=
\bm{z}(0)
+
\int_{0}^{t}
f\bigl(\tau,\bm{z}(\tau)\bigr)\,\mathrm{d}X(\tau),
\quad
\bm{z}(0)=h\bigl(X(0)\bigr).
\end{equation*}
Control path $X(t)$ is typically constructed from observed data using fixed interpolation methods such as linear or cubic splines \citep{morrill_neural_2021}. This introduces a structural prior that may not align with the true dynamics. Empirical results show that interpolation choice is a sensitive hyperparameter with significant impact on performance, and no single method is optimal across tasks \citep{morrill_choice_2022}.

Recent studies replace the fixed interpolant with a learned path from unconstrained networks, such as encoder-decoders \citep{jhin_exit_2022, jhin_learnable_2023}. Others apply attention mechanisms within the Neural CDE framework to reweight temporal contributions \citep{jhin_attentive_2024}. Further work focuses on improving the expressivity of the dynamics by signature transformation \citep{morrill_neural_2021}. These approaches enhance flexibility but may introduce instability when the learned path lacks structural constraints.

\paragraph{Neural Flows.}
\citet{bilos_neural_2021} propose modeling ODE solutions with an invertible transformation, circumventing numerical solvers by directly parameterizing a diffeomorphism in time. Such invertible mappings have been extensively studied in normalizing flows \citep{papamakarios_normalizing_2021}, where tractable Jacobians enable density estimation or generative sampling. 
Formally, Neural Flow \citep{bilos_neural_2021} is designed to learn an invertible mapping \(F(t,\bm{x}(0);\theta_{F})\) that directly parameterizes the solution of an ODE or dynamical process. 
Given an implicit dynamical system governed by,
\begin{equation*}
\frac{\mathrm{d}\bm{z}}{\mathrm{d}t}
=
f\bigl(\bm{z}(t)\bigr),
\quad
\bm{z}(0)=\bm{x}(0),
\end{equation*}
Neural Flow posits a neural function \(F\) satisfying, 
\begin{equation*}
\tfrac{\mathrm{d}}{\mathrm{d}t}\,F(t,\bm{x}(0))=f\bigl(F(t,\bm{x}(0))\bigr),
\quad
F(0,\bm{x}(0))=\bm{x}(0).
\end{equation*}
Crucially, \(F(\cdot)\) is designed to be a diffeomorphism in \(\bm{x}\), ensuring that mapping local volumes is always invertible. This property allows Neural Flow to avoid the need for standard numerical ODE solvers and to guarantee the model’s Jacobian is tractable. 
While originally aimed at continuous normalizing flows for generative tasks, the concept of applying invertible flows to dynamic time series modeling has grown \citep{bilos_neural_2021,oh_dualdynamics_2025}. 

Our approach builds on the observation that Neural CDEs and Neural Flows address complementary aspects of time series modeling. Instead of using a Neural Flow as an end-to-end model, we use it to construct a data-driven control path. This design combines the continuous-time formulation of NDEs with a learnable, invertible transformation that adapts to the observed data, avoiding the limitations of fixed interpolation schemes or unconstrained trajectories.

%-----------------------------------------------------------------------
\section{Methodology}\label{sec:methodology}
We are given a dataset \(\mathcal{D} = \{(\bm{X}_i, y_i)\}_{i=1}^{N}\), where each instance \(\bm{X}_i\) comprises a time-indexed sequence \(\bigl(t_{i}, \bm{x}_{i}\bigr)\). Each \(t_i \subset \mathbb{T}\) represents the set of sampling times, and \(\bm{x}_i \in \mathbb{R}^{d_x}\) is the multivariate observation at those times. We aim to classify each instance with a label \(y_i \in \{1,2,\dots,K\}\). The sampling intervals \(\Delta t_j = t_{j+1} - t_j\) are not necessarily uniform, and certain intervals may be missing data entirely. 
We aim to learn a predictive function $\mathcal{H} : (t_i, \bm{x}_i) \mapsto y_i$ that robustly classifies sequences under irregular sampling.

\subsection{FlowPath Framework}\label{sec:architecture}
Raw observation \(\vx(t)\) includes how missing values create discontinuities and distortions in the time series, disrupting its temporal structure. The proposed learnable flow \(\Phi(t)\) generates a continuous path that preserves the underlying temporal dynamics. Unlike interpolation, which applies a fixed approach, the learnable flow adapts to surrounding patterns, leading to more informative representations for the downstream task.

\paragraph{Learning the Underlying Manifold.}
Let \(F\colon [0,T]\times \mathbb{R}^{d_x}\to \mathbb{R}^{d_x}\) be a neural network parameterized by \(\theta_{F}\) such that the mapping \(F\) is a diffeomorphism in its second argument \citep{bilos_neural_2021}. 
We explicitly define the learnable control path \(\Phi : [0, T] \to \mathbb{R}^{d_x}\) as:
\begin{equation}\label{eq:Flow}
\Phi(t)
=
F\bigl(t,\,\bm{x};\,\theta_{F}\bigr),
\end{equation}
where \(\bm{x}(0) \in \mathbb{R}^{d_x}\) denotes the earliest available observation in the irregular time series. 
Unlike fixed spline-based interpolations, \(\Phi(t)\) is parameterized by a neural network \(F\) with learnable parameters \(\theta_F\), 
allowing the control path to be learned directly from data in a flexible and adaptive manner.

\paragraph{Dynamics on the Learned Manifold.}
We define a hidden state \(\bm{z}(t)\in\mathbb{R}^{d_z}\) governed by the Riemann--Stieltjes integral as suggested by \citet{kidger_neural_2020}:
\begin{equation}\label{eq:FlowPath}
\bm{z}(t)
=
\bm{z}(0)
+\int_{0}^{t}
f\bigl(\tau,\bm{z}(\tau);\theta_{f}\bigr)
\,\mathrm{d}\Phi(\tau). 
\end{equation}
Then, an equivalent ODE form of \Eqref{eq:FlowPath} defines:
\begin{align}\label{eq:FlowPath_ODE}
\dot{\bm{z}}(t)=
f\bigl(t,\bm{z}(t);\theta_f\bigr)\,\dot{\Phi}(t), 
\end{align}
with \(\bm{z}(0)=h\bigl(\bm{x}(0)\bigr)\), and \(h\colon\mathbb{R}^{d_x}\to\mathbb{R}^{d_z}\) an embedding layer. In \Eqref{eq:FlowPath_ODE}, the dot notation $\dot{\bm{z}}(t)$ and $\dot{\Phi}(t)$ denote the time derivatives, i.e., $\frac{d\bm{z}(t)}{dt}$ and $\frac{d\Phi(t)}{dt}$, respectively.
The smoothness assumptions on \(\Phi\) guarantee Lipschitz continuity in time, while invertibility enforces distribution-preserving transformations \citep{papamakarios_normalizing_2021,bilos_neural_2021}.
Since \(\Phi\) is a diffeomorphism, the learned path \(t\;\mapsto\;\Phi(t)\) admits well-defined inverses and Jacobians, which strengthen theoretical guarantees such as existence, uniqueness, and universal approximation for continuous-time processes. 
In practice, numerical solvers treat \(\dot{\Phi}(t)\) as a known function (output of a neural net) and integrate \(\dot{\bm{z}}(t)\) 
forward in time, following the ODE form in \Eqref{eq:FlowPath_ODE}.

\paragraph{Considerations for the Invertible Flow.}
We parameterize the control path using an invertible neural network based on normalizing flows, which provide a flexible framework for learning smooth, bijective transformations \citep{papamakarios_normalizing_2021}. 
Recent theoretical results show that such architectures can approximate any diffeomorphism under suitable conditions \citep{teshima_coupling-based_2020, ishikawa_universal_2023}. This ensures that the learned path $\Phi(t)$ remains invertible and smooth by design, reducing the risk of instability often seen in unconstrained path parameterizations.

Within the FlowPath framework, \(\Phi(t) = F(t, \bm{x}(0); \theta_F)\) acts as a trainable control path that modulates the evolution of the latent state \(\bm{z}(t)\). The function $F$ is designed as an invertible neural network, drawing inspiration from normalizing flows \citep{papamakarios_normalizing_2021} and specifically neural flows that parameterize diffeomorphisms \citep{bilos_neural_2021}. This design ensures that $\Phi(t)$ is a smooth transformation of its inputs, providing FlowPath with enhanced flexibility to model complex temporal dependencies.

\subsection{Properties of FlowPath}\label{sec:property}
The design of FlowPath results in a time-continuous model that (i) preserves information through an invertible flow, (ii) ensures well-posedness of the underlying dynamics, and (iii) promotes strong generalization capability.
We present these three properties below, detailing full explanations in the supplementary material. 

FlowPath uses an invertible flow \(\Phi\) that reshapes the latent distribution without collapsing or tearing probability mass. The following theorem states that the instantaneous change in log-density is exactly driven by the divergence of the controlled dynamics, ensuring no unintended compression or expansion of probability mass.

\begin{theorem}[Preservation of Probability Density]\label{thm:preservation}
Under the FlowPath framework in \Eqref{eq:FlowPath}, let \(\Phi(t)\) be a $C^1$-diffeomorphism, and \(f\) be Lipschitz continuous in \(\vz\). Then the probability density $p(\vz(t))$ of the latent state evolves according to 
\begin{equation}\label{eq:preservation}
\frac{\mathrm{d}}{\mathrm{d}t}\,\log p\bigl(\vz(t)\bigr)
\;=\;
-\,
\mathrm{div}_{\vz}
\Bigl(
  f\bigl(t,\vz(t)\bigr)\,\dot{\Phi}(t)
\Bigr).
\end{equation}
Consequently, \(\Phi\) neither collapses nor arbitrarily expands probability mass, preserving the geometry of the latent distribution.
\end{theorem}

This property, while crucial for continuous normalizing flows in generative modeling~\citep{papamakarios_normalizing_2021}, also benefits our classification task by preserving class‐separating structures in latent space, thereby improving robustness to irregular sampling and missing data.

Next, Theorem~\ref{thm:exist_unique} establishes the standard existence and uniqueness guarantee for the FlowPath dynamics. 
\begin{theorem}[Existence and Uniqueness of Solutions]\label{thm:exist_unique}
Fix $\theta_f$. Let \(f(t,\cdot;\theta_f)\) in \Eqref{eq:FlowPath} be continuous in $t$ and Lipschitz in \(\vz\), and \(\Phi(t)\) be continuously differentiable on \([0,T]\). Then  for any initial condition \(\vz(0)\in\mathbb{R}^{d_z}\), there exists a unique continuous solution 
\begin{equation*}
\bm{z}(t)
=
\bm{z}(0)
+\int_{0}^{t}
f\bigl(s,\bm{z}(s);\theta_{f}\bigr)
\,\mathrm{d}\Phi(s). 
\end{equation*}
\end{theorem}

The universal approximation properties of Neural ODEs and Neural Flows have been extensively studied and established in prior work~\citep{lin_resnet_2018,li_deep_2022,teshima_coupling-based_2020,ishikawa_universal_2023}.  Building upon these results, one can check that FlowPath is a universal approximator under appropriate regularity conditions. Under the assumption that \(\Phi\) is a \(C^1\)-diffeomorphism and \(f\) is Lipschitz, Neural ODEs are universal approximators for any smooth target trajectory $u$ in the uniform norm. Thus, one can choose \(f\) such that within any desired precision $\epsilon'$
\[
  \sup_{t\in[0,T]}\bigl\|\,f(t,z(t)) - \dot\Phi(t)^{-1}\dot u(t)\bigr\| < \epsilon'.
\]
Substituting this into the FlowPath dynamics \(\dot z(t)=f(t,z(t))\,\dot\Phi(t)\) and integrating yields
\begin{align*}
\sup_{t\in[0,T]}\|z(t)-u(t)\|
&\le \int_{0}^{T}\|f(\tau,z(\tau))\,\dot\Phi(\tau)-\dot u(\tau)\|\,\mathrm{d}\tau \\
&\le \int_{0}^{T}\Bigl(\sup_{s\in[0,T]}\|\dot\Phi(s)\|\Bigr)\,\epsilon'\,\mathrm{d}\tau \\
&= \epsilon'\;\Bigl(\sup_{s\in[0,T]}\|\dot\Phi(s)\|\Bigr)\,T.
\end{align*}
Since \(\sup_{t\in[0,T]}\|\dot\Phi(t)\|\) and \(T\) are finite, by choosing \(\epsilon'\) appropriately, FlowPath can approximate any smooth trajectory to within arbitrary tolerance \(\epsilon>0\).

Let \(\mathcal{L}\) be an \(\ell\)-Lipschitz loss in the model’s final output.  We parameterize FlowPath by \(\theta=(\theta_f,\theta_F)\), where \(\theta_f\) specifies the vector field \(f\) and \(\theta_F\) defines the invertible flow \(\Phi\).

\begin{theorem}[Generalization Bound]\label{thm:generalization}
Assume that the flow map \(\Phi(\cdot;\theta_{F})\colon [0,T]\to\mathbb{R}^{d_x}\) is a diffeomorphism with 
\(\sup_{t\in[0,T]}\|\dot{\Phi}(t)\|\le M_{\Phi}\),
and the vector field \(f(\cdot,\cdot;\theta_f)\) is \(L_f\)-Lipschitz in the hidden state.  Let
\(\mathcal{F}_\Theta\) be the class of all such FlowPath predictors.  Given \(n\) i.i.d.\ samples \(\{(X_i,y_i)\}_{i=1}^n\) drawn from $\gP$, define the population risk  by
\[
  \mathcal{R}_{\mathrm{true}}(f)
  = \mathbb{E}_{(X,y)\sim\mathcal{P}}\bigl[\mathcal{L}\bigl(f(X),y\bigr)\bigr],
\]
and the empirical risk by
\[
  \mathcal{R}_{\mathrm{emp}}(f)
  = \frac{1}{n}\sum_{i=1}^n\mathcal{L}\bigl(f(X_i),y_i\bigr).
\]
If \(\widehat f=\arg\min_{f\in\mathcal{F}_\Theta}\mathcal{R}_{\mathrm{emp}}(f)\), then for any \(\delta>0\), with probability at least \(1-\delta\),
\[
\bigl|\mathcal{R}_{\mathrm{true}}(\widehat f)
-\mathcal{R}_{\mathrm{emp}}(\widehat f)\bigr|
\;\le\;
\alpha\,\frac{1}{\sqrt{n}}
\;+\;
\sqrt{\frac{\ln(1/\delta)}{2n}},
\]
where \(\alpha\) depends on \(\ell\), \(L_f\), and \(M_{\Phi}\).
\end{theorem}

This result, derived via Rademacher complexity arguments~\citep{bartlett_spectrally-normalized_2017,chen_generalization_2020}, indicates that, as the number of training samples \(n\) grows, the empirical loss of a FlowPath model closely approximates its true expected loss on unseen data.

%-----------------------------------------------------------------------
\section{Experiments}\label{sec:experiments}

We design experiments to evaluate both the classification accuracy and robustness of our model under varying degrees of missing observations. Specifically, we consider 18 diverse time series classification datasets from UEA \& UCR Data Repository, referenced in \citep{bagnall_uea_2018,h_a_dau_ucr_2019}, utilizing the \texttt{sktime} Python library \citep{loning_sktime_2019}. 
We followed the experimental setup and dataset selection described by \citet{oh_dualdynamics_2025}. 
The original paper considered three categories: `Motion \& Human Activity Recognition (HAR)', `Electrocardiogram (ECG) \& Electroencephalogram (EEG)', and `Sensor' domains, as summarized in the supplementary material. 

In the experiments, three derivative datasets were created from each original dataset by artificially inducing missing data at rates of 30\%, 50\%, and 70\%. Consequently, for each original dataset, there were four distinct settings, including the original data. 
The datasets were split into training, validation, and testing subsets, adhering to a 70/15/15 proportion. 
Lastly, the classification metrics were computed for each setting and averaged with five iterations.

\subsection{Benchmark Methods}
We utilized the conventional \textit{RNN}~\citep{medsker_recurrent_1999}, \textit{LSTM}~\citep{s_hochreiter_long_1997}, and \textit{GRU}~\citep{chung_empirical_2014}, with mean imputation. 
Furthermore, we employed several modifications of the GRU, specifically \textit{GRU-$\Delta t$}~\citep{choi_doctor_2016}, \textit{GRU-D}~\citep{che_recurrent_2018}, which have been tailored to efficiently handle ISTS. 
Additionally, we considered models that are based on the principles of Neural ODEs, such as \textit{GRU-ODE}~\citep{brouwer_gru-ode-bayes_2019}, \textit{ODE-RNN}~\citep{rubanova_latent_2019} and \textit{ODE-LSTM}~\citep{lechner_learning_2020}. 
Moreover, recent advancements of Neural CDEs were also included in the evaluation, such as \textit{Neural CDE}~\citep{kidger_neural_2020}, \textit{Neural RDE}~\citep{morrill_neural_2021}, \textit{ANCDE}~\citep{jhin_attentive_2024}, \textit{EXIT}~\citep{jhin_exit_2022}, \textit{LEAP}~\citep{jhin_learnable_2023}, and \textit{DualDynamics}~\citep{oh_dualdynamics_2025}.
Lastly, we included na\"ive \textit{Neural Flow}~\citep{bilos_neural_2021} with different configurations: ResNet, GRU, and Coupling Flow.

\subsection{Qualitative Evaluation of Key Properties}
We used the `BasicMotions` dataset for a qualitative analysis. The dataset comprises motion data collected from participants performing four distinct activities while wearing a smartwatch.
Each activity was performed five times, with data sampled every 0.1 seconds over a 10-second period. 

\begin{figure}[htbp]
\centering\captionsetup{skip=5pt}
\captionsetup[subfigure]{justification=centering, skip=5pt}
    \subfloat{ 
        \includegraphics[width=\linewidth]{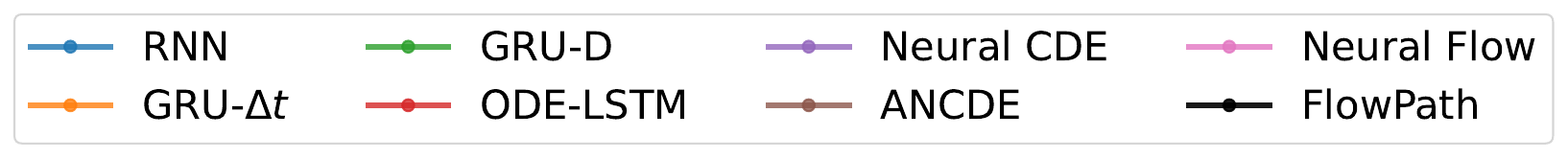}} \\[-10pt]
    \subfloat[Regular]{
      \includegraphics[width=0.48\linewidth]{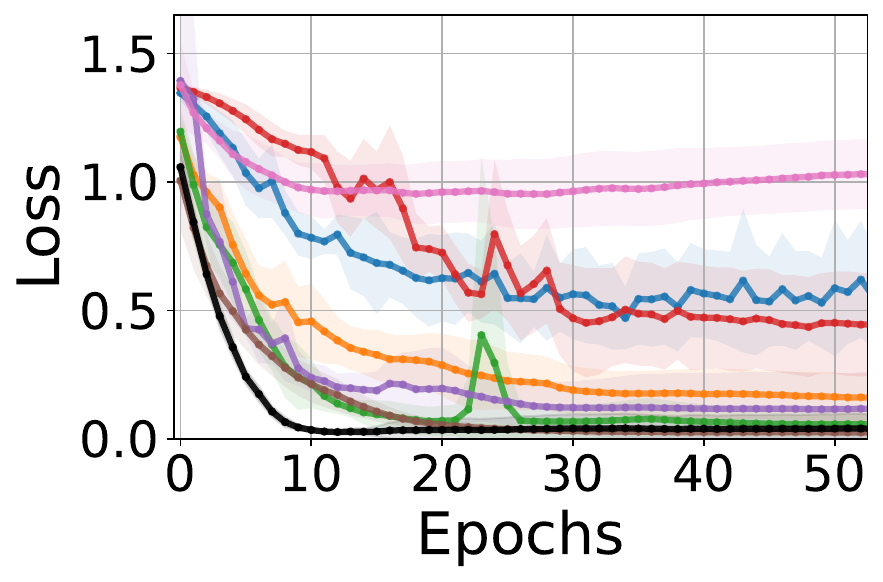}} \hfil
    \subfloat[50\% Missing]{
      \includegraphics[width=0.48\linewidth]{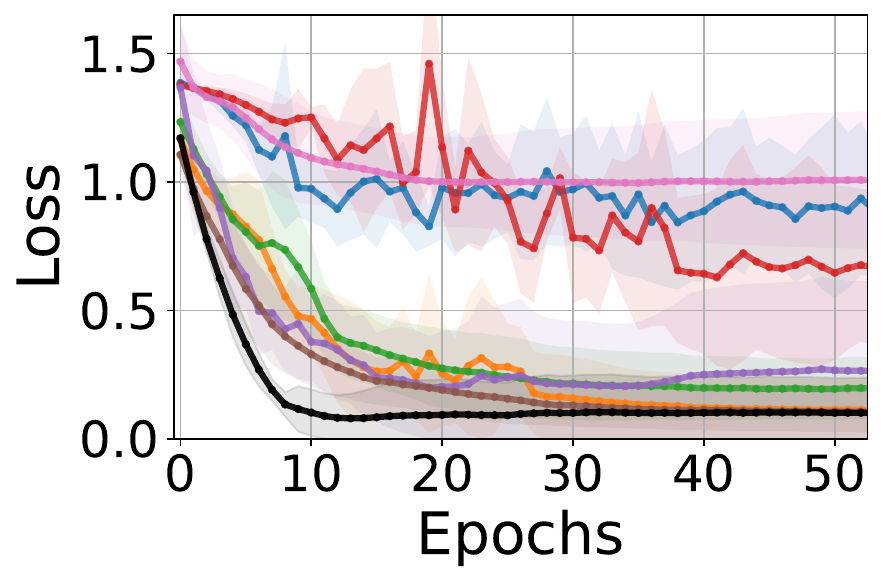}}
\caption{Test loss curves on `BasicMotions' under regular and irregular scenarios using selected methods}
\label{fig:loss_ex}
% \end{figure}
% \\[\baselineskip]
% \begin{figure}[htbp]
\centering\captionsetup{skip=5pt}
\captionsetup[subfigure]{justification=centering, skip=5pt}
    \subfloat{ 
        \includegraphics[width=\linewidth]{figs/run/legend_only.pdf}} \\[-10pt]
    \subfloat[Regular]{
      \includegraphics[width=0.48\linewidth]{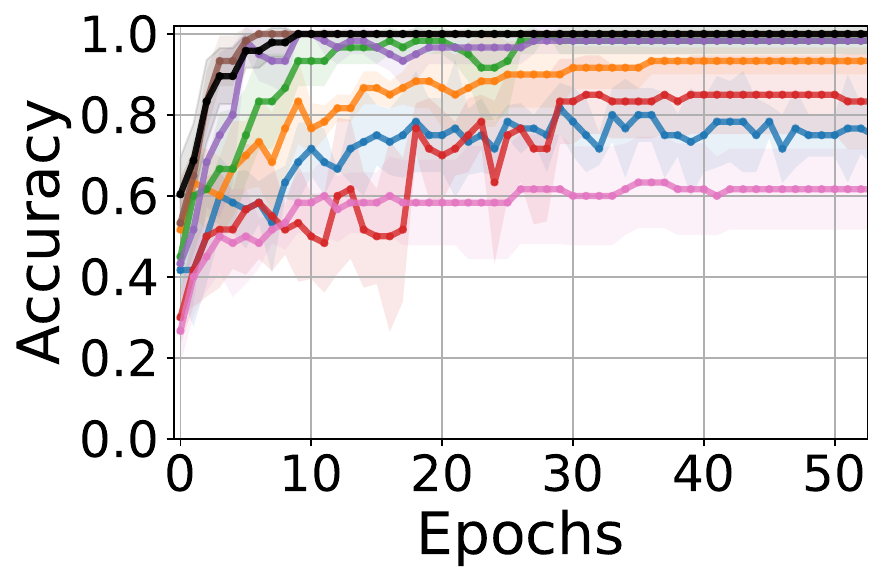}} \hfil
    \subfloat[50\% Missing]{
      \includegraphics[width=0.48\linewidth]{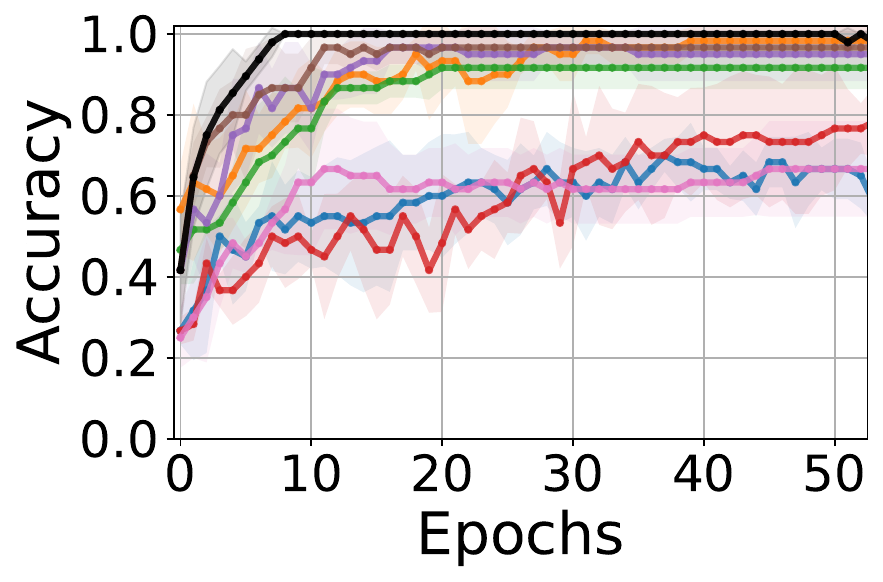}}
\caption{Test accuracy on `BasicMotions' under regular and irregular scenarios using selected methods and FlowPath}
\label{fig:acc_ex}
\end{figure}

\begin{figure}[htbp]
\centering\captionsetup{skip=5pt}
\captionsetup[subfigure]{justification=centering, skip=5pt}
    \subfloat[Irregularly-sampled time series (ISTS)]{
      \includegraphics[width=0.95\linewidth]{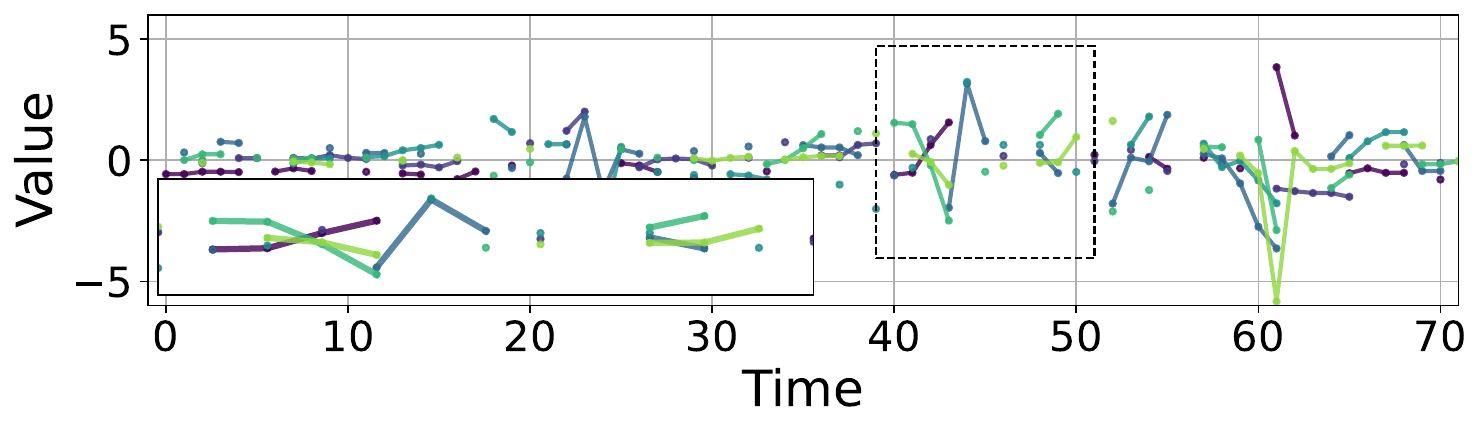}} \\
    \subfloat[MLP-based non-invertible path]{
      \includegraphics[width=0.95\linewidth]{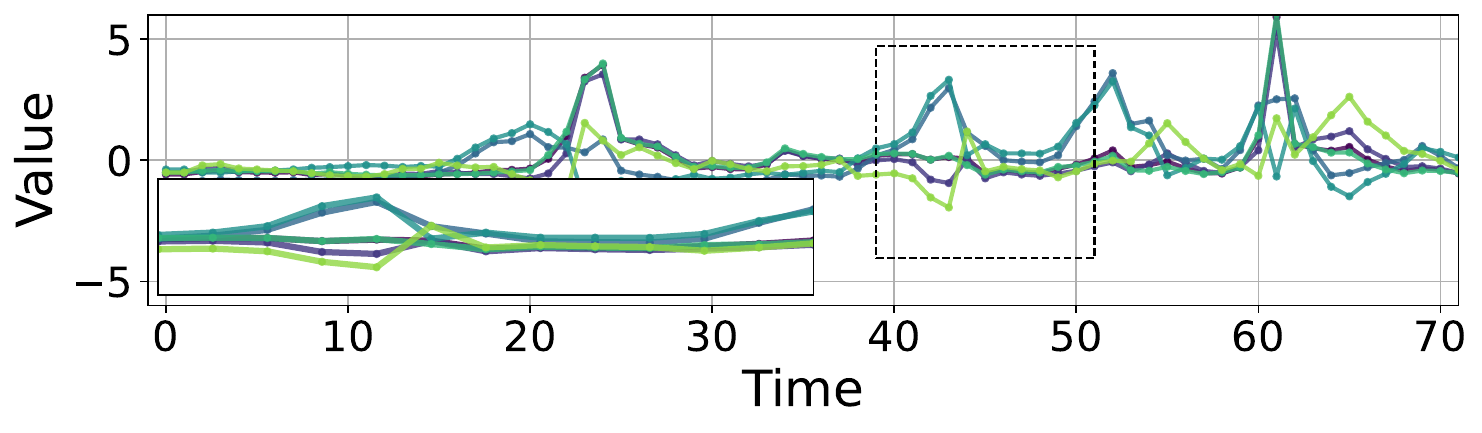}} \\
    \subfloat[Flow-based invertible path]{
      \includegraphics[width=0.95\linewidth]{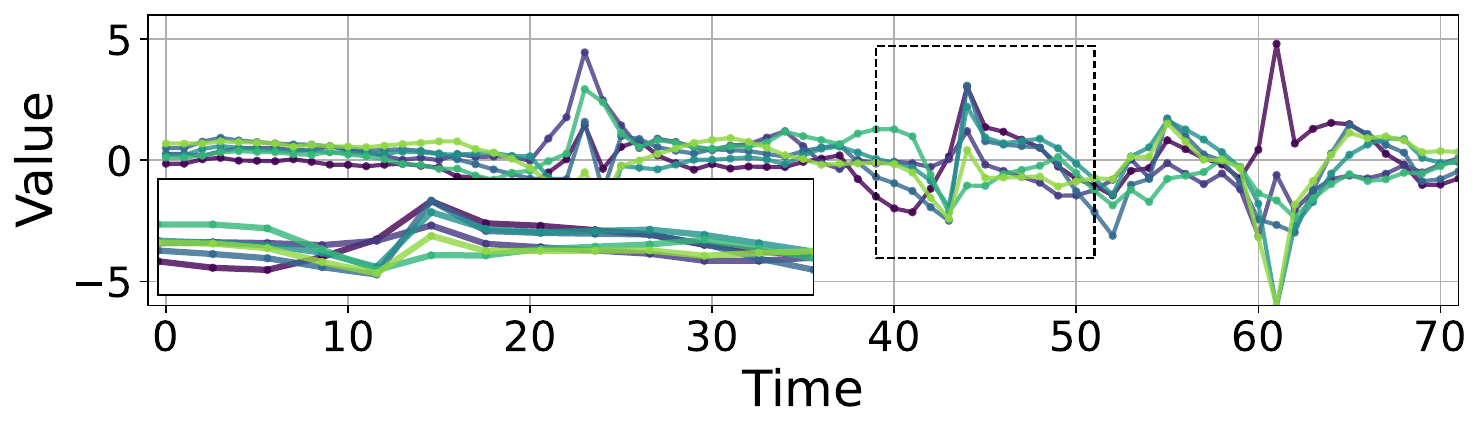}} 
\caption{Qualitative comparison of path construction methods on a sample from the `BasicMotions' dataset with 50\% missingness. (a) Raw and sparse observations. (b) Path generated by a non-invertible MLP, exhibiting a biased curve toward the observed points. (c) Path produced by the proposed FlowPath, capturing a more structured and stable manifold. Each color denotes a separate data dimension.}
\label{fig:flow}
\end{figure}

In Figures~\ref{fig:loss_ex} and \ref{fig:acc_ex}, FlowPath consistently achieves the lowest loss and highest accuracy, demonstrating faster and more stable convergence across both regular and missing data scenarios on the `BasicMotions` dataset. 
This empirical observation of stable learning and robust performance aligns with FlowPath's theoretical properties: the well-posedness of its dynamics (Theorem~\ref{thm:exist_unique}) contributes to stable training, while its generalization capability (Theorem~\ref{thm:generalization}) suggests that strong empirical performance can translate effectively to unseen test data. 

% Figure~\ref{fig:flow} compares path construction methods under 50\% missingness. The MLP-based path (b) appears noisy and overfit to sparse inputs, failing to capture the global structure. In contrast, FlowPath (c) produces a smoother and more coherent path. This stability arises from the invertibility constraint, which promotes information preservation and geometric consistency (Theorem~\ref{thm:preservation}). Additional examples are included in the supplementary material.

Our goal is not to reconstruct the exact path but to learn a meaningful continuous representation $\Phi(t)$ for classification. Figure~\ref{fig:flow} shows a qualitative comparison of the manifolds learned from sparse data. Under highly irregular observations (a), a standard non-invertible multi-layer perceptron (MLP) (b) produces a disordered path that overfits sparse points without capturing the underlying structure. In contrast, FlowPath (c) learns a smoother and more coherent path. 
This stability arises from the invertibility constraint, which promotes information preservation and geometric consistency (Theorem~\ref{thm:preservation}). 
Additional visualizations are provided in the supplementary material.

\subsection{Structural Comparison of Learned Manifolds}
To assess the structure of the learned representations, we analyze trajectories and distributions. 
For clarity, three dimensions from the observation space are shown here. A full analysis is provided in the supplementary material.
Furthermore, we include an extended discussion on NDEs' robustness and manifold learning.

\begin{figure}[H]
\centering\captionsetup{skip=5pt}
  \includegraphics[width=0.90\linewidth]{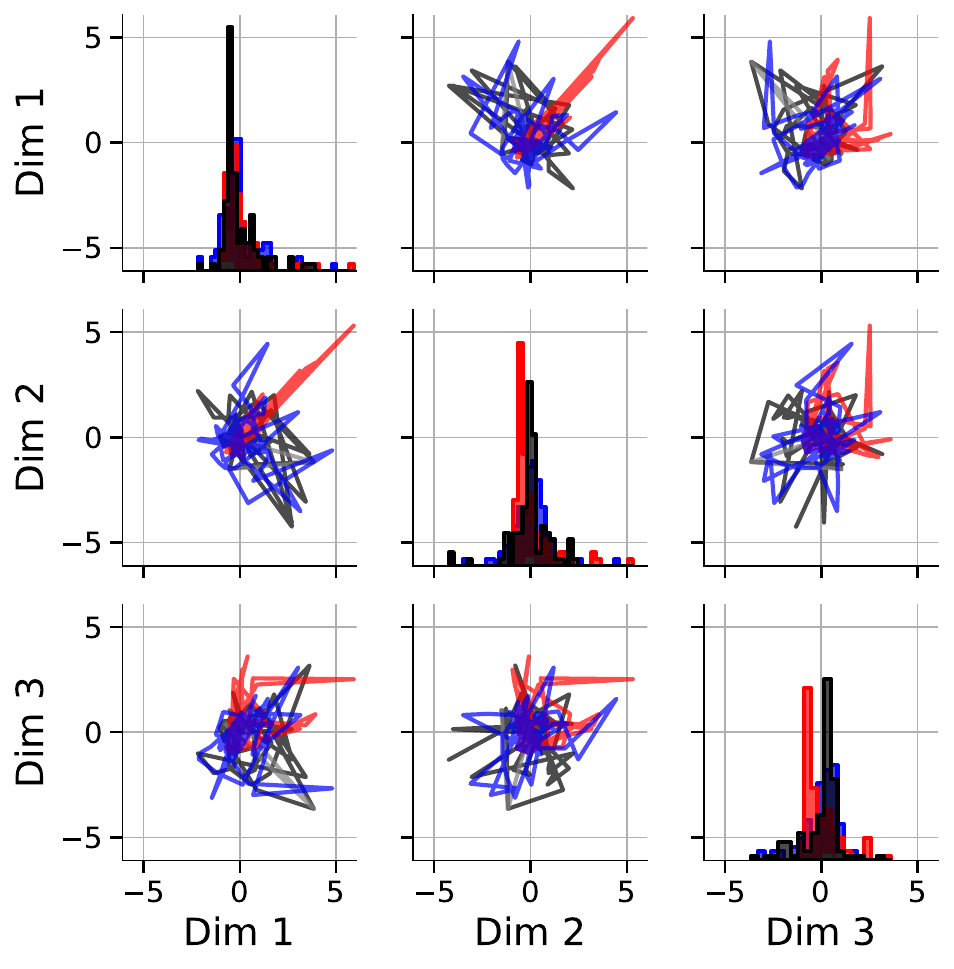}
\caption{
2D projections of learned trajectories from sparse input. Off-diagonal plots show phase-space paths, and diagonal plots show marginal histograms. FlowPath (blue) better matches the ground truth (black) than the MLP (red). % or sparse observations (gray).
}\label{fig:traj_ex}
\end{figure}

\begin{figure}[H]
\centering\captionsetup{skip=5pt}
  \includegraphics[width=0.90\linewidth]{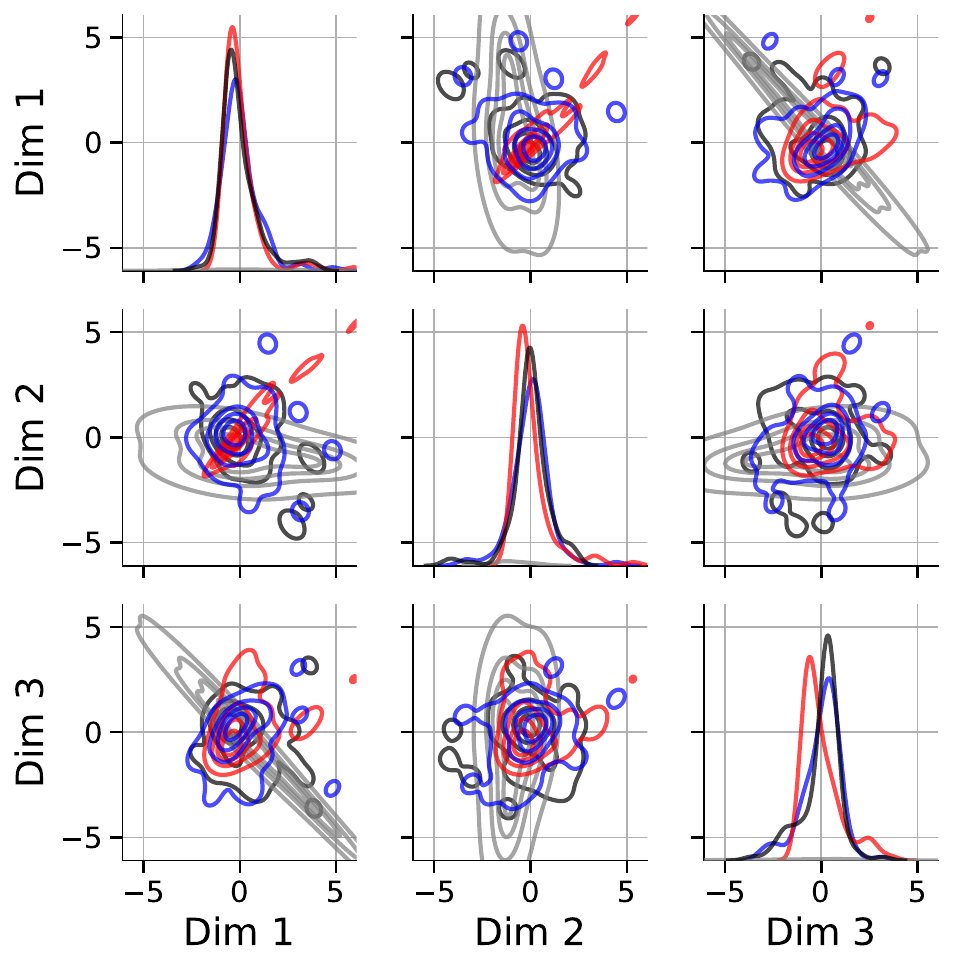}
\caption{
1D and 2D Kernel Density Estimates (KDEs) for each method. FlowPath (blue) aligns more closely with the original distribution (black) than the MLP (red). % or sparse input (gray).
}\label{fig:kde_ex}
\end{figure}

\begin{figure}[H]
\vspace{-2.0em}
\centering\captionsetup{skip=5pt}
\captionsetup[subfigure]{justification=centering, skip=5pt}
    \subfloat[Original data]{
      \includegraphics[width=0.42\linewidth]{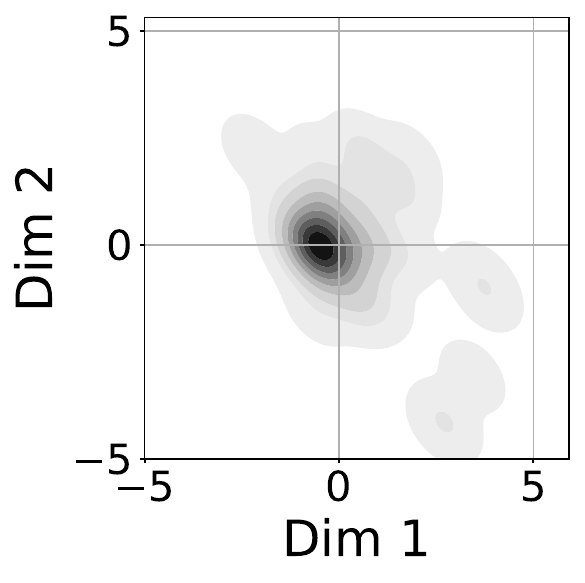}} \hfil
    \subfloat[ISTS observation]{
      \includegraphics[width=0.42\linewidth]{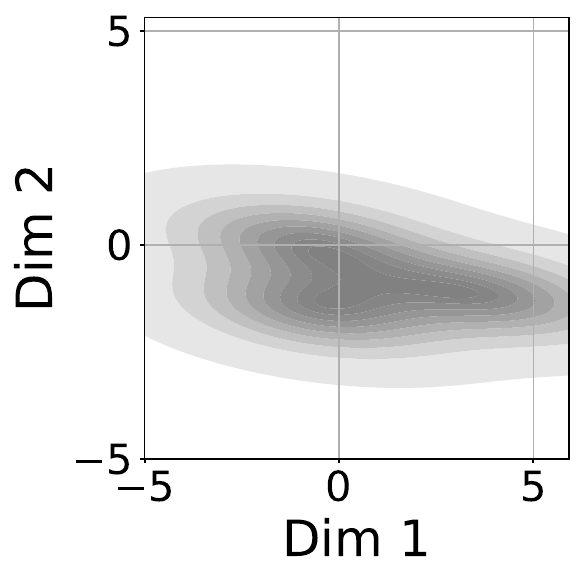}} \\[-5pt]
    \subfloat[MLP-based non-invertible path]{
      \includegraphics[width=0.42\linewidth]{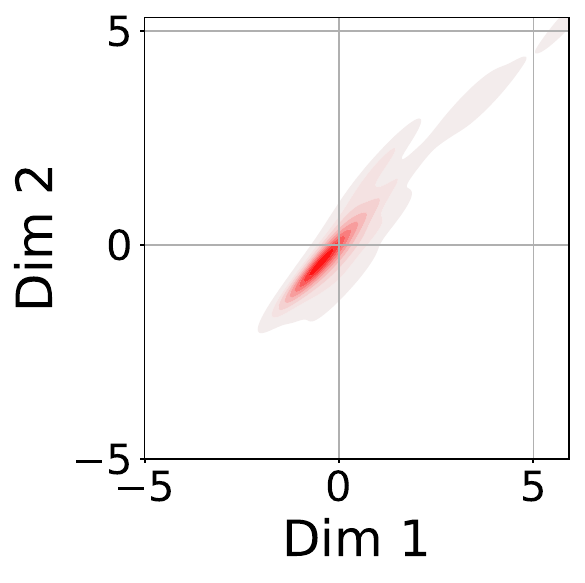}} \hfil
    \subfloat[Flow-based \\ invertible path]{
      \includegraphics[width=0.42\linewidth]{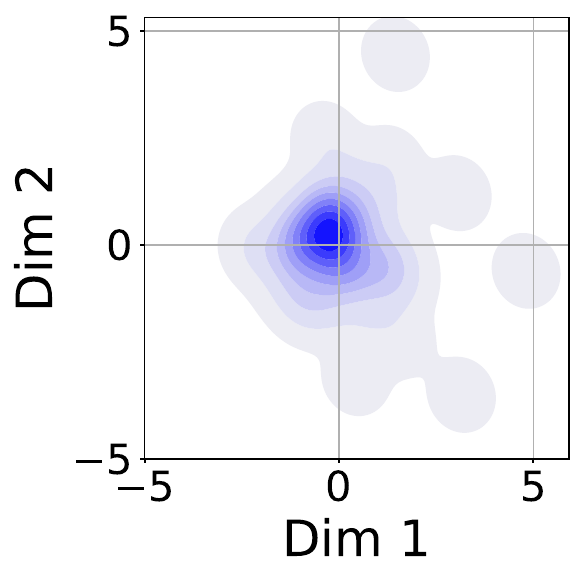}}
\caption{2D KDEs of the learned manifolds between Dim 1 and  Dim 2, using the sparse observation shown in panel (b)}
\label{fig:kde_detail}
\end{figure}

%%%
\begin{table*}[htbp]
\scriptsize\centering\captionsetup{skip=5pt}
% \caption{
% Classification accuracy on 18 benchmark datasets under various scenarios. (Values in parentheses denote the mean standard deviation across different datasets and scenarios. \textbf{Best} and \ul{second-best} results are highlighted, respectively.)}\label{tab:result_all_acc}
\begin{tabular}{@{}lcccccccccc@{}}
\toprule
\multirow{2.5}{*}{\textbf{Methods}} & \multicolumn{2}{c}{\textbf{Regular}}         & \multicolumn{2}{c}{\textbf{30\% Missing}}    & \multicolumn{2}{c}{\textbf{50\% Missing}}    & \multicolumn{2}{c}{\textbf{70\% Missing}}    & \multicolumn{2}{c}{\textbf{Average}}         \\ \cmidrule(lr){2-3}\cmidrule(lr){4-5}\cmidrule(lr){6-7}\cmidrule(lr){8-9}\cmidrule(lr){10-11} % \cmidrule(l){2-11} 
                                  & \textbf{Accuracy}            & \textbf{Rank} & \textbf{Accuracy}            & \textbf{Rank} & \textbf{Accuracy}            & \textbf{Rank} & \textbf{Accuracy}            & \textbf{Rank} & \textbf{Accuracy}            & \textbf{Rank} \\ \midrule
RNN                               & 0.560\avgstd{0.072}          & 10.7          & 0.484\avgstd{0.075}          & 13.3          & 0.471\avgstd{0.082}          & 12.9          & 0.453\avgstd{0.068}          & 13.3          & 0.492\avgstd{0.074}          & 12.6          \\
LSTM                              & 0.588\avgstd{0.067}          & 10.0          & 0.552\avgstd{0.075}          & 9.4           & 0.516\avgstd{0.073}          & 10.5          & 0.505\avgstd{0.067}          & 10.6          & 0.540\avgstd{0.071}          & 10.1          \\
GRU                               & 0.674\avgstd{0.080}          & 6.9           & 0.639\avgstd{0.065}          & 8.0           & 0.611\avgstd{0.076}          & 8.2           & 0.606\avgstd{0.088}          & 8.1           & 0.633\avgstd{0.077}          & 7.8           \\
GRU-$\Delta t$                             & 0.629\avgstd{0.065}          & 9.1           & 0.636\avgstd{0.069}          & 7.6           & 0.651\avgstd{0.068}          & 6.7           & 0.649\avgstd{0.074}          & 7.5           & 0.641\avgstd{0.069}          & 7.7           \\
GRU-D                             & 0.593\avgstd{0.088}          & 10.0          & 0.579\avgstd{0.087}          & 9.8           & 0.580\avgstd{0.075}          & 9.7           & 0.599\avgstd{0.062}          & 9.4           & 0.588\avgstd{0.078}          & 9.7           \\
GRU-ODE                           & 0.663\avgstd{0.072}          & 7.2           & 0.661\avgstd{0.069}          & 6.8           & 0.664\avgstd{0.069}          & 6.5           & 0.659\avgstd{0.081}          & 6.3           & 0.662\avgstd{0.073}          & 6.7           \\
ODE-RNN                           & 0.652\avgstd{0.085}          & 6.8           & 0.632\avgstd{0.076}          & 7.3           & 0.626\avgstd{0.086}          & 7.2           & 0.653\avgstd{0.059}          & 5.8           & 0.641\avgstd{0.076}          & 6.8           \\
ODE-LSTM                          & 0.566\avgstd{0.074}          & 10.4          & 0.518\avgstd{0.069}          & 11.4          & 0.501\avgstd{0.068}          & 12.1          & 0.474\avgstd{0.068}          & 12.3          & 0.515\avgstd{0.070}          & 11.5          \\
Neural CDE                        & 0.681\avgstd{0.073}          & 7.1           & 0.672\avgstd{0.068}          & 7.3           & 0.661\avgstd{0.070}          & 7.0           & 0.652\avgstd{0.091}          & 7.0           & 0.667\avgstd{0.075}          & 7.1           \\
Neural RDE                        & 0.649\avgstd{0.082}          & 7.9           & 0.648\avgstd{0.071}          & 6.9           & 0.633\avgstd{0.078}          & 7.7           & 0.607\avgstd{0.079}          & 8.2           & 0.634\avgstd{0.078}          & 7.7           \\
ANCDE                             & 0.662\avgstd{0.083}          & 7.3           & 0.661\avgstd{0.083}          & 6.8           & 0.639\avgstd{0.080}          & 7.6           & 0.631\avgstd{0.073}          & 7.1           & 0.649\avgstd{0.080}          & 7.2           \\
EXIT                              & 0.595\avgstd{0.087}          & 9.4           & 0.580\avgstd{0.088}          & 9.8           & 0.578\avgstd{0.086}          & 9.5           & 0.564\avgstd{0.072}          & 10.0          & 0.579\avgstd{0.083}          & 9.7           \\
LEAP                              & 0.490\avgstd{0.062}          & 13.1          & 0.459\avgstd{0.070}          & 13.7          & 0.466\avgstd{0.074}          & 12.4          & 0.451\avgstd{0.074}          & 12.7          & 0.466\avgstd{0.070}          & 13.0          \\
DualDynamics                      &  \ul{0.724\avgstd{0.090}}    &  \ul{4.6}     &  \ul{0.720\avgstd{0.088}}    &  \ul{4.9}     &  \ul{0.691\avgstd{0.091}}    &  \ul{4.9}     &  \ul{0.697\avgstd{0.098}}    &  \ul{4.6}     &  \ul{0.708\avgstd{0.092}}    &  \ul{4.8}     \\
Neural Flow                       & 0.530\avgstd{0.069}          & 11.8          & 0.531\avgstd{0.072}          & 9.9           & 0.537\avgstd{0.073}          & 9.4           & 0.535\avgstd{0.082}          & 9.7           & 0.533\avgstd{0.074}          & 10.2          \\ \midrule
\textbf{FlowPath}                 & \textbf{0.731\avgstd{0.083}} & \textbf{3.7}  & \textbf{0.743\avgstd{0.091}} & \textbf{3.1}  & \textbf{0.726\avgstd{0.084}} & \textbf{3.6}  & \textbf{0.718\avgstd{0.090}} & \textbf{3.5}  & \textbf{0.730\avgstd{0.087}} & \textbf{3.5}  \\ \bottomrule
\end{tabular}
\caption{
Classification accuracy on 18 benchmark datasets under various scenarios. (Values in parentheses denote the mean standard deviation across different datasets and scenarios. \textbf{Best} and \ul{second-best} results are highlighted, respectively.)}\label{tab:result_all_acc}
\end{table*}
%%%

Figure~\ref{fig:traj_ex} highlights structural differences in the learned manifolds. Sparse observations (gray) lack temporal continuity due to missing data. The MLP path (red) shows instability and deviation from the original geometry (black), likely overfitting to observed points. FlowPath (blue), in contrast, maintains a smoother and more coherent trajectory, reflecting the benefits of the invertibility constraint.

Figure~\ref{fig:kde_ex} shows KDEs of the learned manifolds. FlowPath achieves better alignment with the true distribution across both marginal and joint spaces. This ability to reconstruct distributional structure under sparsity contributes to its downstream classification performance. To illustrate this in detail, we zoom in on one representative case. 

Figure~\ref{fig:kde_detail} qualitatively analyzes the ability to recover the underlying manifold (a) from only the sparse observations in (b). The non-invertible MLP (c) learns a distorted distribution that is misaligned with the unknown original geometry. In contrast, FlowPath (d) yields a more coherent manifold that better aligns with the original distribution, suggesting that the structural constraint of invertibility is crucial for learning a faithful representation under sparsity.

\subsection{Performance Comparison with 18 Datasets}\label{sec:performance_comparison}

In Table~\ref{tab:result_all_acc}, we present a comparative analysis of average classification accuracy across various scenarios, including the regular scenario and irregular scenarios (with missing rates of 30\%, 50\%, and 70\%), alongside an aggregate average across all settings. 
Notably, FlowPath maintains robustness under high rates of missing data, where other methods experience a significant performance drop. Additionally, FlowPath not only achieves the highest average accuracy but also consistently ranks among the top-performing models overall across the four missingness scenarios.

\begin{table}[htbp]
\scriptsize\centering\captionsetup{skip=5pt}
% \caption{Pairwise comparison ({wins / ties / losses}) against {DualDynamics} across different missing rates.}\label{tab:one_to_one}
\begin{tabular}{lccccc}
\toprule
\textbf{Settings}      & \textbf{Regular} & \textbf{\begin{tabular}[c]{@{}c@{}}30\% \\ Missing\end{tabular}} & \textbf{\begin{tabular}[c]{@{}c@{}}50\% \\ Missing\end{tabular}} & \textbf{\begin{tabular}[c]{@{}c@{}}70\% \\ Missing\end{tabular}} & \textbf{\begin{tabular}[c]{@{}c@{}}All \\ Settings\end{tabular}} \\ \midrule
Wins  & 8  & 11 & 11 & 10 & 40 \\
Ties  & 2  & 1  & 2  & 1  & 6  \\
Losses & 8  & 6  & 5  & 7  & 26 \\
\bottomrule
\end{tabular}
\caption{Pairwise comparison ({wins / ties / losses}) against {DualDynamics} across different missing rates.}\label{tab:one_to_one}
\end{table}

As shown in Table~\ref{tab:one_to_one}, our method outperforms the strongest baseline, {DualDynamics}, under missing-value conditions. Performance is similar in the regular setting, but our model gains more wins as the missing rate increases, showing greater robustness to incomplete observations.
Complete baseline comparisons and pairwise statistical test results are provided in the supplementary material.

Table~\ref{tab:result_all_ablation} presents an ablation study that investigates the sources of FlowPath’s performance improvements. The results indicate that while a simple learned path incorporated with an MLP yields modest gains over the fixed-path Neural CDE baseline, the structurally constrained FlowPath consistently achieves the highest performance across all levels of missingness. Furthermore, additional experiments reported in the supplementary material compare three invertible flow architectures, including ResNet, GRU, and Coupling Flow, demonstrating that FlowPath’s effectiveness is robust to the choice of flow architecture.

\begin{table}[htbp]
\scriptsize\centering\captionsetup{skip=5pt}
% \caption{
% Ablation study comparing the standard Neural CDE, a non-invertible MLP, and the proposed FlowPath.}\label{tab:result_all_ablation}
\begin{tabular}{@{}lC{1.0cm}C{1.0cm}C{1.0cm}C{1.0cm}C{1.0cm}@{}}
\toprule
% \textbf{Settings} & \textbf{Regular}    & \textbf{30\% Missing} & \textbf{50\% Missing} & \textbf{70\% Missing} & \textbf{Average}    \\ \midrule
\textbf{Settings}      & \textbf{Regular} & \textbf{\begin{tabular}[c]{@{}c@{}}30\% \\ Missing\end{tabular}} & \textbf{\begin{tabular}[c]{@{}c@{}}50\% \\ Missing\end{tabular}} & \textbf{\begin{tabular}[c]{@{}c@{}}70\% \\ Missing\end{tabular}} & \textbf{\begin{tabular}[c]{@{}c@{}}All \\ Settings\end{tabular}} \\ \midrule
Neural CDE         & 0.681\avgstd{0.073} & 0.672\avgstd{0.068}   & 0.661\avgstd{0.070}   & 0.652\avgstd{0.091}   & 0.667\avgstd{0.075} \\ % \midrule
\quad + MLP        & 0.705\avgstd{0.077} & 0.700\avgstd{0.084}   & 0.695\avgstd{0.091}   & 0.655\avgstd{0.084}   & 0.689\avgstd{0.084} \\ \midrule
\textbf{FlowPath}          & 0.731\avgstd{0.083} & 0.743\avgstd{0.091}   & 0.726\avgstd{0.084}   & 0.718\avgstd{0.090}   & 0.730\avgstd{0.087} \\ \bottomrule
\end{tabular}
\caption{
Ablation study comparing the standard Neural CDE, a non-invertible MLP, and the proposed FlowPath.}\label{tab:result_all_ablation}
\end{table}

\subsection{Performance-computation Analysis}
\begin{figure}[htbp]
\centering\captionsetup{skip=5pt}
\captionsetup[subfigure]{justification=centering, skip=5pt}
    \subfloat[Performance on `Regular' scenario]{ 
      \includegraphics[width=\linewidth]{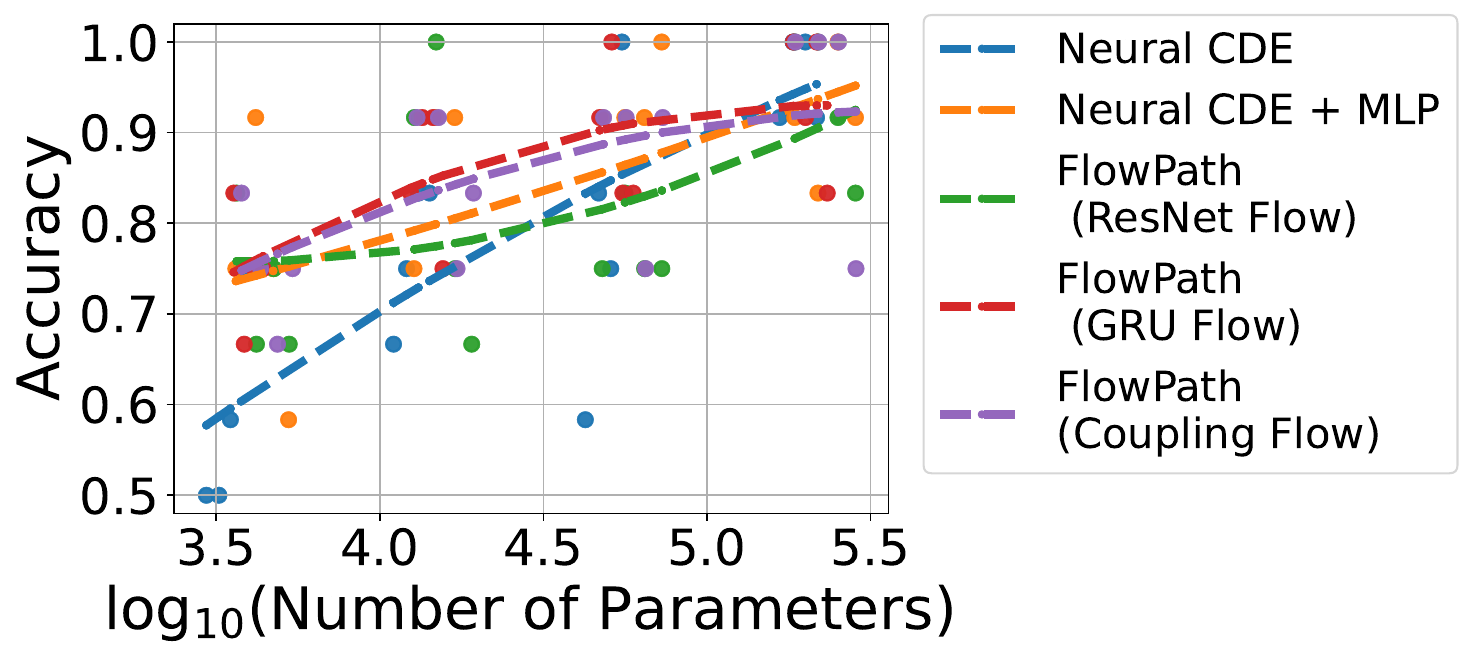}} \\
    \subfloat[Performance on `50\% Missing' scenario]{
      \includegraphics[width=\linewidth]{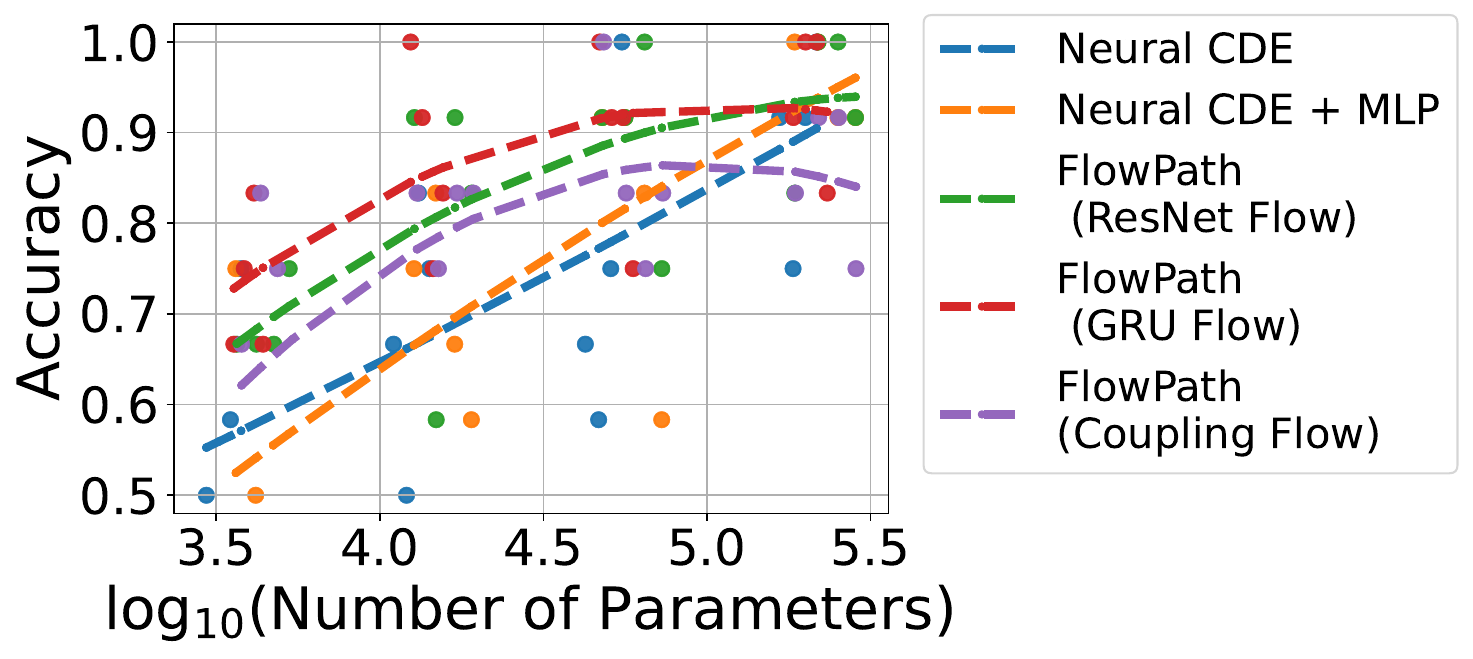}} 
\caption{Analysis of performance-computation trade-off. % on the `BasicMotions' dataset. 
Figures show \(\log_{10}\) of the parameter count versus classification accuracy.
Each colored dot corresponds to specific hyperparameters % (\(n_l\), \(n_h\), and flow model) 
and the dashed line indicates the fitted trend. 
% Notice that FlowPath typically achieves higher accuracy than standard Neural CDE, especially under missingness, at the cost of mildly increased runtime.
}
\vspace{-1.0em}
\label{fig:computation}
\end{figure}

We conduct an in-depth comparison across different hyperparameter settings, varying the flow architecture (ResNet, GRU, Coupling Flow), the number of layers $n_l$, and hidden sizes $n_h$.
Figures~\ref{fig:computation}~(a) and (b) compare classification accuracy to model complexity in the regular and 50\% missing settings, respectively. 
FlowPath variants consistently outperform the Neural CDE baseline at matched parameter counts, with performance gaps widening under higher missingness. MLP-based paths may perform adequately in certain cases; however, their performance significantly deteriorates under irregular scenarios.
These results show that FlowPath’s invertible, learnable path improves performance with moderate computational cost, making it a practical choice.

%-----------------------------------------------------------------------
\section{Experiment: Real-world HAR Dataset}\label{sec:experiments_real}
To evaluate robustness on a real-world task, we used the PAMAP2 dataset \citep{a_reiss_introducing_2012}, a challenging benchmark for sensor-based Human Activity Recognition (HAR). 
Each participant was equipped with three sensors positioned on the wrist, chest, and ankle, as well as a heart rate monitor. The dataset comprises 5333 segments spanning 8 daily living activities, captured using 17 distinct sensor modalities.

We followed the exact preprocessing protocol, data splits, and sensor dropout methodology proposed by \citet{zhang_graph-guided_2022}. 
To simulate irregularity, we randomly removed 60\% of data points from each segment, using the same random mask across all experiments.
After that, we generated scenarios by randomly removing a subset of input sensors with dropout rates ranging from 10\% to 50\%. This approach increases the challenge of the problem, as the omitted sensors are randomly selected for each sample. For every test sample, a subset of sensors was designated as missing, and all observations from these sensors were replaced with zeros.

\subsection{Benchmark Methods}
We compare the proposed FlowPath with the following state-of-the-art baselines, including \textit{Transformer}~\citep{vaswani_attention_2017}, \textit{Trans-mean} (Transformer with mean imputation), \textit{GRU-D}~\citep{chen_neural_2018}, \textit{SeFT}~\citep{horn_set_2020}, \textit{mTAND}~\citep{shukla_multi-time_2021}, \textit{Raindrop}~\citep{zhang_graph-guided_2022}, and \textit{Neural CDE}~\citep{kidger_neural_2020}.

For the original dataset without sensor dropout, we have included additional benchmarks such as \textit{DGM$^2$-O}~\citep{wu_dynamic_2021}, \textit{IP-Net}~\citep{shukla_interpolation-prediction_2019}, \textit{MTGNN}~\citep{wu_connecting_2020}, \textit{TITD}~\citep{ji_titd_2025}, and \textit{CoFormer}~\citep{y_wei_compatible_2023}. 
For fair comparison, we used performance results as reported in the original literature.

\subsection{Performance Comparison}
As detailed in the supplementary material, optimal hyperparameters were selected via grid search. We employed the standard Neural CDE and compared various flow configurations for the proposed FlowPath. Regardless of the flow configuration, FlowPath consistently demonstrated superior classification performance compared to Neural CDE, as shown in Table~\ref{tab:ablation_pam}. Based on hyperparameter tuning, the GRU Flow was selected for this dataset and scenarios.

\begin{table}[htbp]
\scriptsize\centering\captionsetup{skip=5pt}
% \captionof{table}{
% F1 score comparison on the PAMAP2 dataset between Neural CDE and the proposed FlowPath model.}\label{tab:ablation_pam}
\begin{tabular}{@{}cccccc@{}}
\toprule
% \multirow{2.5}{*}{\textbf{Settings}} & \multicolumn{2}{c}{\textbf{Neural CDE}} & \multicolumn{3}{c}{\textbf{FlowPath}}                \\ \cmidrule(lr){2-3} \cmidrule(lr){4-6} 
\multirow{2.5}{*}{\textbf{\begin{tabular}[c]{@{}c@{}}Sensor \\ Dropout\end{tabular}}} & \multicolumn{2}{c}{\textbf{Neural CDE}} & \multicolumn{3}{c}{\textbf{FlowPath}}                \\ \cmidrule(lr){2-3} \cmidrule(lr){4-6} 
                                & \textbf{Spline}     & \textbf{MLP}      & \textbf{ResNet} & \textbf{GRU}   & \textbf{Coupling} \\ \midrule
\textbf{0\%}                & 95.0 \std{0.4}      & 93.9 \std{0.4}    & 94.6 \std{0.5}  & 95.6 \std{0.2} & 94.8 \std{0.2}    \\
\textbf{10\%}           & 87.3 \std{1.3}      & 86.4 \std{1.0}    & 87.2 \std{0.3}  & 88.3 \std{0.4} & 87.4 \std{0.3}    \\
\textbf{20\%}           & 76.8 \std{0.6}      & 75.1 \std{0.8}    & 79.3 \std{1.0}  & 79.1 \std{1.9} & 77.8 \std{0.6}    \\
\textbf{30\%}           & 67.1 \std{1.9}      & 68.2 \std{0.8}    & 71.4 \std{0.7}  & 70.0 \std{0.7} & 70.2 \std{1.0}    \\
\textbf{40\%}           & 57.7 \std{1.6}      & 58.8 \std{2.8}    & 65.9 \std{1.9}  & 59.6 \std{1.0} & 60.8 \std{1.1}    \\
\textbf{50\%}           & 51.6 \std{1.0}      & 50.7 \std{2.0}    & 57.9 \std{0.2}  & 54.1 \std{1.0} & 54.2 \std{3.1}    \\ \bottomrule
\end{tabular}
\captionof{table}{
F1 score comparison on the PAMAP2 dataset between Neural CDE and the proposed FlowPath model.}\label{tab:ablation_pam}
\end{table}

Table~\ref{tab:result_pam_regular} compares classification metrics for various methods under the irregularly-sampled PAMAP2 dataset with all sensors. FlowPath’s learnable invertible flow further boosts performance, indicating its superior handling of ISTS.

\begin{table}[htbp]
\scriptsize\centering\captionsetup{skip=5pt}
% \captionof{table}{
% Classification performance on the PAMAP2 dataset with all sensors under irregular sampling.% (See details in \citet{zhang_graph-guided_2022})
% }\label{tab:result_pam_regular}
% \begin{tabular}{@{}clcccc@{}}
\begin{tabular}{@{}lcccc@{}}
\toprule
% \textbf{}                                                                     & 
\textbf{Methods} & \textbf{Accuracy} & \textbf{Precision} & \textbf{Recall} & \textbf{F1 Score} \\ \midrule
% \parbox[t]{1mm}{\multirow{13}{*}{\rotatebox[origin=c]{90}{\textbf{Original}}}} & 
Transformer                 & 83.5\std{1.5}          & 84.8\std{1.5}          & 86.0\std{1.2}          & 85.0\std{1.3}          \\
Trans-mean                           & 83.7\std{2.3}          & 84.9\std{2.6}          & 86.4\std{2.1}          & 85.1\std{2.4}          \\
GRU-D                                & 83.3\std{1.6}          & 84.6\std{1.2}          & 85.2\std{1.6}          & 84.8\std{1.2}          \\
SeFT                                 & 67.1\std{2.2}          & 70.0\std{2.4}          & 68.2\std{1.5}          & 68.5\std{1.8}          \\
mTAND                                & 74.6\std{4.3}          & 74.3\std{4.0}          & 79.5\std{2.8}          & 76.8\std{3.4}          \\
Raindrop                             & 88.5\std{1.5}          & 89.9\std{1.5}          & 89.9\std{0.6}          & 89.8\std{1.0}          \\
DGM$^2$-O                            & 82.4\std{2.3}          & 85.2\std{1.2}          & 83.9\std{2.3}          & 84.3\std{1.8}          \\
IP-Net                               & 74.3\std{3.8}          & 75.6\std{2.1}          & 77.9\std{2.2}          & 76.6\std{2.8}          \\
MTGNN                                & 83.4\std{1.9}          & 85.2\std{1.7}          & 86.1\std{1.9}          & 85.9\std{2.4}          \\
TITD                                 & 90.7\std{0.5}          & 90.2\std{1.3}          & 90.7\std{0.9}          & 90.5\std{1.5}          \\
CoFormer                             & 91.2\std{0.6}          & 92.4\std{0.7}          & 93.7\std{0.7}          & 92.8\std{0.5}          \\
Neural CDE                           & \ul{94.2\std{0.5}}    & \ul{95.2\std{0.4}}    & \ul{94.8\std{0.5}}    & \ul{95.0\std{0.4}}    \\ \midrule
\textbf{FlowPath}                          & \textbf{94.8\std{0.2}} & \textbf{95.8\std{0.4}} & \textbf{95.5\std{0.2}} & \textbf{95.6\std{0.2}} \\ \bottomrule
\end{tabular}
\captionof{table}{
Classification performance on the PAMAP2 dataset with all sensors under original irregular sampling.% (See details in \citet{zhang_graph-guided_2022})
}\label{tab:result_pam_regular}
\end{table}

Furthermore, Table~\ref{tab:result_pam} summarizes the classification outcomes for the PAMAP2 dataset using a variety of benchmark methods. 
Across a comparison with all benchmark methods, our approach consistently delivers superior performance across all evaluated metrics. 

%%%
\begin{table}[!htb]
\scriptsize\centering\captionsetup{skip=5pt}
% \caption{
% Classification performance on the PAMAP2 dataset with sensor dropout rates ranging from 10\% to 50\%.
% %Classification performance on PAMAP2 dataset with regular setting and sensor dropout from 10\% to 50\%. % \textbf{Best} and \ul{second-best} results are highlighted, respectively.
% }
% \label{tab:result_pam}
\begin{tabular}{@{}L{0.1cm}lcccc@{}}
\toprule
% \textbf{Settings}                                                                 
~ & \textbf{Methods} & \textbf{Accuracy} & \textbf{Precision} & \textbf{Recall} & \textbf{F1 Score} \\ \midrule
\multirow{8}{*}{\rotatebox{90}{\makebox[0pt][c]{\textbf{10\% Missing}}}} & Transformer                            & 60.9 \std{12.8}   & 58.4 \std{18.4}    & 59.1 \std{16.2} & 56.9 \std{18.9}   \\
                                                                                  & Trans-mean                             & 62.4 \std{3.5}    & 59.6 \std{7.2}     & 63.7 \std{8.1}  & 62.7 \std{6.4}    \\
                                                                                  & GRU-D                                  & 68.4 \std{3.7}    & 74.2 \std{3.0}     & 70.8 \std{4.2}  & 72.0 \std{3.7}    \\
                                                                                  & SeFT                                   & 40.0 \std{1.9}    & 40.8 \std{3.2}     & 41.0 \std{0.7}  & 39.9 \std{1.5}    \\
                                                                                  & mTAND                                  & 53.4 \std{2.0}    & 54.8 \std{2.7}     & 57.0 \std{1.9}  & 55.9 \std{2.2}    \\
                                                                                  & Raindrop                               & 76.7 \std{1.8}    & 79.9 \std{1.7}     & 77.9 \std{2.3}  & 78.6 \std{1.8}    \\
                                                                                  & Neural CDE                             & \ul{85.8 \std{1.4}} & \ul{88.8 \std{1.2}}  & \ul{86.3 \std{1.3}} & \ul{87.3 \std{1.3}}  \\ \cmidrule(l){2-6} 
                                                                                  & \textbf{FlowPath}                      & \textbf{86.6 \std{1.1}} & \textbf{89.5 \std{0.8}}  & \textbf{87.4 \std{0.4}} & \textbf{88.3 \std{0.4}}  \\ \midrule
\multirow{8}{*}{\rotatebox{90}{\makebox[0pt][c]{\textbf{20\% Missing}}}} & Transformer                            & 62.3 \std{11.5}   & 65.9 \std{12.7}    & 61.4 \std{13.9} & 61.8 \std{15.6}   \\
                                                                                  & Trans-mean                             & 56.8 \std{4.1}    & 59.4 \std{3.4}     & 53.2 \std{3.9}  & 55.3 \std{3.5}    \\
                                                                                  & GRU-D                                  & 64.8 \std{0.4}    & 69.8 \std{0.8}     & 65.8 \std{0.5}  & 67.2 \std{0.0}    \\
                                                                                  & SeFT                                   & 34.2 \std{2.8}    & 34.9 \std{5.2}     & 34.6 \std{2.1}  & 33.3 \std{2.7}    \\
                                                                                  & mTAND                                  & 45.6 \std{1.6}    & 49.2 \std{2.1}     & 49.0 \std{1.6}  & 49.0 \std{1.0}    \\
                                                                                  & Raindrop                               & 71.3 \std{2.5}    & 75.8 \std{2.2}     & 72.5 \std{2.0}  & 73.4 \std{2.1}    \\
                                                                                  & Neural CDE                             & \ul{74.4 \std{0.5}} & \ul{81.0 \std{0.8}}  & \ul{74.7 \std{1.0}} & \ul{76.8 \std{0.6}}  \\ \cmidrule(l){2-6} 
                                                                                  & \textbf{FlowPath}                      & \textbf{77.0 \std{1.7}} & \textbf{82.4 \std{1.4}}  & \textbf{77.1 \std{2.3}} & \textbf{79.1 \std{1.9}}  \\ \midrule
\multirow{8}{*}{\rotatebox{90}{\makebox[0pt][c]{\textbf{30\% Missing}}}} & Transformer                            & 52.0 \std{11.9}   & 55.2 \std{15.3}    & 50.1 \std{13.3} & 48.4 \std{18.2}   \\
                                                                                  & Trans-mean                             & 65.1 \std{1.9}    & 63.8 \std{1.2}     & \textbf{67.9 \std{1.8}} & 64.9 \std{1.7}    \\
                                                                                  & GRU-D                                  & 58.0 \std{2.0}    & 63.2 \std{1.7}     & 58.2 \std{3.1}  & 59.3 \std{3.5}    \\
                                                                                  & SeFT                                   & 31.7 \std{1.5}    & 31.0 \std{2.7}     & 32.0 \std{1.2}  & 28.0 \std{1.6}    \\
                                                                                  & mTAND                                  & 34.7 \std{5.5}    & 43.4 \std{4.0}     & 36.3 \std{4.7}  & 39.5 \std{4.4}    \\
                                                                                  & Raindrop                               & 60.3 \std{3.5}    & 68.1 \std{3.1}     & 60.3 \std{3.6}  & 61.9 \std{3.9}    \\
                                                                                  & Neural CDE                             & \ul{65.5 \std{1.8}} & \ul{73.1 \std{1.8}}  & 64.9 \std{2.3}  & \ul{67.1 \std{1.9}}  \\ \cmidrule(l){2-6} 
                                                                                  & \textbf{FlowPath}                      & \textbf{67.3 \std{0.2}} & \textbf{75.1 \std{0.4}}  & \ul{67.7 \std{0.7}}   & \textbf{70.0 \std{0.7}}  \\ \midrule
\multirow{8}{*}{\rotatebox{90}{\makebox[0pt][c]{\textbf{40\% Missing}}}} & Transformer                            & 43.8 \std{14.0}   & 44.6 \std{23.0}    & 40.5 \std{15.9} & 40.2 \std{20.1}   \\
                                                                                  & Trans-mean                             & 48.7 \std{2.7}    & 55.8 \std{2.6}     & 54.2 \std{3.0}  & 55.1 \std{2.9}    \\
                                                                                  & GRU-D                                  & 47.7 \std{1.4}    & 63.4 \std{1.6}     & 44.5 \std{0.5}  & 47.5 \std{0.0}    \\
                                                                                  & SeFT                                   & 26.8 \std{2.6}    & 24.1 \std{3.4}     & 28.0 \std{1.2}  & 23.3 \std{3.0}    \\
                                                                                  & mTAND                                  & 23.7 \std{1.0}    & 33.9 \std{6.5}     & 26.4 \std{1.6}  & 29.3 \std{1.9}    \\
                                                                                  & Raindrop                               & \ul{57.0 \std{3.1}} & 65.4 \std{2.7}     & \ul{56.7 \std{3.1}} & \ul{58.9 \std{2.5}}  \\
                                                                                  & Neural CDE                             & 55.8 \std{0.5}    & \textbf{67.7 \std{1.5}}  & 54.5 \std{2.3}  & 57.7 \std{1.6}    \\ \cmidrule(l){2-6} 
                                                                                  & \textbf{FlowPath}                      & \textbf{58.1 \std{0.1}} & \ul{67.0 \std{0.9}}  & \textbf{57.1 \std{0.7}} & \textbf{59.6 \std{1.0}}  \\ \midrule
\multirow{8}{*}{\rotatebox{90}{\makebox[0pt][c]{\textbf{50\% Missing}}}} & Transformer                            & 43.2 \std{2.5}    & 52.0 \std{2.5}     & 36.9 \std{3.1}  & 41.9 \std{3.2}    \\
                                                                                  & Trans-mean                             & 46.4 \std{1.4}    & 59.1 \std{3.2}     & 43.1 \std{2.2}  & 46.5 \std{3.1}    \\
                                                                                  & GRU-D                                  & 49.7 \std{1.2}    & 52.4 \std{0.3}     & 42.5 \std{1.7}  & 47.5 \std{1.2}    \\
                                                                                  & SeFT                                   & 26.4 \std{1.4}    & 23.0 \std{2.9}     & 27.5 \std{0.4}  & 23.5 \std{1.8}    \\
                                                                                  & mTAND                                  & 20.9 \std{3.1}    & 35.1 \std{6.1}     & 23.0 \std{3.2}  & 27.7 \std{3.9}    \\
                                                                                  & Raindrop                               & 47.2 \std{4.4}    & 59.4 \std{3.9}     & 44.8 \std{5.3}  & 47.6 \std{5.2}    \\
                                                                                  & Neural CDE                             & \ul{49.9 \std{1.7}} & \ul{65.5 \std{0.8}}  & \ul{48.9 \std{1.5}} & \ul{51.6 \std{1.0}}  \\ \cmidrule(l){2-6} 
                                                                                  & \textbf{FlowPath}                      & \textbf{52.0 \std{0.6}} & \textbf{66.3 \std{1.9}}  & \textbf{51.3 \std{0.7}} & \textbf{54.1 \std{1.0}}  \\ \bottomrule
\end{tabular}
\caption{
Classification performance on the PAMAP2 dataset with sensor dropout rates ranging from 10\% to 50\%.
%Classification performance on PAMAP2 dataset with regular setting and sensor dropout from 10\% to 50\%. % \textbf{Best} and \ul{second-best} results are highlighted, respectively.
\vspace{-1.0em}
}
\label{tab:result_pam}
\end{table}
%%%

\section{Experiment: Real-world Medical Dataset}

We further evaluated model performance on a medical classification benchmark on the PhysioNet Sepsis dataset~\citep{reyna_early_2020}, which includes 40,335 patients and 34 temporal variables. Following~\citet{kidger_neural_2020}, we compared models with and without observation intensity (OI), an index reflecting patient severity, and measured classification performance by AUROC.

Benchmark models and reported performance from \citet{kidger_neural_2020}, \citet{oh_dualdynamics_2025}, and \citet{oh_comprehensive_2025}. We also include variants of stable Neural SDEs \citep{oh_stable_2024}, including Neural Langevin-type SDE (LSDE), Neural Linear Noise SDE (LNSDE), and Neural Geometric SDE (GSDE).
% For further implementation details and benchmark methods, please refer to the original studies.
For implementation details and benchmark methods, please consult the original studies.

%%%
\begin{table}[htb]
\scriptsize\centering\captionsetup{skip=5pt}
% \caption{AUROC Comparison on PhysioNet Sepsis}\label{tab:classification}
\begin{tabular}{@{}lC{2.1cm}C{2.1cm}@{}}
\toprule
\multirow{2.5}{*}{\textbf{Methods}} & \multicolumn{2}{c}{\textbf{Test AUROC}}                                     \\ \cmidrule(l){2-3} 
\multicolumn{1}{c}{}                                  & \textbf{With OI}                     & \textbf{Without OI}                  \\ \midrule
GRU-$\Delta t$                                        & 0.878 \std{0.006} & 0.840 \std{0.007}   \\
GRU-D                                                 & 0.871 \std{0.022} & 0.850 \std{0.013}   \\
GRU-ODE                                               & 0.852 \std{0.010} & 0.771 \std{0.024}   \\
ODE-RNN                                               & 0.874 \std{0.016} & 0.833 \std{0.020}   \\
Latent-ODE                                            & 0.787 \std{0.011} & 0.495 \std{0.002}   \\
ACE-NODE                                              & 0.804 \std{0.010} & 0.514 \std{0.003}   \\
Neural CDE                                            & 0.880 \std{0.006} & 0.776 \std{0.009}   \\
ANCDE                                                 & 0.900 \std{0.002} & 0.823 \std{0.003}   \\
EXIT                                                  & 0.913 \std{0.002} & 0.836 \std{0.003}   \\
DualDynamics                                          & 0.918 \std{0.003} & 0.873 \std{0.004}   \\
Neural SDE                                            & 0.799 \std{0.007} & 0.796 \std{0.006}   \\
Neural LSDE                                           & 0.909 \std{0.004} & 0.879 \std{0.008}   \\
Neural LNSDE                                          & 0.911 \std{0.002} & 0.881 \std{0.002}   \\
Neural GSDE                                           & 0.909 \std{0.001} & 0.884 \std{0.002}   \\ \midrule
\textbf{FlowPath}                                     &                   &                     \\
\quad -- ResNet                                       & 0.919\std{0.005}  & 0.869\std{0.006}    \\
\quad -- GRU                                          & 0.918\std{0.005}  & 0.870\std{0.005}    \\
\quad -- Coupling                                     & 0.916\std{0.006}  & 0.866\std{0.002}    \\ \bottomrule
\end{tabular}
\caption{AUROC Comparison on PhysioNet Sepsis}\label{tab:classification}
\end{table}
%%%

As shown in Table~\ref{tab:classification}, FlowPath variants achieve the highest or near-highest AUROC among all baselines. % This shows that the proposed flow-based formulation improves classification robustness on the PhysioNet Sepsis dataset.
These results suggest that learning continuous flow variations offers a reliable inductive bias for modeling complex temporal dynamics in time series analysis.

%-----------------------------------------------------------------------
\section{Conclusion}\label{sec:conclusion}
We present \textit{FlowPath}, a framework for modeling ISTS by learning the control path of a Neural CDE through an invertible neural flow. This approach replaces fixed interpolation with a data-adaptive, structure-preserving transformation that more accurately captures the underlying temporal geometry. Experimental results on 18 benchmark datasets and a real-world case study demonstrate improved classification accuracy and robustness, particularly under conditions of high missingness. 
These findings highlight the importance of modeling the geometry of the control path, not just the dynamics along it, to build reliable continuous-time models for real-world applications.

While FlowPath improves robustness by learning a well-structured control path, its use of invertible flows introduces additional parameters. Exploring more efficient architectures could reduce this overhead. Furthermore, our evaluation also focuses on classification, and the performance of FlowPath on tasks such as forecasting and generative modeling remains an important area for future investigation.

%-----------------------------------------------------------------------
\section{Acknowledgments}
We thank the teams and individuals for their efforts in the real-world dataset preparation and curation for our research, especially the UEA \& UCR repository for the numerous datasets that we extensively analyzed.

This research was supported by Basic Science Research Program through the National Research Foundation of Korea (NRF) funded by the Ministry of Education (RS-2024-00407852); 
the Institute of Information \& Communications Technology Planning \& Evaluation(IITP) grant funded by the Korea government(MSIT)(No.RS-2020-II201336, Artificial Intelligence graduate school support(UNIST));
the Institute of Information \& Communications Technology Planning \& Evaluation(IITP) grant funded by the Korea government(MSIT) (No. RS-2025-25442824, AI STAR Fellowship); 
the National Research Foundation of Korea(NRF) grant funded by the Korea government(MSIT) (No.RS-2023-00218913, RS-2025-00563597, No. RS-2025-02216640)

% {
% \small
\bibliography{references}
% }

% \clearpage
% \newpage
% \input{ReproducibilityChecklist/checklist}

%%%%%%%%%%%%%%%%%%%%%%%%%%%%%%%%%%%%%%%%%%%%%%%%%%%%%%%%%%%%
\clearpage
\newpage
\appendix
\onecolumn

\renewcommand{\theequation}{\arabic{equation}}
\setcounter{equation}{0} % Reset equation counter for the appendix

\section{Theoretical Considerations}\label{appendix:theory}
This section provides expanded explanations and clarifications for the theoretical claims stated in the manuscript. We restate key assumptions where necessary and detail each step of the arguments.

\subsection{Proof of Theorem~\ref{thm:preservation}}

\begin{proof}
Fix \(\theta=(\theta_f,\theta_F)\) and omit the explicit dependence on \(\theta\).  Set
\[
  v(t,\vz)=f(t,\vz)\dot\Phi(t),
  \quad\text{where}\quad
  \dot\vz(t)=v\bigl(t,\vz(t)\bigr).
\]
Since \(\Phi\) is a \(C^1\) diffeomorphism and \(f\) is Lipschitz in \(\vz\), the vector field \(v(t,\vz)\) is \(C^1\) in \(\vz\) and continuous in \(t\).  By the continuity of Liouville equation, the density \(p(\vz,t)\) satisfies
\[
  \frac{\partial p}{\partial t}(\vz,t)
  + \nabla_{\vz}\!\cdot\bigl(p(\vz,t)\,v(t,\vz)\bigr)
  = 0.
\]
Along the solution \(\vz(t)\), the total time‐derivative of \(\log p\) is
\begin{align*}
  \frac{\rd}{\rd t}\log p\bigl(\vz(t),t\bigr)   &= \frac{1}{p}\Bigl(\frac{\partial p}{\partial t}      + \dot\vz(t)^{\mathsf T}\nabla_{\vz}p\Bigr)\\
  &= -\frac{1}{p}\,\nabla_{\vz}\cdot\bigl(p v\bigr)      = -\nabla_{\vz}\cdot\,v\bigl(t,\vz(t)\bigr).
\end{align*}
Substituting \(v(t,\vz)=f(t,\vz)\dot\Phi(t)\) gives
\[
  \frac{\rd}{\rd t}\log p\bigl(\vz(t)\bigr)
  = -\mathrm{div}_{\vz}\Bigl(f\bigl(t,\vz(t)\bigr)\dot\Phi(t)\Bigr),
\]
Since \(v\) is \(C^1\) in \(\vz\) over the compact domain \([0,T]\times\mathbb{R}^{d_z}\), its divergence is bounded.  Hence the density cannot collapse to zero or blow up to infinity in finite time.
\end{proof}

\subsection{Proof of Theorem~\ref{thm:exist_unique}}

\begin{proof}
Fix \(\theta_f\) and \(\theta_F\), and omit explicit dependence on \(\theta=(\theta_f,\theta_F)\).  Recall the FlowPath integral equation
\[
  \vz(t)   = \vz(0)   + \int_{0}^{t} f\bigl(\tau,\vz(\tau);\theta_f\bigr)\dot{\Phi}\bigl(\tau;\theta_F\bigr)\rd\tau.
\]
Define the combined vector field
\[
  v\bigl(t,\vz\bigr)
  = f\bigl(t,\vz;\theta_f\bigr)\,\dot{\Phi}\bigl(t;\theta_F\bigr).
\]
Since \(\Phi(\cdot;\theta_F)\) is \(C^1\) on \([0,T]\), set
\[
  M_\Phi =\sup_{t\in[0,T]}\|\dot{\Phi}(t;\theta_F)\| < \infty.
\]
By assumption, \(f(t,\cdot;\theta_f)\) is \(L_f\)-Lipschitz in \(\vz\), uniformly in \(t\).  Hence for any \(\vz_1,\vz_2\),
\begin{align*}
  \|v(t,\vz_1)-v(t,\vz_2)\|
  &= \bigl\|f(t,\vz_1)\dot{\Phi}(t)
         - f(t,\vz_2)\dot{\Phi}(t)\bigr\|\\
  &\le M_\Phi\|f(t,\vz_1)-f(t,\vz_2)\|
  \le L_f M_\Phi\|\vz_1-\vz_2\|.
\end{align*}
Thus \(v(t,\vz)\) is Lipschitz in \(\vz\) and continuous in \(t\). Applying the Picard–Lindelöf theorem to the ODE
\[
  \dot{\vz}(t) = v\bigl(t,\vz(t)\bigr),
  \quad \vz(0)=\vz_0,
\]
we conclude there is a unique continuous solution \(\vz(t)\) on \([0,T]\).  
\end{proof}

\subsection{Proof of Theorem~\ref{thm:generalization}}

\begin{proof}
Assume throughout that the time horizon \(T>0\) is fixed and finite, and assume that
\[
  \|f(t,\vz;\theta_f)\|\leq F_\infty
  \quad\text{for all }t\in[0,T],\vz\in\R^{d_z}.
\]
Fix two parameter vectors \(\theta_i=(\theta_{f,i},\theta_{F,i})\) and write \(\vz_i(t)=\vz_{\theta_i}(t)\) for \(i=1,2\).  Let
\[
  \Delta\vz(t)=\vz_1(t)-\vz_2(t),
  \quad
  \Delta\theta=(\Delta\theta_f,\Delta\theta_F)
  =(\theta_{f,1}-\theta_{f,2},\,\theta_{F,1}-\theta_{F,2}).
\]
By the FlowPath integral,
\[
  \Delta\vz(t)
  = \int_{0}^{t}
    \Bigl[
      f\bigl(s,\vz_1(s);\theta_{f,1}\bigr)\dot\Phi\bigl(s;\theta_{F,1}\bigr)
      -f\bigl(s,\vz_2(s);\theta_{f,2}\bigr)\dot\Phi\bigl(s;\theta_{F,2}\bigr)
    \Bigr]\rd s.
\]
We add and subtract intermediate terms to split the integrand:
\begin{align*}
  f_1\dot\Phi_1 - f_2\dot\Phi_2 &=\bigl[f_1 - f(s,\vz_1;\theta_{f,2})\bigr]\dot\Phi_1
   + f(s,\vz_1;\theta_{f,2})\bigl[\dot\Phi_1-\dot\Phi_2\bigr]\\
  & +\bigl[f(s,\vz_1;\theta_{f,2}) - f(s,\vz_2;\theta_{f,2})\bigr]\dot\Phi_2,
\end{align*}
where \(f_i=f(s,\vz_i(s);\theta_{f,i})\) and \(\dot\Phi_i=\dot\Phi(s;\theta_{F,i})\).  Using
\[
  \|\dot\Phi_i\|\leq M_\Phi,
  \quad
  \|\dot\Phi_1-\dot\Phi_2\|\leq L_\Phi\|\Delta\theta_F\|,
  \quad
  \|f(s,\vz;\theta_{f,1}) - f(s,\vz;\theta_{f,2})\|\leq L_\theta\|\Delta\theta_f\|,
\]
and the Lipschitz constant \(L_f\) in \(\vz\), we get
\[
  \|\Delta\vz(t)\|
\leq
  \Bigl(L_\theta\,M_\Phi + F_\infty\,L_\Phi\Bigr)\|\Delta\theta\|\,t
  + L_f\,M_\Phi\int_{0}^{t}\|\Delta\vz(s)\|\rd s.
\]
By Grönwall’s inequality, we confirm that $\vz$ is Lipschitz:
\[
  \|\Delta\vz(T)\|   \leq   A\|\Delta\theta\|.
\]
where $A:=\bigl(L_\theta M_\Phi + F_\infty L_\Phi\bigr)T\,
  e^{L_f M_\Phi T}$. Since the loss \(\mathcal{L}\) is also \(\ell\)-Lipschitz,
\[
  \bigl|\mathcal{L}(\vz_1(T)) - \mathcal{L}(\vz_2(T))\bigr|
  \;\le\;
  \ell\,\|\Delta\vz(T)\|
  \;\le\;
  L\,\|\Delta\theta\|,
  \quad
  L=\ell\,A.
\]
Thus \(\mathcal{F}_\Theta\) is \(L\)-Lipschitz in \(\theta\). Now, we exactly follow the standard argument in ~\citep{bartlett_spectrally-normalized_2017,chen_generalization_2020}. If \(\Theta\) has diameter \(D\), covering‐number arguments yield
\[
  \mathcal{N}\bigl(\mathcal{F}_\Theta,\eta\bigr)
  \leq
  \Bigl(1 + \tfrac{L\,D}{\eta}\Bigr)^{d}.
\]
A Dudley‐entropy‐integral bound then shows
\[
  \mathrm{Rad}(\mathcal{F}_\Theta)
  = \mathcal{O}\!\Bigl(\frac{L\,D\,\sqrt{d}}{\sqrt{n}}\Bigr)
  =: \kappa\frac{1}{\sqrt{n}}.
\]
Finally, symmetrization and McDiarmid’s concentration imply that, for any \(\delta>0\), with probability at least \(1-\delta\),
\[
  \bigl|\mathcal{R}_{\mathrm{true}}(\widehat f)
  -\mathcal{R}_{\mathrm{emp}}(\widehat f)\bigr|
  \leq
  2\mathrm{Rad}(\mathcal{F}_\Theta)
  + \sqrt{\frac{\ln(1/\delta)}{2n}}
  \leq
  \alpha\frac{1}{\sqrt{n}}
  + \sqrt{\frac{\ln(1/\delta)}{2n}},
\]
with \(\alpha=2\kappa\).  This completes the proof.
\end{proof}

\section{Implementation details of FlowPath}\label{appendix:implementation}
\subsection{Perspectives on NDEs' Robustness and Manifold Learning}
FlowPath's contributions can advance two ongoing research thrusts in the NDE community: the development of robust models, and the integration of manifold learning to respect the intrinsic geometry of data.

One strategy for robust NDEs is to constrain the learned dynamics through physical or mathematical priors on the vector field $f$. For example, stability can be enforced via conservation laws \citep{white_stabilized_2023}, or by designing stable classes of stochastic systems to mitigate noise sensitivity \citep{oh_stable_2024}, in the perspective of Neural Stochastic Differential Equations (Neural SDEs). 
Compared to that, FlowPath leaves the dynamics function $f$ unconstrained and instead achieves robustness by imposing a strong structural prior on the control path $\Phi(t)$ itself. This regularizes the system by ensuring the data manifold that drives the dynamics is well-behaved, rather than by limiting the flexibility of the dynamics that evolve upon it.

Another active area focuses on learning the data manifold explicitly, moving beyond the Euclidean assumptions of standard NDEs. One approach constrains the dynamics to remain tangent to a learned manifold, as in Neural Manifold ODEs~\citep{lou_neural_2020}. Related advances in generative modeling apply continuous flows or diffusion processes on Riemannian manifolds \citep{mathieu_riemannian_2020,huang_riemannian_2022}, and recent work in metric flow matching emphasizes the importance of respecting intrinsic geometry \citep{chen_flow_2024,kapusniak_metric_2024}. 
FlowPath shares this manifold-aware motivation but with a more modular design. Our method learns a data-driven manifold to define the control path, while the dynamics themselves evolve in the ambient space of the NDE.
This decouples the learning of the geometric structure from the evolution rule.

These different philosophies highlight a trade-off between expressivity and structure. Models like Characteristic-Neural ODEs \citep{xu_characteristic_2023} are designed for maximum flexibility, enabling them to learn complex, non-homeomorphic transformations that can represent intersecting trajectories. 
In contrast, FlowPath intentionally leverages the structural constraint of a diffeomorphism as a powerful inductive bias. We posit that for robust classification from sparse observations, preserving the topology of the input data via an invertible map is a more beneficial constraint than allowing for arbitrary topological changes. This reflects a clear modeling choice, prioritizing a stable and well-behaved latent representation as the foundation for robust downstream performance.

%%%
\subsection{Design of Flow Model}\label{sec:flow}
In our framework, the trainable control path \(\Phi(t)\) is realized through an invertible mapping \(F\colon [0,T]\times \mathbb{R}^{d_x}\to \mathbb{R}^{d_x}\) following \citet{bilos_neural_2021}. Building on classical results in dynamical systems, we ensure \(\Phi\) maintains bijectivity and differentiability, preserving local volumes in the latent space. We implement three architecture choices:

\paragraph{ResNet Flow.} Extending residual networks to continuous time \citep{chen_neural_2018}, we define:
\begin{equation*}
    F(t,\bm{z}) = \bm{z} + \varphi(t) \, f\bigl(t,\bm{z}\bigr),
\end{equation*}
where \(\varphi\) encodes time dependence and \(f\) is a Lipschitz neural network. Invertibility follows from spectral normalization \citep{behrmann_invertible_2019}.

\paragraph{GRU Flow.} Following \citet{brouwer_gru-ode-bayes_2019}, we adapt the GRU architecture to continuous time:
\begin{equation*}
    F(t,\bm{z}) = \bm{z} + \varphi(t) \bigl(1 - f_{1}(t,\bm{z})\bigr) \odot \bigl(f_{2}(t,\bm{z}) - \bm{z}\bigr),
\end{equation*}
preserving memory capabilities while ensuring invertibility through careful gating constraints.

\paragraph{Coupling Flow.} Inspired by normalizing flows \citep{dinh_nice_2015,dinh_density_2017}, we partition \(\bm{z}\) and apply:
\begin{equation*}
    F(t)_{d_1} = \bm{z}_{d_1} \exp\bigl(u(t,\bm{z}_{d_2})\varphi_u(t)\bigr) + v(t,\bm{z}_{d_2})\varphi_v(t),
\end{equation*}
where \(d_1 \cup d_2\) forms a complete partition, enabling efficient Jacobian computation.

We design our flow architecture to balance expressiveness with computational efficiency, guided by \citet{papamakarios_normalizing_2021}’s normalizing-flow framework. In practice, we leverage Hutchinson’s trace estimator \citep{hutchinson_stochastic_1989} for efficient Jacobian-related computations, further reducing the overhead of flow-based modeling~\citep{grathwohl_ffjord_2019,bilos_neural_2021,oh_dualdynamics_2025}.
The flow architecture is designed to balance computational efficiency and representational expressiveness. We optimize its configuration during the tuning process and validate our choices through an ablation study.

\subsection{Adjoint-sensitive method of FlowPath}\label{sec:adjoint}

The adjoint method is a powerful technique often used to calculate the gradients of systems governed by differential equations \citep{pollini_adjoint_2018}. 
The adjoint-sensitive method, proposed to solve Neural ODEs, reduces the memory requirements and computational burden that would otherwise be associated with backpropagating gradients through the solution of an ODE or CDE~\citep{chen_neural_2018,kidger_neural_2020}.
The primary concept of the adjoint-sensitive method is to introduce an adjoint state, usually denoted as $\bm{a}(t)$, which satisfies an auxiliary differential equation running backward in time.
Algorithm~\ref{alg:FlowPath_adjoint} summarizes the procedure for performing the parameter update using the adjoint-sensitivity framework in FlowPath.
\begin{algorithm}[!ht]
\small\caption{Adjoint-sensitivity method for FlowPath}\label{alg:FlowPath_adjoint}
\begin{algorithmic}[1]
% \REQUIRE Learning rate \(\eta\)
\STATE Initialize parameters \(\theta = \{\theta_f,\;\theta_{F},\}, \;\theta_{\text{MLP}}\)
\STATE \textbf{Forward:} 
\STATE numerically solve \Eqref{eq:FlowPath} for \(\vz(t)\) over \(t\in[0,T]\)
% \STATE Compute loss \(\mathcal{L}\bigl(\vz(T)\bigr)\)
\STATE Compute loss \(\mathcal{L}\bigl(\text{MLP}(\vz(T);\;\theta_{\text{MLP}}),\;y\bigr)\)

\STATE \textbf{Backward:}
\STATE \quad (a) Set \(\bm{a}(T)=\tfrac{\partial\mathcal{L}}{\partial \vz(T)}\)
\STATE \quad (b) Numerically integrate backward (\(t=T\to 0\))
\begin{align*}
\frac{\mathrm{d}\bm{a}(t)}{\mathrm{d}t}
=
-\,
\bm{a}(t)^{\mathsf{T}}
\,
\frac{\partial}{\partial \vz}
\Bigl(
f\bigl(t,\vz(t);\theta_f\bigr)\,\dot{\Phi}(t;\theta_{F})
\Bigr),
% \quad
% t=T\to 0
\end{align*}
\STATE \quad (c) Compute
\begin{align*}
\frac{\partial\mathcal{L}}{\partial \theta}
=
\int_{0}^{T}
\bm{a}(t)^{\mathsf{T}}
\,
\frac{\partial}{\partial \theta}
\Bigl[
f\bigl(t,\vz(t);\theta_f\bigr)\,\dot{\Phi}(t;\theta_{F})\Bigr]
\,\mathrm{d}t
\end{align*}

\STATE \textbf{Update:} \(\theta \;\leftarrow\;\theta - \eta\,\tfrac{\partial\mathcal{L}}{\partial \theta}\),  where \(\eta\) is learning rate 

\STATE \textbf{Return:} \(\theta\)
\end{algorithmic}
\end{algorithm}
 
% \paragraph{Forward Dynamics.}
% Let \(\vz(0)=h(\bm{x}(0))\). Over \([0,T]\),
% \begin{equation*}\label{eq:flowpath_forward}
% \vz(t)
% =
% \vz(0)
% +
% \int_{0}^{t}
% f\bigl(\tau,\vz(\tau);\theta_f\bigr)\,\dot{\Phi}\bigl(\tau;\theta_{F}\bigr)\,\mathrm{d}\tau,
% \end{equation*}
% where \(\Phi\) is learned from parameters \(\theta_{F}\) and the vector field \(f(\cdot,\cdot;\theta_f)\) is a neural net. 
Denote all parameters collectively by \(\theta=(\theta_f,\theta_{F})\). Let \(\mathcal{L}\) be a scalar loss depending on \(\vz(T)\).
%
% \paragraph{Adjoint State.}
Then, define an adjoint \(\bm{a}(t)\in\mathbb{R}^{d_z}\) by integrating backward in time:
\begin{equation*}
\frac{\mathrm{d}\bm{a}(t)}{\mathrm{d}t}
=
-\,
\bm{a}(t)^{\mathsf{T}}
\,
\frac{\partial}{\partial \vz}
\Bigl[
f\bigl(t,\vz(t);\theta_f\bigr)\,\dot{\Phi}\bigl(t;\theta_{F}\bigr)
\Bigr],
% \quad
% \bm{a}(T)
% =
% \frac{\partial\mathcal{L}}{\partial \vz(T)}.
\end{equation*}
where \(\bm{a}(T) = \frac{\partial\mathcal{L}}{\partial \vz(T)}\).
Solving this ODE backward (\(t=T\to 0\)) provides a continuous \(\bm{a}(t)\) without needing to store all states \(\{\vz(t)\}\).
%
% \paragraph{Gradient Computation.}
Once \(\bm{a}(t)\) is determined, the derivative of \(\mathcal{L}\) with respect to \(\theta\) is
\begin{equation*}
\frac{\partial\mathcal{L}}{\partial \theta}
=
\int_{0}^{T}
\bm{a}(t)^{\mathsf{T}}
\,
\frac{\partial}{\partial \theta}
\Bigl[
f\bigl(t,\vz(t);\theta_f\bigr)\,\dot{\Phi}\bigl(t;\theta_{F}\bigr)
\Bigr]
\,\mathrm{d}t,
\end{equation*}
which also involves evaluating \(\vz(t)\) at certain query points.
%
% \paragraph{Memory Efficiency.}
By avoiding explicit backprop through the entire forward trajectory, only the final state \(\vz(T)\) and the backward ODE for \(\bm{a}(t)\) are integrated. 
This approach significantly reduces memory overhead for training large models or datasets.

Then, \(\vz(T)\) is fed into a multi-layer perceptron (MLP) classifier, and the cross-entropy loss is computed between the predicted label \(\hat{y} = \text{MLP}(\vz(T);\;\theta_{\text{MLP}})\) and the true label \(y\).

% \clearpage
\section{Detailed results on time series classification with missingness}\label{appendix:result_all}
\subsection{Experimental settings}
The experiments were conducted on an Ubuntu 22.04 LTS server, which had an Intel(R) Xeon(R) Gold 6242 CPU and a cluster of NVIDIA A100 40GB GPUs. The source code for our experiments is available at \url{https://github.com/yongkyung-oh/FlowPath}.

\paragraph{Description of datasets.}
We performed classification experiments using 18 datasets spanning three distinct domains from the University of East Anglia (UEA) and the University of California Riverside (UCR) Time Series Classification Repository\footnote{\url{http://www.timeseriesclassification.com/}}~\citep{bagnall_uea_2018,h_a_dau_ucr_2019}. These experiments utilized the Python library \texttt{sktime}~\citep{loning_sktime_2019}. 
As summarized in Table~\ref{tab:data}, datasets in the Sensor domain contain only univariate time series, while datasets in the other domains include both univariate and multivariate time series. To address the issue of varying sequence lengths, all time series were uniformly scaled to match the length of the longest sequence~\citep{oh_stable_2024,oh_dualdynamics_2025,oh_tandem_2025}. This scaling approach ensured consistent input dimensions across datasets for modeling. 

After aligning the series lengths, random missing values, including 30\%, 50\%, and 70\%, were introduced into each time series to simulate real-world scenarios. The datasets were then partitioned into training, validation, and testing subsets, using a ratio of 70\%, 15\%, and 15\% respectively. Unlike the original splits from \citet{bagnall_uea_2018}, which varied across datasets, this partitioning ensured a uniform evaluation framework. 
Finally, the altered time series for each variable were integrated into a unified dataset. To evaluate model robustness, multiple cross-validation iterations were performed, each with a unique random seed for variability and statistical reliability.

\begin{table*}[htbp]
\scriptsize\centering\captionsetup{skip=5pt}
\caption{Description of datasets for the classification task}\label{tab:data}
\begin{tabular}{@{}clrrrr@{}}
\toprule
% \textbf{Domain}                         & \multicolumn{1}{c}{\textbf{Dataset}} & \multicolumn{1}{c}{\textbf{\begin{tabular}[c]{@{}c@{}}Total number of \\ samples\end{tabular}}} & \multicolumn{1}{c}{\textbf{\begin{tabular}[c]{@{}c@{}}Number of \\ classes\end{tabular}}} & \multicolumn{1}{c}{\textbf{\begin{tabular}[c]{@{}c@{}}Dimension of \\ time series\end{tabular}}} & \multicolumn{1}{c}{\textbf{\begin{tabular}[c]{@{}c@{}}Length of \\ time series\end{tabular}}} \\ \midrule
\textbf{Domain} & \textbf{Dataset} & \textbf{Total number of samples} & \textbf{Number of classes} & \textbf{Dimension of time series} & \textbf{Length of time series} \\ \midrule
\multirow{6}{*}{\textbf{Motion \& HAR}} & \textbf{BasicMotions}                & 80                                                   & 4                                              & 6                                                     & 100                                                \\
                                        & \textbf{Epilepsy}                    & 275                                                  & 4                                              & 3                                                     & 206                                                \\
                                        & \textbf{PickupGestureWiimoteZ}       & 100                                                  & 10                                             & 1                                                     & 29-361                                             \\
                                        & \textbf{ShakeGestureWiimoteZ}        & 100                                                  & 10                                             & 1                                                     & 40-385                                             \\
                                        & \textbf{ToeSegmentation1}            & 268                                                  & 2                                              & 1                                                     & 277                                                \\
                                        & \textbf{ToeSegmentation2}            & 166                                                  & 2                                              & 1                                                     & 343                                                \\ \midrule
\multirow{6}{*}{\textbf{ECG \& EEG}}    & \textbf{Blink}                       & 950                                                  & 2                                              & 4                                                     & 510                                                \\
                                        & \textbf{ECG200}                      & 200                                                  & 2                                              & 1                                                     & 96                                                 \\
                                        & \textbf{SelfRegulationSCP1}          & 561                                                  & 2                                              & 6                                                     & 896                                                \\
                                        & \textbf{SelfRegulationSCP2}          & 380                                                  & 2                                              & 7                                                     & 1152                                               \\
                                        & \textbf{StandWalkJump}               & 27                                                   & 3                                              & 4                                                     & 2500                                               \\
                                        & \textbf{TwoLeadECG}                  & 1162                                                 & 2                                              & 1                                                     & 82                                                 \\ \midrule
\multirow{6}{*}{\textbf{Sensor}}        & \textbf{DodgerLoopDay}               & 158                                                  & 7                                              & 1                                                     & 288                                                \\
                                        & \textbf{DodgerLoopGame}              & 158                                                  & 2                                              & 1                                                     & 288                                                \\
                                        & \textbf{DodgerLoopWeekend}           & 158                                                  & 2                                              & 1                                                     & 288                                                \\
                                        & \textbf{Lightning2}                  & 121                                                  & 2                                              & 1                                                     & 637                                                \\
                                        & \textbf{Lightning7}                  & 143                                                  & 7                                              & 1                                                     & 319                                                \\
                                        & \textbf{Trace}                       & 200                                                  & 4                                              & 1                                                     & 275                                                \\ \bottomrule
\end{tabular}
\end{table*}

\paragraph{Training strategies.}
In every model and dataset, we seek to minimize the validation loss by optimizing the hyperparameters as suggested by \citet{oh_stable_2024}. The hyperparameters are optimized using the following scheme: Learning rate $\eta \in \{10^{-4}, 10^{-1}\}$, sampled from a log-uniform distribution; Number of layers $n_l \in \{1, 2, 3, 4\}$, determined via grid search; Hidden vector dimensions $n_h \in \{16, 32, 64, 128\}$, selected through grid search; Batch size chosen from $\{16, 32, 64, 128\}$, adjusted according to the total data size.

It is important that for the sake of fairness and integrity in comparative analysis, we employed the original source code provided by the authors of the aforementioned models.
In regard to hyperparameter tuning, we undertook a systematic approach for each model and dataset. This was facilitated by the employment of the Python library \texttt{ray}\footnote{\url{https://github.com/ray-project/ray}}~\citep{moritz_ray_2018,liaw_tune_2018}. 
The hyperparameters were calibrated based on the original dataset, and the same configuration was consistently applied to scenarios incorporating missing data, ensuring consistency in experimental settings.

For RNN-based methods, these hyperparameters are used in the fully-connected layer, while for NDE-based methods, they are employed in the embedding layer and vector fields. Furthermore, Euler or Euler--Maruyama method is used to solve NDE-based methods.
In case of the Neural Flow and the proposed FlowPath, three different flow models are considered with original source code~\footnote{\url{https://github.com/mbilos/neural-flows-experiments}}. 
As suggested in \citet{bilos_neural_2021}, we optimize the best flow model for each dataset. Also, we compare the performance of the different flow models in the ablation study. 
Each model undergoes training for a total of 100 epochs, with the best-performing model selected based on the lowest validation loss. The early-stopping is applied when the loss does not improve for 10 consecutive epochs.

\subsection{Comparative Results with different domains}
To ensure a comprehensive evaluation of our approach and to facilitate fair comparison with existing methods, we reproduce and extend the full benchmark suite with 18 datasets in three domains. This includes conducting experiments across all three major domains—Motion \& Human Activity Recognition (HAR), Electrocardiogram (ECG) \& Electroencephalogram (EEG), and various real-world datasets—spanning a total of 18 publicly available datasets. 
By aligning with this established benchmark, we aim to demonstrate the generalizability and robustness of our method across heterogeneous time series modalities.

%%%
%%%
\begin{table*}[!htb]
\scriptsize\centering\captionsetup{skip=5pt}
% \caption{Classification accuracy across all three domains, and the average results for all benchmarks and ablations.
% }\label{tab:result_all_0}
\begin{tabular}{@{}lcccccccccc@{}}
\toprule
\multirow{2.5}{*}{\textbf{Methods}} & \multicolumn{2}{c}{\textbf{Regular}}         & \multicolumn{2}{c}{\textbf{30\% Missing}}    & \multicolumn{2}{c}{\textbf{50\% Missing}}    & \multicolumn{2}{c}{\textbf{70\% Missing}}    & \multicolumn{2}{c}{\textbf{Average}}         \\ \cmidrule(lr){2-3}\cmidrule(lr){4-5}\cmidrule(lr){6-7}\cmidrule(lr){8-9}\cmidrule(lr){10-11} % \cmidrule(l){2-11} 
                                  & \textbf{Accuracy}            & \textbf{Rank} & \textbf{Accuracy}            & \textbf{Rank} & \textbf{Accuracy}            & \textbf{Rank} & \textbf{Accuracy}            & \textbf{Rank} & \textbf{Accuracy}            & \textbf{Rank} \\ \midrule
RNN                               & 0.560\avgstd{0.072}          & 10.7          & 0.484\avgstd{0.075}          & 13.3          & 0.471\avgstd{0.082}          & 12.9          & 0.453\avgstd{0.068}          & 13.3          & 0.492\avgstd{0.074}          & 12.6          \\
LSTM                              & 0.588\avgstd{0.067}          & 10.0          & 0.552\avgstd{0.075}          & 9.4           & 0.516\avgstd{0.073}          & 10.5          & 0.505\avgstd{0.067}          & 10.6          & 0.540\avgstd{0.071}          & 10.1          \\
GRU                               & 0.674\avgstd{0.080}          & 6.9           & 0.639\avgstd{0.065}          & 8.0           & 0.611\avgstd{0.076}          & 8.2           & 0.606\avgstd{0.088}          & 8.1           & 0.633\avgstd{0.077}          & 7.8           \\
GRU-$\Delta t$                             & 0.629\avgstd{0.065}          & 9.1           & 0.636\avgstd{0.069}          & 7.6           & 0.651\avgstd{0.068}          & 6.7           & 0.649\avgstd{0.074}          & 7.5           & 0.641\avgstd{0.069}          & 7.7           \\
GRU-D                             & 0.593\avgstd{0.088}          & 10.0          & 0.579\avgstd{0.087}          & 9.8           & 0.580\avgstd{0.075}          & 9.7           & 0.599\avgstd{0.062}          & 9.4           & 0.588\avgstd{0.078}          & 9.7           \\
GRU-ODE                           & 0.663\avgstd{0.072}          & 7.2           & 0.661\avgstd{0.069}          & 6.8           & 0.664\avgstd{0.069}          & 6.5           & 0.659\avgstd{0.081}          & 6.3           & 0.662\avgstd{0.073}          & 6.7           \\
ODE-RNN                           & 0.652\avgstd{0.085}          & 6.8           & 0.632\avgstd{0.076}          & 7.3           & 0.626\avgstd{0.086}          & 7.2           & 0.653\avgstd{0.059}          & 5.8           & 0.641\avgstd{0.076}          & 6.8           \\
ODE-LSTM                          & 0.566\avgstd{0.074}          & 10.4          & 0.518\avgstd{0.069}          & 11.4          & 0.501\avgstd{0.068}          & 12.1          & 0.474\avgstd{0.068}          & 12.3          & 0.515\avgstd{0.070}          & 11.5          \\
Neural CDE                        & 0.681\avgstd{0.073}          & 7.1           & 0.672\avgstd{0.068}          & 7.3           & 0.661\avgstd{0.070}          & 7.0           & 0.652\avgstd{0.091}          & 7.0           & 0.667\avgstd{0.075}          & 7.1           \\
Neural RDE                        & 0.649\avgstd{0.082}          & 7.9           & 0.648\avgstd{0.071}          & 6.9           & 0.633\avgstd{0.078}          & 7.7           & 0.607\avgstd{0.079}          & 8.2           & 0.634\avgstd{0.078}          & 7.7           \\
ANCDE                             & 0.662\avgstd{0.083}          & 7.3           & 0.661\avgstd{0.083}          & 6.8           & 0.639\avgstd{0.080}          & 7.6           & 0.631\avgstd{0.073}          & 7.1           & 0.649\avgstd{0.080}          & 7.2           \\
EXIT                              & 0.595\avgstd{0.087}          & 9.4           & 0.580\avgstd{0.088}          & 9.8           & 0.578\avgstd{0.086}          & 9.5           & 0.564\avgstd{0.072}          & 10.0          & 0.579\avgstd{0.083}          & 9.7           \\
LEAP                              & 0.490\avgstd{0.062}          & 13.1          & 0.459\avgstd{0.070}          & 13.7          & 0.466\avgstd{0.074}          & 12.4          & 0.451\avgstd{0.074}          & 12.7          & 0.466\avgstd{0.070}          & 13.0          \\
DualDynamics                      &  \ul{0.724\avgstd{0.090}}    &  \ul{4.6}     &  \ul{0.720\avgstd{0.088}}    &  \ul{4.9}     &  \ul{0.691\avgstd{0.091}}    &  \ul{4.9}     &  \ul{0.697\avgstd{0.098}}    &  \ul{4.6}     &  \ul{0.708\avgstd{0.092}}    &  \ul{4.8}     \\
Neural Flow                       & 0.530\avgstd{0.069}          & 11.8          & 0.531\avgstd{0.072}          & 9.9           & 0.537\avgstd{0.073}          & 9.4           & 0.535\avgstd{0.082}          & 9.7           & 0.533\avgstd{0.074}          & 10.2          \\ \midrule
\textbf{FlowPath}                 & \textbf{0.731\avgstd{0.083}} & \textbf{3.7}  & \textbf{0.743\avgstd{0.091}} & \textbf{3.1}  & \textbf{0.726\avgstd{0.084}} & \textbf{3.6}  & \textbf{0.718\avgstd{0.090}} & \textbf{3.5}  & \textbf{0.730\avgstd{0.087}} & \textbf{3.5}  \\ \bottomrule
\end{tabular}
\\[0.5em] % [\baselineskip]
\begin{tabular}{@{}lccccc@{}}
\toprule
\textbf{Settings} & \textbf{Regular}    & \textbf{30\% Missing} & \textbf{50\% Missing} & \textbf{70\% Missing} & \textbf{Average}    \\ \midrule
Neural CDE        & 0.681\avgstd{0.073} & 0.672\avgstd{0.068}   & 0.661\avgstd{0.070}   & 0.652\avgstd{0.091}   & 0.667\avgstd{0.075} \\
\quad + MLP       & 0.705\avgstd{0.077} & 0.700\avgstd{0.084}   & 0.695\avgstd{0.091}   & 0.655\avgstd{0.084}   & 0.689\avgstd{0.084} \\ \midrule
FlowPath          & 0.731\avgstd{0.083} & 0.743\avgstd{0.091}   & 0.726\avgstd{0.084}   & 0.718\avgstd{0.090}   & 0.730\avgstd{0.087} \\
\quad with ResNet Flow   & 0.705\avgstd{0.077} & 0.724\avgstd{0.081}   & 0.697\avgstd{0.084}   & 0.693\avgstd{0.070}   & 0.705\avgstd{0.078} \\
\quad with GRU Flow      & 0.715\avgstd{0.087} & 0.713\avgstd{0.076}   & 0.701\avgstd{0.065}   & 0.681\avgstd{0.080}   & 0.702\avgstd{0.077} \\
\quad with Coupling Flow & 0.713\avgstd{0.085} & 0.720\avgstd{0.084}   & 0.708\avgstd{0.084}   & 0.691\avgstd{0.100}   & 0.708\avgstd{0.088} \\ \bottomrule
\end{tabular}
\caption{Classification accuracy across all three domains, and the average results for all benchmarks and ablations.
}\label{tab:result_all_0}
\end{table*}
%%%
%%%

Table~\ref{tab:result_all_0} summarizes performance across all 18 datasets. Table~\ref{tab:result_all_1} reports results on Motion \& HAR datasets, Table~\ref{tab:result_all_2} covers ECG \& EEG datasets, and Table~\ref{tab:result_all_3} presents results for Sensor datasets.
Through these experiments, we perform a comprehensive comparison between FlowPath and several benchmark methods, and further conduct an ablation study to examine the specific impact and contribution of the flow-based modeling component within the overall FlowPath framework.

%%%
%%%
\begin{table*}[!htb]
\scriptsize\centering\captionsetup{skip=5pt}
% \caption{Classification accuracy across `Motion \& HAR' domains, and the average results for all benchmarks and ablations.
% }\label{tab:result_all_1}
\begin{tabular}{@{}lcccccccccc@{}}
\toprule
\multirow{2.5}{*}{\textbf{Methods}} & \multicolumn{2}{c}{\textbf{Regular}}         & \multicolumn{2}{c}{\textbf{30\% Missing}}    & \multicolumn{2}{c}{\textbf{50\% Missing}}    & \multicolumn{2}{c}{\textbf{70\% Missing}}    & \multicolumn{2}{c}{\textbf{Average}}         \\ \cmidrule(lr){2-3}\cmidrule(lr){4-5}\cmidrule(lr){6-7}\cmidrule(lr){8-9}\cmidrule(lr){10-11} % \cmidrule(l){2-11} 
                                  & \textbf{Accuracy}            & \textbf{Rank} & \textbf{Accuracy}            & \textbf{Rank} & \textbf{Accuracy}            & \textbf{Rank} & \textbf{Accuracy}            & \textbf{Rank} & \textbf{Accuracy}            & \textbf{Rank} \\ \midrule
RNN                               & 0.518\avgstd{0.062}          & 11.8          & 0.445\avgstd{0.077}          & 14.3          & 0.427\avgstd{0.061}          & 14.3          & 0.415\avgstd{0.070}          & 13.6          & 0.451\avgstd{0.067}          & 13.5          \\
LSTM                              & 0.568\avgstd{0.089}          & 10.3          & 0.522\avgstd{0.092}          & 10.8          & 0.494\avgstd{0.094}          & 11.1          & 0.467\avgstd{0.090}          & 12.5          & 0.513\avgstd{0.091}          & 11.2          \\
GRU                               & 0.696\avgstd{0.095}          & 7.0           & 0.629\avgstd{0.076}          & 9.6           & 0.612\avgstd{0.089}          & 8.7           & 0.631\avgstd{0.111}          & 7.3           & 0.642\avgstd{0.093}          & 8.1           \\
GRU-$\Delta t$                             & 0.613\avgstd{0.076}          & 8.2           & 0.624\avgstd{0.079}          & 7.0           & 0.663\avgstd{0.064}          & 6.6           & 0.624\avgstd{0.077}          & 9.0           & 0.631\avgstd{0.074}          & 7.7           \\
GRU-D                             & 0.587\avgstd{0.077}          & 10.3          & 0.563\avgstd{0.089}          & 10.4          & 0.540\avgstd{0.056}          & 12.0          & 0.564\avgstd{0.054}          & 10.3          & 0.563\avgstd{0.069}          & 10.8          \\
GRU-ODE                           & 0.662\avgstd{0.081}          & 8.1           & 0.647\avgstd{0.091}          & 8.8           & 0.656\avgstd{0.072}          & 8.2           & 0.657\avgstd{0.084}          & 6.7           & 0.656\avgstd{0.082}          & 7.9           \\
ODE-RNN                           & 0.621\avgstd{0.087}          & 9.0           & 0.603\avgstd{0.078}          & 8.0           & 0.588\avgstd{0.095}          & 8.1           & 0.627\avgstd{0.064}          & 6.9           & 0.610\avgstd{0.081}          & 8.0           \\
ODE-LSTM                          & 0.550\avgstd{0.083}          & 11.5          & 0.507\avgstd{0.080}          & 11.2          & 0.475\avgstd{0.102}          & 12.4          & 0.426\avgstd{0.094}          & 12.3          & 0.489\avgstd{0.089}          & 11.8          \\
Neural CDE                        &  \ul{0.751\avgstd{0.096}}    & 5.3           &  \ul{0.728\avgstd{0.080}}    & 5.4           & 0.717\avgstd{0.100}          & 5.6           &  \ul{0.738\avgstd{0.111}}    & 5.7           &  \ul{0.733\avgstd{0.097}}    & 5.5           \\
Neural RDE                        & 0.702\avgstd{0.091}          & 5.8           & 0.703\avgstd{0.077}          & 5.7           & 0.704\avgstd{0.088}          & 6.0           & 0.665\avgstd{0.095}          & 7.1           & 0.693\avgstd{0.088}          & 6.1           \\
ANCDE                             & 0.710\avgstd{0.068}          &  \ul{5.0}     & 0.716\avgstd{0.075}          &  \ul{4.7}     &  \ul{0.724\avgstd{0.074}}    &  \ul{3.8}     & 0.703\avgstd{0.077}          & 5.0           & 0.713\avgstd{0.074}          &  \ul{4.6}     \\
EXIT                              & 0.548\avgstd{0.086}          & 9.7           & 0.584\avgstd{0.073}          & 8.5           & 0.567\avgstd{0.086}          & 9.5           & 0.534\avgstd{0.097}          & 10.8          & 0.558\avgstd{0.086}          & 9.6           \\
LEAP                              & 0.397\avgstd{0.084}          & 14.2          & 0.374\avgstd{0.086}          & 14.5          & 0.381\avgstd{0.077}          & 12.9          & 0.335\avgstd{0.062}          & 13.5          & 0.372\avgstd{0.077}          & 13.8          \\
DualDynamics                      & 0.704\avgstd{0.082}          & 5.8           & 0.702\avgstd{0.074}          & 5.3           & 0.696\avgstd{0.088}          & 4.8           & 0.723\avgstd{0.123}          &  \ul{3.6}     & 0.706\avgstd{0.092}          & 4.8           \\
Neural Flow                       & 0.498\avgstd{0.082}          & 10.3          & 0.502\avgstd{0.085}          & 10.2          & 0.504\avgstd{0.081}          & 9.7           & 0.516\avgstd{0.093}          & 9.3           & 0.505\avgstd{0.085}          & 9.9           \\ \midrule
\textbf{FlowPath}                 & \textbf{0.770\avgstd{0.079}} & \textbf{3.8}  & \textbf{0.799\avgstd{0.080}} & \textbf{1.8}  & \textbf{0.774\avgstd{0.092}} & \textbf{2.6}  & \textbf{0.782\avgstd{0.094}} & \textbf{2.5}  & \textbf{0.781\avgstd{0.086}} & \textbf{2.6}  \\ \bottomrule
\end{tabular}
\\[0.5em] % [\baselineskip]
\begin{tabular}{@{}lccccc@{}}
\toprule
\textbf{Settings} & \textbf{Regular}    & \textbf{30\% Missing} & \textbf{50\% Missing} & \textbf{70\% Missing} & \textbf{Average}    \\ \midrule
Neural CDE        & 0.751\avgstd{0.096} & 0.728\avgstd{0.080}   & 0.717\avgstd{0.100}   & 0.738\avgstd{0.111}   & 0.733\avgstd{0.097} \\
\quad + MLP       & 0.737\avgstd{0.114} & 0.744\avgstd{0.105}   & 0.749\avgstd{0.110}   & 0.707\avgstd{0.111}   & 0.734\avgstd{0.110} \\ \midrule
FlowPath          & 0.770\avgstd{0.079} & 0.799\avgstd{0.080}   & 0.774\avgstd{0.092}   & 0.782\avgstd{0.094}   & 0.781\avgstd{0.086} \\
\quad with ResNet Flow   & 0.762\avgstd{0.073} & 0.794\avgstd{0.074}   & 0.761\avgstd{0.092}   & 0.768\avgstd{0.078}   & 0.771\avgstd{0.079} \\
\quad with GRU Flow      & 0.775\avgstd{0.080} & 0.776\avgstd{0.081}   & 0.765\avgstd{0.075}   & 0.775\avgstd{0.090}   & 0.773\avgstd{0.082} \\
\quad with Coupling Flow & 0.761\avgstd{0.073} & 0.748\avgstd{0.075}   & 0.758\avgstd{0.068}   & 0.758\avgstd{0.085}   & 0.756\avgstd{0.075} \\ \bottomrule
\end{tabular}
\caption{Classification accuracy across `Motion \& HAR' domains, and the average results for all benchmarks and ablations.
}\label{tab:result_all_1}
\end{table*}
%%%
%%%
\begin{table*}[!htb]
\scriptsize\centering\captionsetup{skip=5pt}
% \caption{Classification accuracy across `ECG \& EEG' domains, and the average results for all benchmarks and ablations.
% }\label{tab:result_all_2}
\begin{tabular}{@{}lcccccccccc@{}}
\toprule
\multirow{2.5}{*}{\textbf{Methods}} & \multicolumn{2}{c}{\textbf{Regular}}         & \multicolumn{2}{c}{\textbf{30\% Missing}}    & \multicolumn{2}{c}{\textbf{50\% Missing}}    & \multicolumn{2}{c}{\textbf{70\% Missing}}    & \multicolumn{2}{c}{\textbf{Average}}         \\ \cmidrule(lr){2-3}\cmidrule(lr){4-5}\cmidrule(lr){6-7}\cmidrule(lr){8-9}\cmidrule(lr){10-11} % \cmidrule(l){2-11} 
                                  & \textbf{Accuracy}            & \textbf{Rank} & \textbf{Accuracy}            & \textbf{Rank} & \textbf{Accuracy}            & \textbf{Rank} & \textbf{Accuracy}            & \textbf{Rank} & \textbf{Accuracy}            & \textbf{Rank} \\ \midrule
RNN                               & 0.604\avgstd{0.075}          & 8.8           & 0.532\avgstd{0.058}          & 12.6          & 0.545\avgstd{0.062}          & 11.3          & 0.506\avgstd{0.056}          & 13.0          & 0.547\avgstd{0.063}          & 11.4          \\
LSTM                              & 0.609\avgstd{0.054}          & 9.8           & 0.592\avgstd{0.066}          & 7.7           & 0.568\avgstd{0.057}          & 8.9           & 0.546\avgstd{0.048}          & 9.2           & 0.579\avgstd{0.056}          & 8.9           \\
GRU                               & 0.692\avgstd{0.073}          & 5.0           &  \ul{0.704\avgstd{0.059}}    &  \ul{5.3}     & 0.658\avgstd{0.068}          & 7.4           & 0.661\avgstd{0.068}          & 6.8           & 0.679\avgstd{0.067}          & 6.1           \\
GRU-$\Delta t$                             & 0.602\avgstd{0.047}          & 11.3          & 0.616\avgstd{0.052}          & 9.6           & 0.624\avgstd{0.062}          & 7.8           & 0.616\avgstd{0.069}          & 9.0           & 0.614\avgstd{0.057}          & 9.4           \\
GRU-D                             & 0.621\avgstd{0.056}          & 8.8           & 0.606\avgstd{0.066}          & 9.8           & 0.618\avgstd{0.055}          & 9.0           & 0.585\avgstd{0.061}          & 10.9          & 0.608\avgstd{0.060}          & 9.6           \\
GRU-ODE                           & 0.650\avgstd{0.057}          & 7.8           & 0.661\avgstd{0.057}          & 6.3           & 0.653\avgstd{0.063}          & 7.3           & 0.633\avgstd{0.080}          & 7.6           & 0.649\avgstd{0.064}          & 7.2           \\
ODE-RNN                           & 0.664\avgstd{0.068}          & 5.6           & 0.650\avgstd{0.050}          & 6.7           & 0.644\avgstd{0.048}          & 8.2           & 0.647\avgstd{0.050}          &  \ul{5.2}     & 0.651\avgstd{0.054}          & 6.4           \\
ODE-LSTM                          & 0.580\avgstd{0.058}          & 8.5           & 0.546\avgstd{0.065}          & 10.9          & 0.550\avgstd{0.054}          & 11.8          & 0.545\avgstd{0.036}          & 11.7          & 0.555\avgstd{0.053}          & 10.7          \\
Neural CDE                        & 0.643\avgstd{0.040}          & 7.9           & 0.632\avgstd{0.038}          & 8.4           & 0.625\avgstd{0.041}          & 8.4           & 0.596\avgstd{0.069}          & 7.8           & 0.624\avgstd{0.047}          & 8.1           \\
Neural RDE                        & 0.591\avgstd{0.065}          & 11.3          & 0.621\avgstd{0.060}          & 7.6           & 0.602\avgstd{0.070}          & 8.8           & 0.577\avgstd{0.073}          & 8.9           & 0.598\avgstd{0.067}          & 9.1           \\
ANCDE                             & 0.617\avgstd{0.075}          & 8.8           & 0.612\avgstd{0.060}          & 9.1           & 0.596\avgstd{0.059}          & 10.2          & 0.590\avgstd{0.072}          & 7.7           & 0.604\avgstd{0.066}          & 8.9           \\
EXIT                              & 0.599\avgstd{0.092}          & 10.2          & 0.552\avgstd{0.084}          & 11.4          & 0.563\avgstd{0.092}          & 10.8          & 0.574\avgstd{0.044}          & 9.8           & 0.572\avgstd{0.078}          & 10.5          \\
LEAP                              & 0.572\avgstd{0.051}          & 11.4          & 0.527\avgstd{0.080}          & 12.9          & 0.546\avgstd{0.093}          & 10.9          & 0.540\avgstd{0.086}          & 11.5          & 0.546\avgstd{0.077}          & 11.7          \\
DualDynamics                      & \textbf{0.708\avgstd{0.119}} &  \ul{4.2}     & \textbf{0.747\avgstd{0.086}} & 5.7           &  \ul{0.703\avgstd{0.086}}    &  \ul{4.8}     &  \ul{0.683\avgstd{0.096}}    & 6.0           & \textbf{0.710\avgstd{0.097}} &  \ul{5.2}     \\
Neural Flow                       & 0.547\avgstd{0.061}          & 13.4          & 0.572\avgstd{0.061}          & 8.4           & 0.586\avgstd{0.080}          & 7.3           & 0.582\avgstd{0.087}          & 7.3           & 0.572\avgstd{0.072}          & 9.1           \\ \midrule
\textbf{FlowPath}                 &  \ul{0.698\avgstd{0.056}}    & \textbf{3.2}  & 0.701\avgstd{0.090}          & \textbf{3.7}  & \textbf{0.711\avgstd{0.076}} & \textbf{3.1}  & \textbf{0.692\avgstd{0.085}} & \textbf{3.7}  &  \ul{0.701\avgstd{0.077}}    & \textbf{3.4}  \\ \bottomrule
\end{tabular}
\\[0.5em] % [\baselineskip]
\begin{tabular}{@{}lccccc@{}}
\toprule
\textbf{Settings} & \textbf{Regular}    & \textbf{30\% Missing} & \textbf{50\% Missing} & \textbf{70\% Missing} & \textbf{Average}    \\ \midrule
Neural CDE        & 0.643\avgstd{0.040} & 0.632\avgstd{0.038}   & 0.625\avgstd{0.041}   & 0.596\avgstd{0.069}   & 0.624\avgstd{0.047} \\
\quad + MLP       & 0.654\avgstd{0.049} & 0.664\avgstd{0.068}   & 0.666\avgstd{0.087}   & 0.631\avgstd{0.045}   & 0.654\avgstd{0.062} \\ \midrule
FlowPath          & 0.698\avgstd{0.056} & 0.701\avgstd{0.090}   & 0.711\avgstd{0.076}   & 0.692\avgstd{0.085}   & 0.701\avgstd{0.077} \\
\quad with ResNet Flow   & 0.679\avgstd{0.054} & 0.688\avgstd{0.068}   & 0.681\avgstd{0.072}   & 0.667\avgstd{0.058}   & 0.679\avgstd{0.063} \\
\quad with GRU Flow      & 0.665\avgstd{0.073} & 0.671\avgstd{0.069}   & 0.675\avgstd{0.052}   & 0.637\avgstd{0.070}   & 0.662\avgstd{0.066} \\
\quad with Coupling Flow & 0.670\avgstd{0.075} & 0.676\avgstd{0.084}   & 0.676\avgstd{0.092}   & 0.668\avgstd{0.111}   & 0.673\avgstd{0.090} \\ \bottomrule
\end{tabular}
\caption{Classification accuracy across `ECG \& EEG' domains, and the average results for all benchmarks and ablations.
}\label{tab:result_all_2}
\end{table*}
%%%
%%%
\begin{table*}[!htb]
\scriptsize\centering\captionsetup{skip=5pt}
% \caption{Classification accuracy across `Sensor' domains, and the average results for all benchmarks and ablations.
% }\label{tab:result_all_3}
\begin{tabular}{@{}lcccccccccc@{}}
\toprule
\multirow{2.5}{*}{\textbf{Methods}} & \multicolumn{2}{c}{\textbf{Regular}}         & \multicolumn{2}{c}{\textbf{30\% Missing}}    & \multicolumn{2}{c}{\textbf{50\% Missing}}    & \multicolumn{2}{c}{\textbf{70\% Missing}}    & \multicolumn{2}{c}{\textbf{Average}}         \\ \cmidrule(lr){2-3}\cmidrule(lr){4-5}\cmidrule(lr){6-7}\cmidrule(lr){8-9}\cmidrule(lr){10-11} % \cmidrule(l){2-11} 
                                  & \textbf{Accuracy}            & \textbf{Rank} & \textbf{Accuracy}            & \textbf{Rank} & \textbf{Accuracy}            & \textbf{Rank} & \textbf{Accuracy}            & \textbf{Rank} & \textbf{Accuracy}            & \textbf{Rank} \\ \midrule
RNN                               & 0.557\avgstd{0.078}          & 11.6          & 0.476\avgstd{0.089}          & 13.1          & 0.441\avgstd{0.123}          & 13.2          & 0.437\avgstd{0.079}          & 13.4          & 0.478\avgstd{0.092}          & 12.8          \\
LSTM                              & 0.588\avgstd{0.059}          & 10.0          & 0.542\avgstd{0.067}          & 9.8           & 0.484\avgstd{0.069}          & 11.5          & 0.503\avgstd{0.065}          & 10.1          & 0.529\avgstd{0.065}          & 10.3          \\
GRU                               & 0.633\avgstd{0.073}          & 8.8           & 0.585\avgstd{0.059}          & 9.2           & 0.563\avgstd{0.072}          & 8.4           & 0.526\avgstd{0.084}          & 10.0          & 0.577\avgstd{0.072}          & 9.1           \\
GRU-$\Delta t$                             & 0.673\avgstd{0.072}          & 7.8           & 0.666\avgstd{0.075}          & 6.1           & 0.668\avgstd{0.080}          & 5.6           & \textbf{0.708\avgstd{0.075}} & 4.5           & 0.679\avgstd{0.076}          & 6.0           \\
GRU-D                             & 0.569\avgstd{0.130}          & 10.8          & 0.569\avgstd{0.105}          & 9.3           & 0.583\avgstd{0.115}          & 8.1           & 0.647\avgstd{0.070}          & 6.8           & 0.592\avgstd{0.105}          & 8.7           \\
GRU-ODE                           & 0.677\avgstd{0.079}          & 5.8           & 0.674\avgstd{0.059}          &  \ul{5.3}     &  \ul{0.684\avgstd{0.072}}    & \textbf{4.0}  &  \ul{0.687\avgstd{0.079}}    & 4.7           & 0.680\avgstd{0.072}          & 4.9           \\
ODE-RNN                           & 0.673\avgstd{0.098}          & 5.8           & 0.642\avgstd{0.101}          & 7.2           & 0.647\avgstd{0.115}          & 5.3           & 0.684\avgstd{0.062}          & 5.3           & 0.662\avgstd{0.094}          & 5.9           \\
ODE-LSTM                          & 0.567\avgstd{0.083}          & 11.1          & 0.502\avgstd{0.061}          & 12.2          & 0.479\avgstd{0.049}          & 12.2          & 0.453\avgstd{0.075}          & 12.9          & 0.500\avgstd{0.067}          & 12.1          \\
Neural CDE                        & 0.650\avgstd{0.082}          & 8.0           & 0.657\avgstd{0.087}          & 7.9           & 0.641\avgstd{0.069}          & 7.1           & 0.623\avgstd{0.094}          & 7.5           & 0.643\avgstd{0.083}          & 7.6           \\
Neural RDE                        & 0.653\avgstd{0.089}          & 6.7           & 0.620\avgstd{0.076}          & 7.3           & 0.594\avgstd{0.077}          & 8.4           & 0.580\avgstd{0.069}          & 8.7           & 0.612\avgstd{0.078}          & 7.8           \\
ANCDE                             & 0.660\avgstd{0.106}          & 8.0           & 0.656\avgstd{0.115}          & 6.7           & 0.598\avgstd{0.108}          & 9.0           & 0.601\avgstd{0.069}          & 8.7           & 0.629\avgstd{0.100}          & 8.1           \\
EXIT                              & 0.637\avgstd{0.082}          & 8.3           & 0.605\avgstd{0.107}          & 9.5           & 0.606\avgstd{0.081}          & 8.3           & 0.584\avgstd{0.074}          & 9.3           & 0.608\avgstd{0.086}          & 8.9           \\
LEAP                              & 0.500\avgstd{0.052}          & 13.8          & 0.475\avgstd{0.044}          & 13.7          & 0.471\avgstd{0.054}          & 13.5          & 0.478\avgstd{0.073}          & 13.0          & 0.481\avgstd{0.056}          & 13.5          \\
DualDynamics                      & \textbf{0.760\avgstd{0.070}} & \textbf{4.0}  &  \ul{0.712\avgstd{0.106}}    & \textbf{3.9}  & 0.673\avgstd{0.100}          &  \ul{5.0}     & \ul{0.687\avgstd{0.076}}          & \textbf{4.3}  & \textbf{0.708\avgstd{0.088}} & \textbf{4.3}  \\
Neural Flow                       & 0.546\avgstd{0.065}          & 11.5          & 0.520\avgstd{0.071}          & 11.2          & 0.521\avgstd{0.058}          & 11.3          & 0.507\avgstd{0.068}          & 12.4          & 0.524\avgstd{0.065}          & 11.6          \\ \midrule
\textbf{FlowPath}                 &  \ul{0.726\avgstd{0.116}}    &  \ul{4.2}     & \textbf{0.729\avgstd{0.103}} & \textbf{3.9}  & \textbf{0.693\avgstd{0.085}} & 5.2           & 0.679\avgstd{0.092}          &  \ul{4.4}     &  \ul{0.707\avgstd{0.099}}    &  \ul{4.4}     \\ \bottomrule
\end{tabular}
\\[0.5em] % [\baselineskip]
\begin{tabular}{@{}lccccc@{}}
\toprule
\textbf{Settings} & \textbf{Regular}    & \textbf{30\% Missing} & \textbf{50\% Missing} & \textbf{70\% Missing} & \textbf{Average}    \\ \midrule
Neural CDE        & 0.650\avgstd{0.082} & 0.657\avgstd{0.087}   & 0.641\avgstd{0.069}   & 0.623\avgstd{0.094}   & 0.643\avgstd{0.083} \\
\quad + MLP       & 0.723\avgstd{0.069} & 0.691\avgstd{0.079}   & 0.668\avgstd{0.076}   & 0.626\avgstd{0.097}   & 0.677\avgstd{0.080} \\ \midrule
FlowPath          & 0.726\avgstd{0.116} & 0.729\avgstd{0.103}   & 0.693\avgstd{0.085}   & 0.679\avgstd{0.092}   & 0.707\avgstd{0.099} \\
\quad with ResNet Flow   & 0.675\avgstd{0.104} & 0.690\avgstd{0.100}   & 0.647\avgstd{0.087}   & 0.646\avgstd{0.075}   & 0.664\avgstd{0.091} \\
\quad with GRU Flow      & 0.704\avgstd{0.107} & 0.692\avgstd{0.077}   & 0.664\avgstd{0.069}   & 0.631\avgstd{0.078}   & 0.673\avgstd{0.083} \\
\quad with Coupling Flow & 0.707\avgstd{0.108} & 0.736\avgstd{0.092}   & 0.690\avgstd{0.094}   & 0.647\avgstd{0.105}   & 0.695\avgstd{0.100} \\ \bottomrule
\end{tabular}
\caption{Classification accuracy across `Sensor' domains, and the average results for all benchmarks and ablations.
}\label{tab:result_all_3}
\end{table*}
%%%
%%%

\paragraph{Pairwise Statistical Test.}
We perform pairwise comparisons between FlowPath and each baseline model across four missingness levels, as well as an aggregated setting that includes all levels. 
Statistical tests use one-sided Wilcoxon signed-rank tests \citep{demsar_statistical_2006}, with Holm-Bonferroni adjustment for multiple comparisons \citep{giacalone_bonferroni-holm_2018}. Significance is determined at the adjusted threshold of $p<0.05$ \citep{benavoli_should_2016}.

%%%
\begin{table}[H]
\scriptsize\centering\captionsetup{skip=5pt}
% \caption{Pairwise comparison of FlowPath against all baselines across missing rates using Wilcoxon signed-rank tests
% % (Each cell shows FlowPath's Win/Tie/Loss counts across 18 datasets, averaged over 5 runs per dataset. (*) denotes statistically significant superiority of FlowPath) %, $p < 0.05$, one-sided Wilcoxon test).
% }\label{tab:result_analysis}
\begin{tabular}{@{}lccccc@{}}
\toprule
% \textbf{Methods}      & \textbf{Regular} & \textbf{\begin{tabular}[c]{@{}c@{}}30\% \\ Missing\end{tabular}} & \textbf{\begin{tabular}[c]{@{}c@{}}50\% \\ Missing\end{tabular}} & \textbf{\begin{tabular}[c]{@{}c@{}}70\% \\ Missing\end{tabular}} & \textbf{\begin{tabular}[c]{@{}c@{}}All \\ Settings\end{tabular}} \\ \midrule
\textbf{Methods} & \textbf{Regular} & \textbf{30\% Missing} & \textbf{50\% Missing} & \textbf{70\% Missing} & \textbf{All Settings} \\ \midrule
RNN                   & {16/0/2}(*)       & {18/0/0}(*) & {18/0/0}(*) & {18/0/0}(*) & {70/0/2}(*)  \\
LSTM                  & {16/0/2}(*)       & {17/0/1}(*) & {17/0/1}(*) & {18/0/0}(*) & {68/0/4}(*)  \\
GRU                   & {12/0/6}          & {14/0/4}(*) & {13/0/5}(*) & {14/0/4}    & {53/0/19}(*) \\
GRU-$\Delta t$                 & {13/0/5}          & {13/2/3}(*) & {12/1/5}    & {11/2/5}    & {49/5/18}(*) \\
GRU-D                 & {16/0/2}(*)       & {16/1/1}(*) & {15/0/3}(*) & {13/1/4}(*) & {60/2/10}(*) \\
GRU-ODE               & {13/1/4}(*)       & {12/0/6}(*) & {12/1/5}    & {11/0/7}    & {48/2/22}(*) \\
ODE-RNN               & {15/0/3}          & {13/1/4}(*) & {12/0/6}    & {10/0/8}    & {50/1/21}(*) \\
ODE-LSTM              & {17/0/1}(*)       & {17/1/0}(*) & {18/0/0}(*) & {17/0/1}(*) & {69/1/2}(*)  \\
Neural CDE            & {12/1/5}          & {16/1/1}(*) & {13/1/4}(*) & {16/0/2}(*) & {57/3/12}(*) \\
Neural RDE            & {15/0/3}(*)       & {15/0/3}(*) & {14/1/3}(*) & {17/0/1}(*) & {61/1/10}(*) \\
ANCDE                 & {13/2/3}(*)       & {15/1/2}(*) & {15/2/1}(*) & {17/0/1}(*) & {60/5/7}(*)  \\
EXIT                  & {18/0/0}(*)       & {17/0/1}(*) & {16/0/2}(*) & {17/0/1}(*) & {68/0/4}(*)  \\
LEAP                  & {18/0/0}(*)       & {18/0/0}(*) & {17/0/1}(*) & {17/0/1}(*) & {70/0/2}(*)  \\
DualDynamics          & {8/2/8}           & {11/1/6}    & {11/2/5}    & {10/1/7}    & {40/6/26}(*) \\
Neural Flow           & {16/1/1}(*)       & {16/0/2}(*) & {16/0/2}(*) & {16/1/1}(*) & {64/2/6}(*) \\ \bottomrule
\end{tabular}
\caption{Pairwise comparison of FlowPath against all baselines across missing rates using Wilcoxon signed-rank tests
% (Each cell shows FlowPath's Win/Tie/Loss counts across 18 datasets, averaged over 5 runs per dataset. (*) denotes statistically significant superiority of FlowPath) %, $p < 0.05$, one-sided Wilcoxon test).
}\label{tab:result_analysis}
\end{table}
%%%

In Table~\ref{tab:result_analysis}, we report the number of datasets where FlowPath significantly outperforms (Wins), performs comparably (Ties), or underperforms (Losses) relative to the baseline, based on average results over five independent runs. 
Across all conditions, FlowPath demonstrates consistent advantages, with statistically significant improvements in most cases. 
These trends are maintained even under severe levels of missing data, suggesting that FlowPath effectively captures the underlying structure of partially observed time series. 
The overall results highlight the robustness and generalization benefits introduced by the proposed invertible manifold learning framework.

\subsection{Comprehensive analysis of FlowPath}
\begin{figure}[htbp]
\centering\captionsetup{skip=5pt}
\captionsetup[subfigure]{justification=centering, skip=5pt}
    \subfloat[Irregularly-sampled time series (ISTS)]{
      \includegraphics[width=0.32\linewidth]{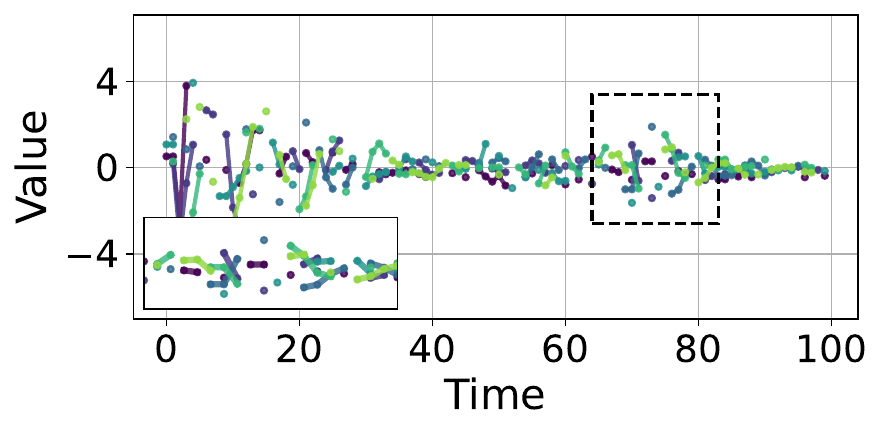}} \hfil
    \subfloat[MLP-based non-invertible path]{
      \includegraphics[width=0.32\linewidth]{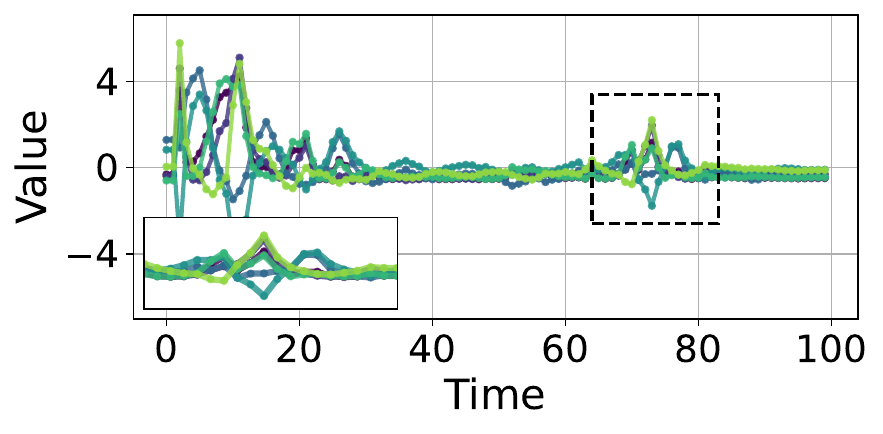}} \hfil
    \subfloat[Flow-based invertible path]{
      \includegraphics[width=0.32\linewidth]{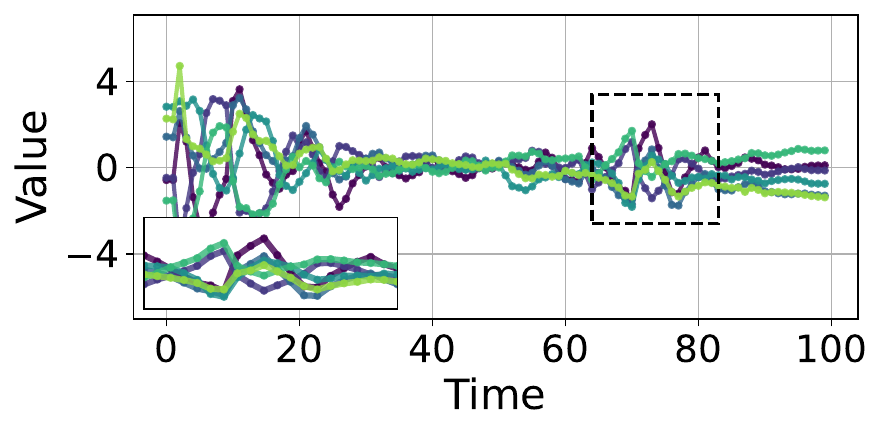}}
\caption{Continuous-time modeling through FlowPath (`BasicMotions' dataset, % with 50\% missing rate, 
instance 1)}
\label{fig:flow_ex1}
% \end{figure}
\medskip
% \begin{figure}[htbp]
\centering\captionsetup{skip=5pt}
\captionsetup[subfigure]{justification=centering, skip=5pt}
    \subfloat[Irregularly-sampled time series (ISTS)]{
      \includegraphics[width=0.32\linewidth]{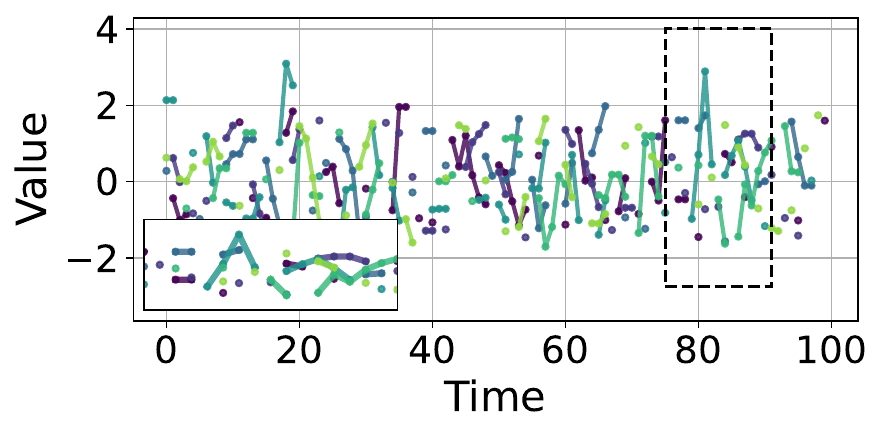}} \hfil
    \subfloat[MLP-based non-invertible path]{
      \includegraphics[width=0.32\linewidth]{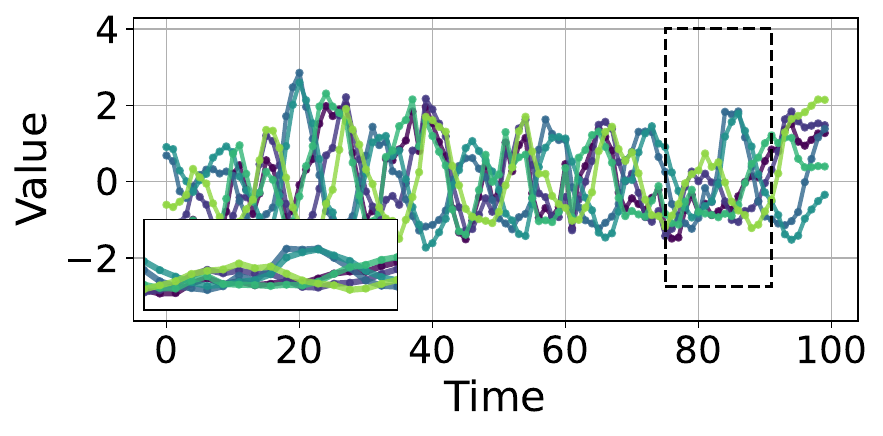}} \hfil
    \subfloat[Flow-based invertible path]{
      \includegraphics[width=0.32\linewidth]{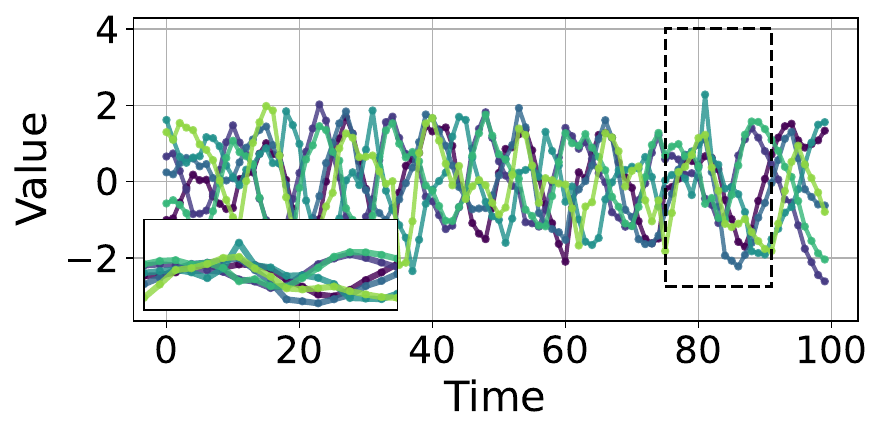}}
\caption{Continuous-time modeling through FlowPath (`BasicMotions' dataset, % with 50\% missing rate, 
instance 2)}
\label{fig:flow_ex2}
% \end{figure}
\medskip
% \begin{figure}[htbp]
\centering\captionsetup{skip=5pt}
\captionsetup[subfigure]{justification=centering, skip=5pt}
    \subfloat[Irregularly-sampled time series (ISTS)]{
      \includegraphics[width=0.32\linewidth]{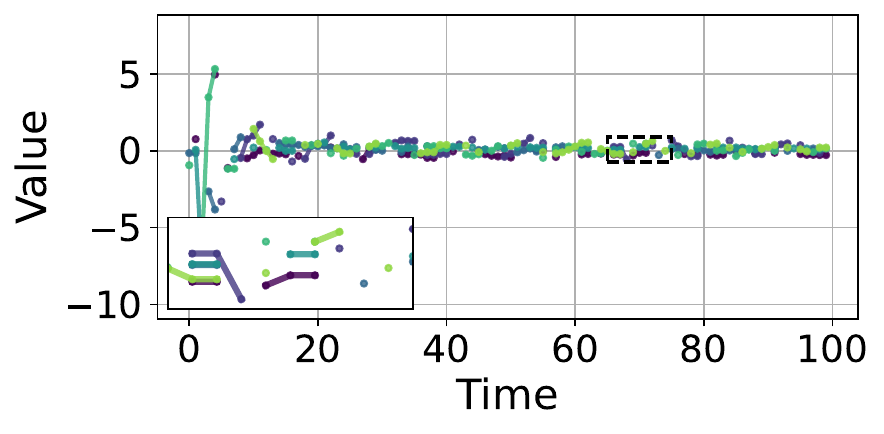}} \hfil
    \subfloat[MLP-based non-invertible path]{
      \includegraphics[width=0.32\linewidth]{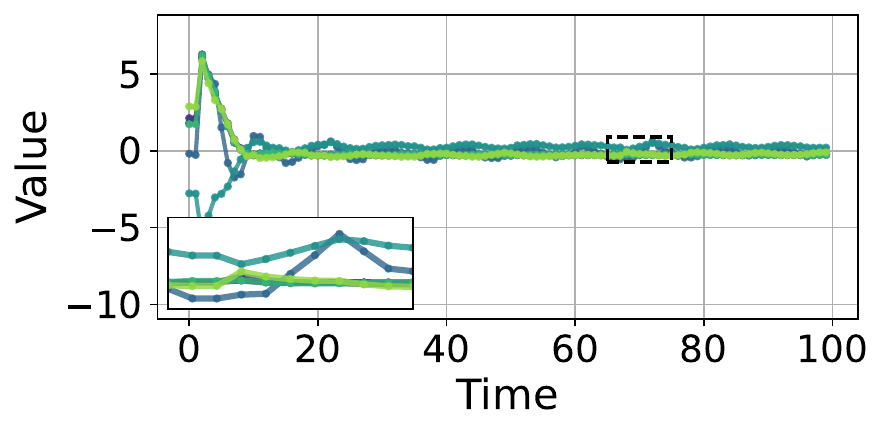}} \hfil
    \subfloat[Flow-based invertible path]{
      \includegraphics[width=0.32\linewidth]{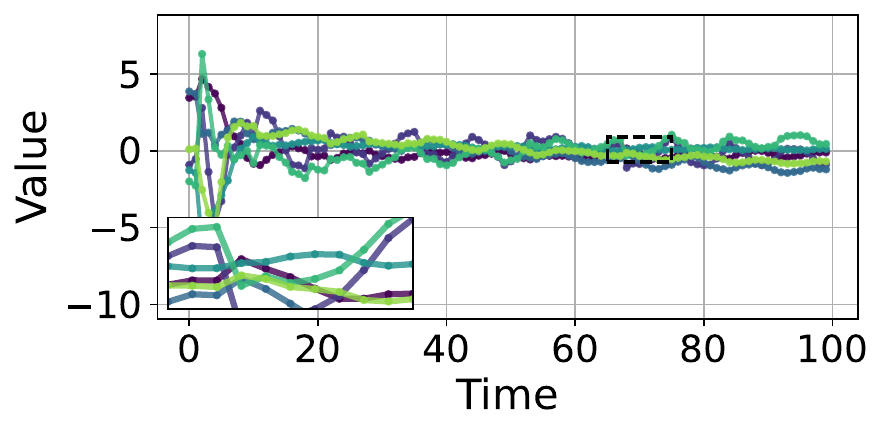}}
\caption{Continuous-time modeling through FlowPath (`BasicMotions' dataset, % with 50\% missing rate, 
instance 3)}
\label{fig:flow_ex3}
% \end{figure}
\medskip
% \begin{figure}[htbp]
\centering\captionsetup{skip=5pt}
\captionsetup[subfigure]{justification=centering, skip=5pt}
    \subfloat[Irregularly-sampled time series (ISTS)]{
      \includegraphics[width=0.32\linewidth]{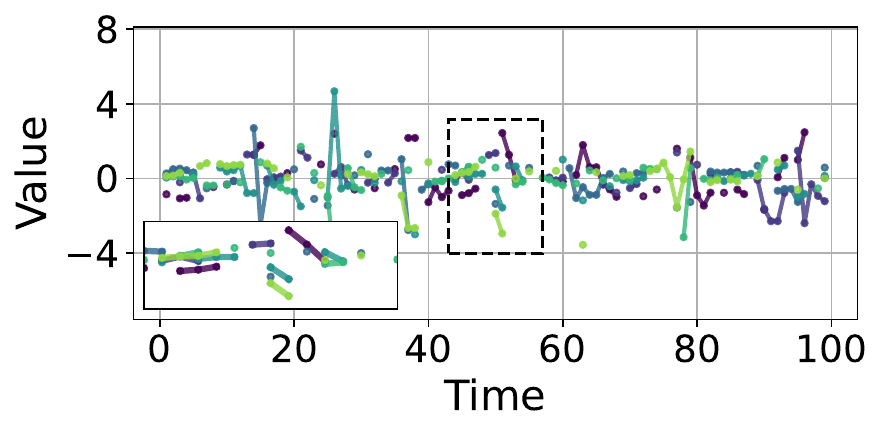}} \hfil
    \subfloat[MLP-based non-invertible path]{
      \includegraphics[width=0.32\linewidth]{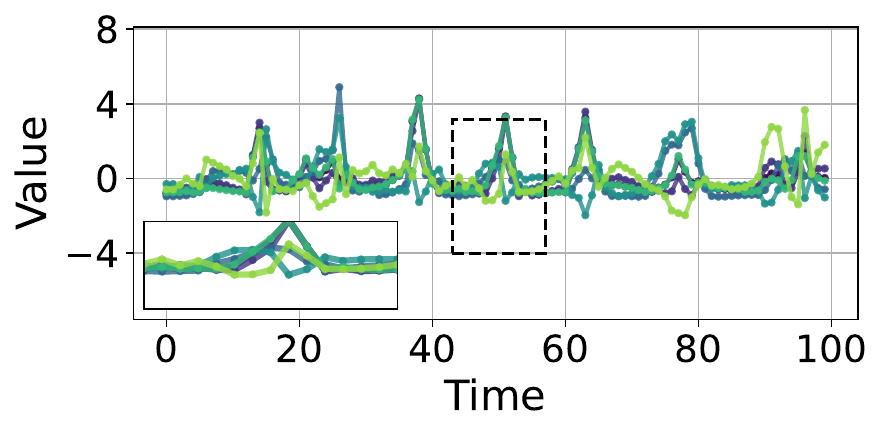}} \hfil
    \subfloat[Flow-based invertible path]{
      \includegraphics[width=0.32\linewidth]{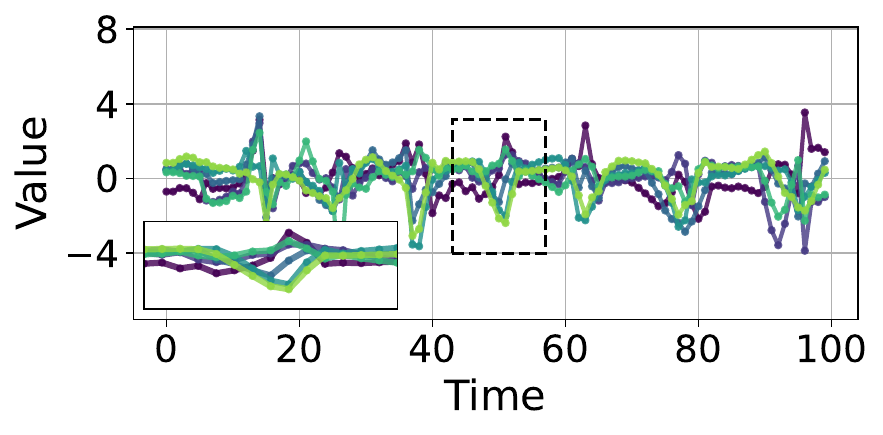}}
\caption{Continuous-time modeling through FlowPath (`BasicMotions' dataset, % with 50\% missing rate, 
instance 4)}
\label{fig:flow_ex4}
\end{figure}

\paragraph{Showcase of FlowPath's Learned Path.}
Figure~\ref{fig:flow_ex1}, \ref{fig:flow_ex2}, \ref{fig:flow_ex3}, and \ref{fig:flow_ex4}  present multiple examples, which are randomly chosen from the test set, demonstrating the effectiveness of FlowPath in handling missing data. Our goal is not reconstructing the missing values but learning a more meaningful representation that benefits downstream tasks, particularly time series classification. 

Each row corresponds to a different instance from the `BasicMotions' dataset with a 50\% missing rate. 
Given the highly irregular raw observations (a), a standard non-invertible MLP (b) learns a chaotic and unstructured path, suggesting it has overfit to the sparse points without capturing the true underlying dynamics. 
In contrast, the path learned by FlowPath (c) is visibly more stable and well-structured, reflecting the inductive bias of the invertible flow toward dynamically plausible manifolds.

\begin{figure}[htbp]
\centering\captionsetup{skip=5pt}
  \includegraphics[width=\linewidth]{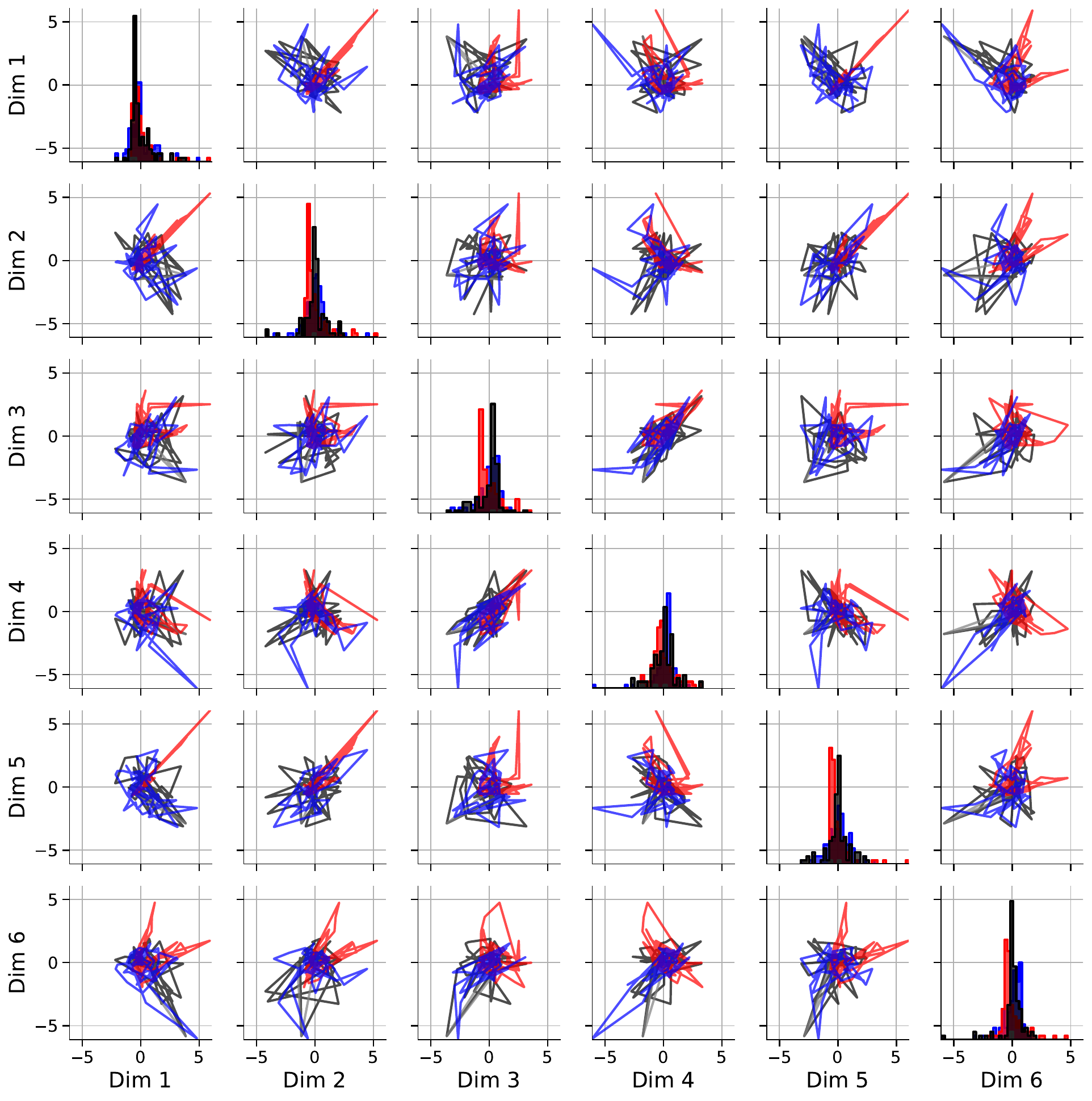}
\caption{
Learned manifolds visualized as 2D trajectories. Off-diagonal plots show phase-space paths, while diagonal plots display marginal histograms. The missing observation (gray) lacks dynamical information due to irregular sampling. The MLP path (red) deviates from the original geometry (black) and exhibits erratic behavior, suggesting overfitting to sparse inputs. In contrast, FlowPath (blue) more consistently captures the global structure and density of the original path. These differences highlight the role of invertibility in promoting stable and coherent representations under data sparsity.
}\label{fig:traj_all}
\end{figure}

\begin{figure}[htbp]
\centering\captionsetup{skip=5pt}
  \includegraphics[width=\linewidth]{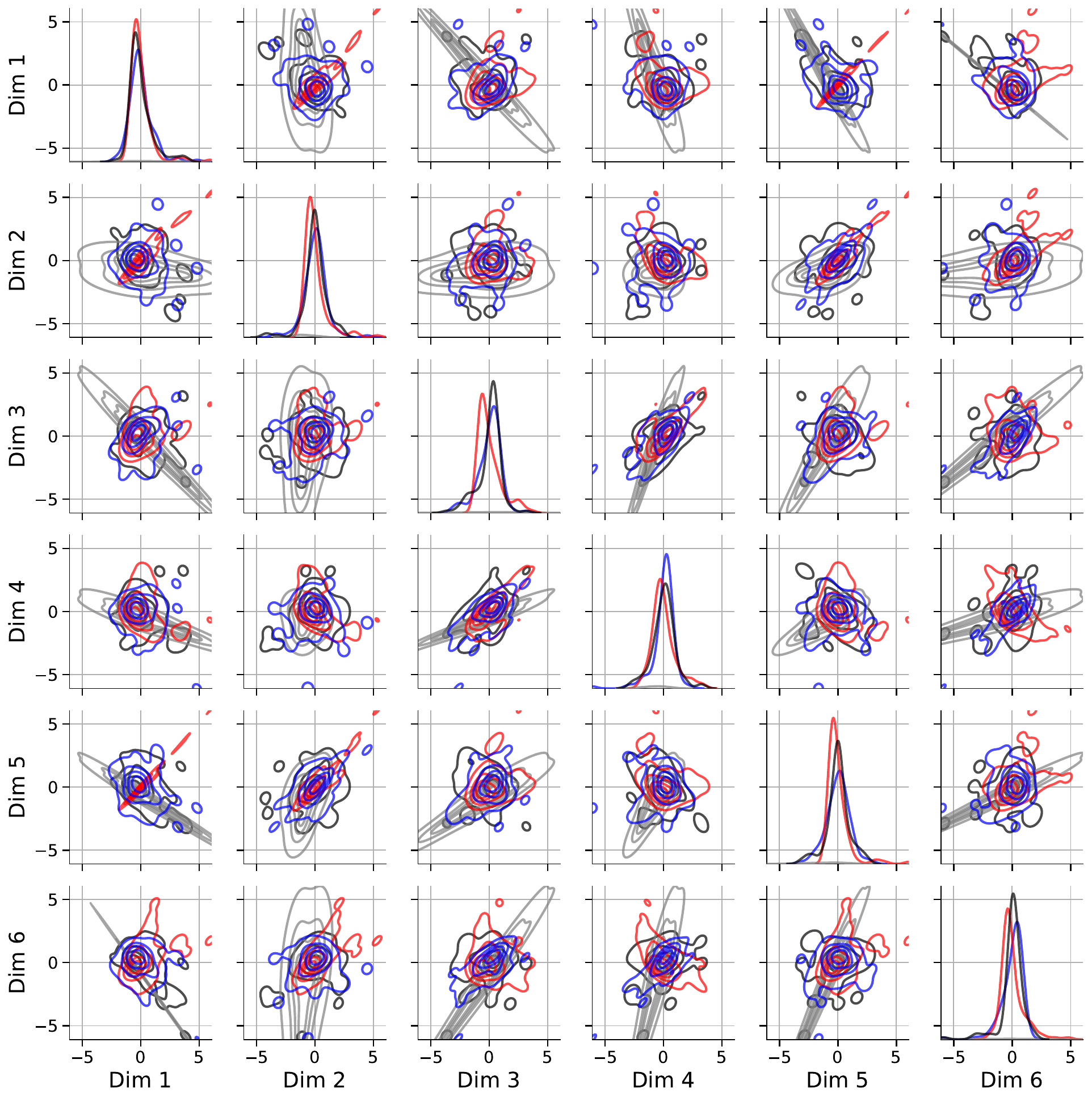}
\caption{
1D and 2D kernel density estimates of the learned manifolds.
FlowPath (blue) shows greater overlap with the original distribution (black) than the sparse observations (gray) and the MLP (red). The sparse observations are diffuse, and the MLP yields a distribution that diverges from the ground truth. In contrast, FlowPath exhibits closer alignment with the target density across projection spaces. While this does not imply full recovery of the underlying distribution, it indicates that FlowPath better approximates its underlying structure under sparse sampling.
}\label{fig:kde_all}
\end{figure}

\paragraph{Detailed Qualitative Visualization of Learned Manifolds.}
To assess the structure of the learned representations, we analyze the trajectories and distributions on the `BasicMotions' dataset with $50\%$ missingness. Figure~\ref{fig:traj_all} visualizes 2D trajectories to assess geometric alignment, while Figure~\ref{fig:kde_all} shows 1D and 2D kernel density estimates to compare distributional overlap.

\clearpage
\newpage
\section{Detailed results on real-world dataset with sensor dropout}\label{appendix:result_pam}
The PAMAP2 dataset~\citep{a_reiss_introducing_2012}\footnote{\url{https://archive.ics.uci.edu/dataset/231/pamap2+physical+activity+monitoring}} contains daily activity data from nine individuals, recorded using three inertial measurement units (placed on the wrist, chest, and ankle) and a heart rate monitor.
To evaluate model robustness under sensor dropout scenarios, we adapted this dataset following the preprocessing method proposed by \citet{zhang_graph-guided_2022}\footnote{\url{https://github.com/mims-harvard/Raindrop}}.

%%%
\begin{table}[htbp]
\scriptsize\centering\captionsetup{skip=5pt}
\vspace{-1.0em}
% \caption{Results of hyperparameter tuning for `PAMAP2' with original irregular scenario}\label{tab:hyperparameter_pam}
\subfloat[Accuracy]{%
\begin{tabular}{@{}ccccccc@{}}
\toprule
\multirow{2.5}{*}{\textbf{$n_l$}} & \multirow{2.5}{*}{\textbf{$n_h$}} & \multicolumn{2}{c}{\textbf{Neural CDE}} & \multicolumn{3}{c}{\textbf{FlowPath}}                \\ \cmidrule(lr){3-4} \cmidrule(lr){5-7} 
                                &                                 & \textbf{Spline}     & \textbf{MLP}      & \textbf{ResNet} & \textbf{GRU}   & \textbf{Coupling} \\ \midrule
\multirow{4}{*}{\textbf{1}}     & \textbf{16}                     & 35.9 \std{0.6}      & 83.0 \std{1.0}    & 56.6 \std{5.1}  & 49.3 \std{3.0} & 42.0 \std{1.2}    \\
                                & \textbf{32}                     & 45.1 \std{2.2}      & 86.0 \std{1.0}    & 57.6 \std{1.8}  & 58.4 \std{3.2} & 59.2 \std{2.3}    \\
                                & \textbf{64}                     & 52.6 \std{3.2}      & 87.8 \std{0.3}    & 68.1 \std{2.4}  & 69.3 \std{3.0} & 66.7 \std{1.6}    \\
                                & \textbf{128}                    & 64.5 \std{3.0}      & 88.0 \std{0.8}    & 78.4 \std{2.7}  & 77.6 \std{1.5} & 76.0 \std{4.5}    \\ \midrule
\multirow{4}{*}{\textbf{2}}     & \textbf{16}                     & 45.1 \std{5.3}      & 81.8 \std{2.4}    & 77.1 \std{1.5}  & 73.7 \std{1.5} & 68.4 \std{0.5}    \\
                                & \textbf{32}                     & 61.1 \std{6.0}      & 86.0 \std{0.3}    & 81.6 \std{3.0}  & 75.5 \std{0.9} & 77.2 \std{1.2}    \\
                                & \textbf{64}                     & 76.1 \std{3.1}      & 87.6 \std{0.3}    & 84.3 \std{0.6}  & 79.7 \std{0.3} & 84.1 \std{0.7}    \\
                                & \textbf{128}                    & 85.2 \std{2.0}      & 92.4 \std{0.7}    & 87.5 \std{0.5}  & 87.2 \std{0.5} & 89.1 \std{0.4}    \\ \midrule
\multirow{4}{*}{\textbf{3}}     & \textbf{16}                     & 79.1 \std{1.9}      & 79.0 \std{1.9}    & 82.0 \std{0.3}  & 80.4 \std{0.4} & 82.1 \std{2.1}    \\
                                & \textbf{32}                     & 83.0 \std{3.8}      & 87.8 \std{0.9}    & 88.6 \std{1.2}  & 88.1 \std{0.7} & 87.1 \std{0.6}    \\
                                & \textbf{64}                     & 88.4 \std{1.5}      & 91.6 \std{0.3}    & 90.9 \std{0.3}  & 90.8 \std{0.6} & 90.8 \std{0.8}    \\
                                & \textbf{128}                    & 91.6 \std{1.3}      & 94.0 \std{1.2}    & 92.4 \std{0.2}  & 93.4 \std{1.2} & 92.8 \std{0.8}    \\ \midrule
\multirow{4}{*}{\textbf{4}}     & \textbf{16}                     & 81.5 \std{2.8}      & 71.5 \std{1.0}    & 83.8 \std{1.1}  & 83.8 \std{0.7} & 84.7 \std{1.2}    \\
                                & \textbf{32}                     & 87.8 \std{0.5}      & 83.7 \std{1.9}    & 89.4 \std{0.4}  & 90.9 \std{0.5} & 90.8 \std{1.0}    \\
                                & \textbf{64}                     & 91.9 \std{0.8}      & 92.9 \std{1.3}    & 93.1 \std{0.2}  & 93.4 \std{1.0} & 92.8 \std{0.3}    \\
                                & \textbf{128}                    & 94.2 \std{0.5}      & 93.3 \std{0.3}    & 94.1 \std{0.4}  & 94.8 \std{0.2} & 94.4 \std{0.1}    \\ \bottomrule
\end{tabular}} \hfil
\subfloat[Precision]{%
\begin{tabular}{@{}ccccccc@{}}
\toprule
\multirow{2.5}{*}{\textbf{$n_l$}} & \multirow{2.5}{*}{\textbf{$n_h$}} & \multicolumn{2}{c}{\textbf{Neural CDE}} & \multicolumn{3}{c}{\textbf{FlowPath}}                \\ \cmidrule(lr){3-4} \cmidrule(lr){5-7}
                                &                                 & \textbf{Spline}     & \textbf{MLP}      & \textbf{ResNet} & \textbf{GRU}   & \textbf{Coupling} \\ \midrule
\multirow{4}{*}{\textbf{1}}     & \textbf{16}                     & 39.7 \std{3.2}      & 84.6 \std{1.1}    & 56.7 \std{6.7}  & 52.5 \std{4.1} & 37.1 \std{12.5}   \\
                                & \textbf{32}                     & 49.2 \std{3.4}      & 87.3 \std{0.5}    & 58.5 \std{1.8}  & 60.5 \std{3.8} & 60.5 \std{1.0}    \\
                                & \textbf{64}                     & 55.4 \std{3.9}      & 89.1 \std{0.8}    & 69.3 \std{1.4}  & 71.8 \std{3.2} & 67.9 \std{2.3}    \\
                                & \textbf{128}                    & 67.8 \std{3.1}      & 90.0 \std{0.9}    & 79.6 \std{2.8}  & 79.0 \std{1.6} & 77.1 \std{4.9}    \\ \midrule
\multirow{4}{*}{\textbf{2}}     & \textbf{16}                     & 54.0 \std{7.4}      & 83.4 \std{2.3}    & 79.3 \std{1.6}  & 77.1 \std{1.9} & 70.2 \std{1.9}    \\
                                & \textbf{32}                     & 64.7 \std{5.6}      & 87.3 \std{0.6}    & 83.3 \std{3.0}  & 76.6 \std{1.0} & 78.9 \std{0.9}    \\
                                & \textbf{64}                     & 78.8 \std{3.1}      & 89.3 \std{0.6}    & 86.0 \std{0.6}  & 81.8 \std{0.7} & 86.8 \std{1.1}    \\
                                & \textbf{128}                    & 87.1 \std{1.8}      & 93.6 \std{0.6}    & 88.9 \std{0.6}  & 88.4 \std{0.3} & 90.5 \std{0.5}    \\ \midrule
\multirow{4}{*}{\textbf{3}}     & \textbf{16}                     & 82.5 \std{1.4}      & 82.1 \std{1.8}    & 83.4 \std{0.4}  & 82.3 \std{0.1} & 84.8 \std{2.2}    \\
                                & \textbf{32}                     & 85.7 \std{2.9}      & 89.8 \std{0.7}    & 90.1 \std{1.1}  & 90.1 \std{0.8} & 89.1 \std{0.2}    \\
                                & \textbf{64}                     & 90.3 \std{1.2}      & 93.2 \std{0.2}    & 92.2 \std{0.5}  & 92.5 \std{0.6} & 92.1 \std{0.3}    \\
                                & \textbf{128}                    & 93.0 \std{0.9}      & 95.2 \std{0.9}    & 93.6 \std{0.4}  & 94.7 \std{0.9} & 93.8 \std{1.1}    \\ \midrule
\multirow{4}{*}{\textbf{4}}     & \textbf{16}                     & 84.9 \std{2.0}      & 74.5 \std{3.0}    & 85.6 \std{1.0}  & 86.1 \std{0.9} & 86.6 \std{1.3}    \\
                                & \textbf{32}                     & 89.5 \std{0.8}      & 85.9 \std{1.4}    & 90.9 \std{0.2}  & 92.6 \std{0.5} & 92.0 \std{1.5}    \\
                                & \textbf{64}                     & 93.3 \std{0.6}      & 93.8 \std{0.8}    & 94.2 \std{0.6}  & 94.4 \std{0.9} & 93.9 \std{0.5}    \\
                                & \textbf{128}                    & 95.2 \std{0.4}      & 94.2 \std{0.2}    & 94.8 \std{0.4}  & 95.8 \std{0.4} & 95.1 \std{0.3}    \\ \bottomrule
\end{tabular}} \\ 
\subfloat[Recall]{%
\begin{tabular}{@{}ccccccc@{}}
\toprule
\multirow{2.5}{*}{\textbf{$n_l$}} & \multirow{2.5}{*}{\textbf{$n_h$}} & \multicolumn{2}{c}{\textbf{Neural CDE}} & \multicolumn{3}{c}{\textbf{FlowPath}}                \\ \cmidrule(lr){3-4} \cmidrule(lr){5-7}
                                &                                 & \textbf{Spline}     & \textbf{MLP}      & \textbf{ResNet} & \textbf{GRU}   & \textbf{Coupling} \\ \midrule
\multirow{4}{*}{\textbf{1}}     & \textbf{16}                     & 33.7 \std{1.7}      & 85.0 \std{1.1}    & 55.3 \std{4.0}  & 48.4 \std{3.7} & 39.6 \std{2.3}    \\
                                & \textbf{32}                     & 44.4 \std{2.5}      & 87.6 \std{1.0}    & 58.0 \std{3.6}  & 60.2 \std{5.1} & 59.9 \std{1.7}    \\
                                & \textbf{64}                     & 53.9 \std{3.8}      & 88.9 \std{0.5}    & 69.0 \std{3.9}  & 69.8 \std{4.2} & 67.6 \std{1.3}    \\
                                & \textbf{128}                    & 66.0 \std{3.0}      & 89.5 \std{0.7}    & 79.8 \std{2.6}  & 79.0 \std{1.7} & 77.7 \std{4.6}    \\ \midrule
\multirow{4}{*}{\textbf{2}}     & \textbf{16}                     & 45.5 \std{6.1}      & 84.2 \std{2.4}    & 80.0 \std{2.0}  & 75.7 \std{1.4} & 69.9 \std{0.7}    \\
                                & \textbf{32}                     & 62.4 \std{7.1}      & 88.3 \std{0.2}    & 83.3 \std{4.1}  & 76.8 \std{1.9} & 78.6 \std{1.5}    \\
                                & \textbf{64}                     & 78.3 \std{3.6}      & 89.1 \std{0.3}    & 86.2 \std{0.9}  & 82.6 \std{0.3} & 85.6 \std{1.2}    \\
                                & \textbf{128}                    & 86.9 \std{1.7}      & 93.0 \std{0.6}    & 88.3 \std{0.8}  & 88.3 \std{0.9} & 89.7 \std{0.7}    \\ \midrule
\multirow{4}{*}{\textbf{3}}     & \textbf{16}                     & 81.3 \std{2.1}      & 81.3 \std{2.6}    & 84.0 \std{0.2}  & 82.0 \std{0.2} & 84.3 \std{2.4}    \\
                                & \textbf{32}                     & 84.8 \std{3.0}      & 89.1 \std{0.8}    & 89.9 \std{0.8}  & 89.0 \std{0.9} & 88.0 \std{0.8}    \\
                                & \textbf{64}                     & 89.4 \std{1.4}      & 92.1 \std{0.2}    & 91.9 \std{0.3}  & 91.8 \std{0.2} & 91.8 \std{0.6}    \\
                                & \textbf{128}                    & 92.3 \std{1.3}      & 94.4 \std{1.0}    & 93.1 \std{0.6}  & 94.3 \std{1.1} & 93.3 \std{0.5}    \\ \midrule
\multirow{4}{*}{\textbf{4}}     & \textbf{16}                     & 83.5 \std{2.6}      & 73.4 \std{2.3}    & 85.1 \std{0.8}  & 85.2 \std{0.6} & 86.3 \std{0.8}    \\
                                & \textbf{32}                     & 89.2 \std{0.5}      & 86.0 \std{2.0}    & 90.5 \std{0.6}  & 91.3 \std{0.5} & 91.7 \std{0.9}    \\
                                & \textbf{64}                     & 92.7 \std{0.6}      & 93.2 \std{1.4}    & 93.6 \std{0.4}  & 93.9 \std{0.6} & 93.3 \std{0.4}    \\
                                & \textbf{128}                    & 94.8 \std{0.5}      & 93.6 \std{0.6}    & 94.4 \std{0.7}  & 95.5 \std{0.2} & 94.6 \std{0.3}    \\ \bottomrule
\end{tabular}} \hfil
\subfloat[F1 Score]{%
\begin{tabular}{@{}ccccccc@{}}
\toprule
\multirow{2.5}{*}{\textbf{$n_l$}} & \multirow{2.5}{*}{\textbf{$n_h$}} & \multicolumn{2}{c}{\textbf{Neural CDE}} & \multicolumn{3}{c}{\textbf{FlowPath}}                \\ \cmidrule(lr){3-4} \cmidrule(lr){5-7}
                                &                                 & \textbf{Spline}     & \textbf{MLP}      & \textbf{ResNet} & \textbf{GRU}   & \textbf{Coupling} \\ \midrule
\multirow{4}{*}{\textbf{1}}     & \textbf{16}                     & 33.6 \std{2.8}      & 84.5 \std{1.0}    & 53.5 \std{6.1}  & 47.0 \std{3.4} & 34.0 \std{6.2}    \\
                                & \textbf{32}                     & 44.8 \std{2.4}      & 87.4 \std{0.6}    & 57.4 \std{2.7}  & 59.9 \std{4.4} & 59.5 \std{1.5}    \\
                                & \textbf{64}                     & 54.0 \std{3.6}      & 88.9 \std{0.5}    & 68.6 \std{2.5}  & 70.3 \std{3.8} & 67.6 \std{1.7}    \\
                                & \textbf{128}                    & 66.5 \std{3.1}      & 89.7 \std{0.5}    & 79.5 \std{2.7}  & 78.8 \std{1.6} & 77.2 \std{4.7}    \\ \midrule
\multirow{4}{*}{\textbf{2}}     & \textbf{16}                     & 46.3 \std{6.5}      & 83.4 \std{2.2}    & 78.9 \std{1.2}  & 75.6 \std{1.1} & 69.3 \std{0.8}    \\
                                & \textbf{32}                     & 62.5 \std{6.9}      & 87.5 \std{0.3}    & 83.1 \std{3.6}  & 76.5 \std{1.3} & 78.5 \std{0.9}    \\
                                & \textbf{64}                     & 78.3 \std{3.3}      & 89.1 \std{0.2}    & 85.9 \std{0.6}  & 82.0 \std{0.5} & 86.1 \std{1.0}    \\
                                & \textbf{128}                    & 86.9 \std{1.7}      & 93.2 \std{0.6}    & 88.5 \std{0.8}  & 88.3 \std{0.6} & 89.9 \std{0.2}    \\ \midrule
\multirow{4}{*}{\textbf{3}}     & \textbf{16}                     & 81.7 \std{1.8}      & 81.3 \std{2.3}    & 83.5 \std{0.2}  & 82.0 \std{0.2} & 84.3 \std{2.2}    \\
                                & \textbf{32}                     & 85.0 \std{3.0}      & 89.4 \std{0.8}    & 90.0 \std{0.9}  & 89.5 \std{0.8} & 88.4 \std{0.5}    \\
                                & \textbf{64}                     & 89.8 \std{1.3}      & 92.6 \std{0.2}    & 91.9 \std{0.3}  & 92.1 \std{0.4} & 91.9 \std{0.3}    \\
                                & \textbf{128}                    & 92.6 \std{1.1}      & 94.8 \std{0.9}    & 93.3 \std{0.5}  & 94.5 \std{0.9} & 93.5 \std{0.8}    \\ \midrule
\multirow{4}{*}{\textbf{4}}     & \textbf{16}                     & 84.0 \std{2.3}      & 71.5 \std{3.3}    & 85.2 \std{0.8}  & 85.5 \std{0.7} & 86.3 \std{1.0}    \\
                                & \textbf{32}                     & 89.2 \std{0.5}      & 85.9 \std{1.8}    & 90.6 \std{0.4}  & 91.8 \std{0.2} & 91.7 \std{1.2}    \\
                                & \textbf{64}                     & 93.0 \std{0.6}      & 93.5 \std{1.1}    & 93.8 \std{0.5}  & 94.1 \std{0.7} & 93.5 \std{0.4}    \\
                                & \textbf{128}                    & 95.0 \std{0.4}      & 93.9 \std{0.4}    & 94.6 \std{0.5}  & 95.6 \std{0.2} & 94.8 \std{0.2}    \\ \bottomrule
\end{tabular}}
% \vspace{-1.0em}
\caption{Results of hyperparameter tuning for `PAMAP2' with original irregular scenario}\label{tab:hyperparameter_pam}
\end{table}
%%%

\subsection{Data Preprocessing.}
Users one through eight were included in the analysis, while the ninth subject was excluded due to insufficient sensor readings. The signal data were subsequently segmented into units using a 600-second window with a 50\% overlap between segments. Initially, the PAMAP2 dataset included 18 daily living activities, but we removed activities represented by fewer than 500 samples, resulting in a dataset with 8 activities. Consequently, the modified PAM dataset comprises 5333 segments of sensor signal data. Each segment is collected from 17 sensors, capturing 600 continuous data points at a 100 Hz sampling rate. To introduce irregularity into the time series, we randomly eliminated 60\% of the data points from each segment, ensuring the selection of removed observations was random yet consistent across all experiments. 

\subsection{Comparison with benchmark methods.}
\begin{figure}[htbp]
\centering\captionsetup{skip=5pt}
\captionsetup[subfigure]{justification=centering, skip=5pt}
    \subfloat[Accuracy]{
      \includegraphics[width=0.48\linewidth]{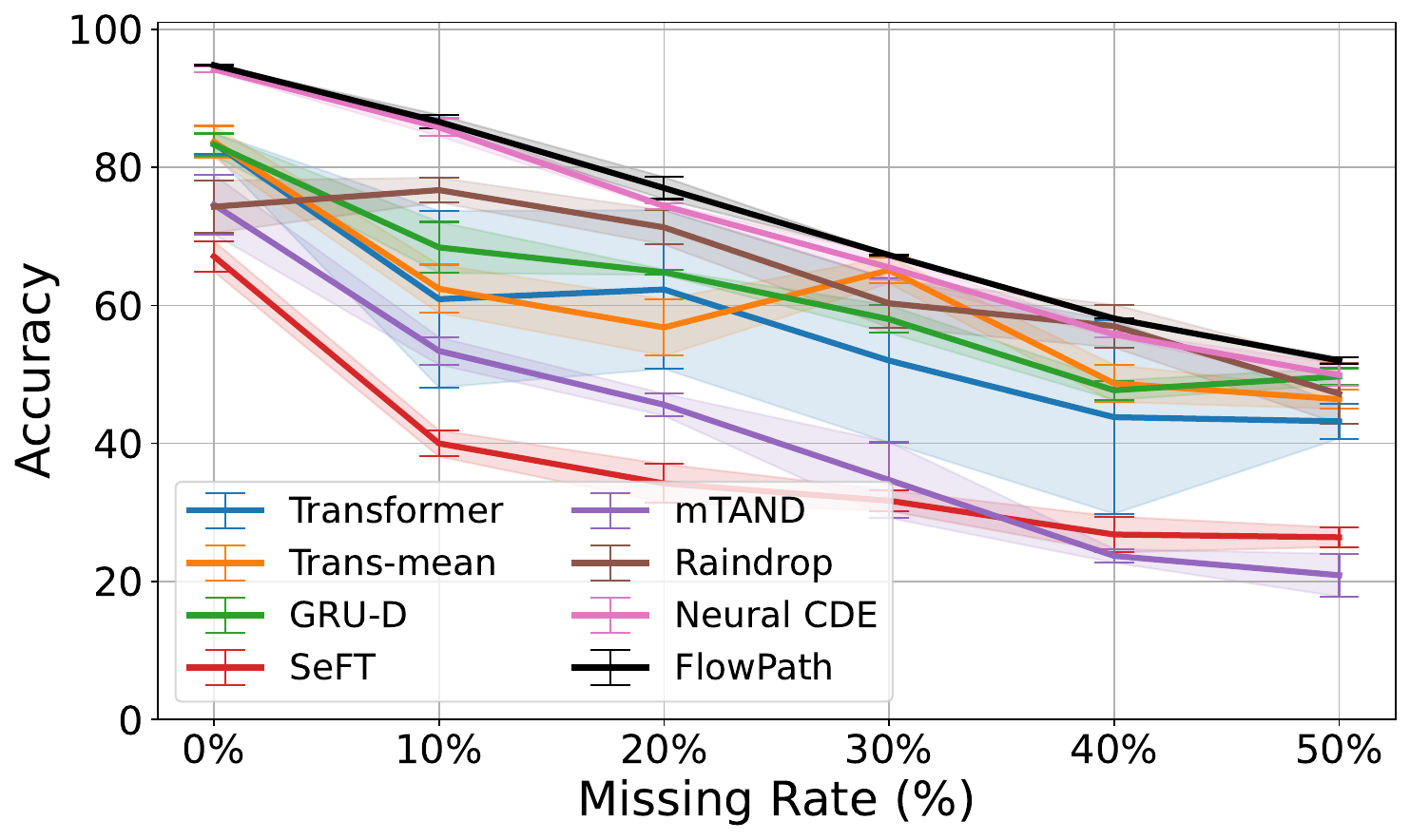}} \hfil
    \subfloat[Precision]{
      \includegraphics[width=0.48\linewidth]{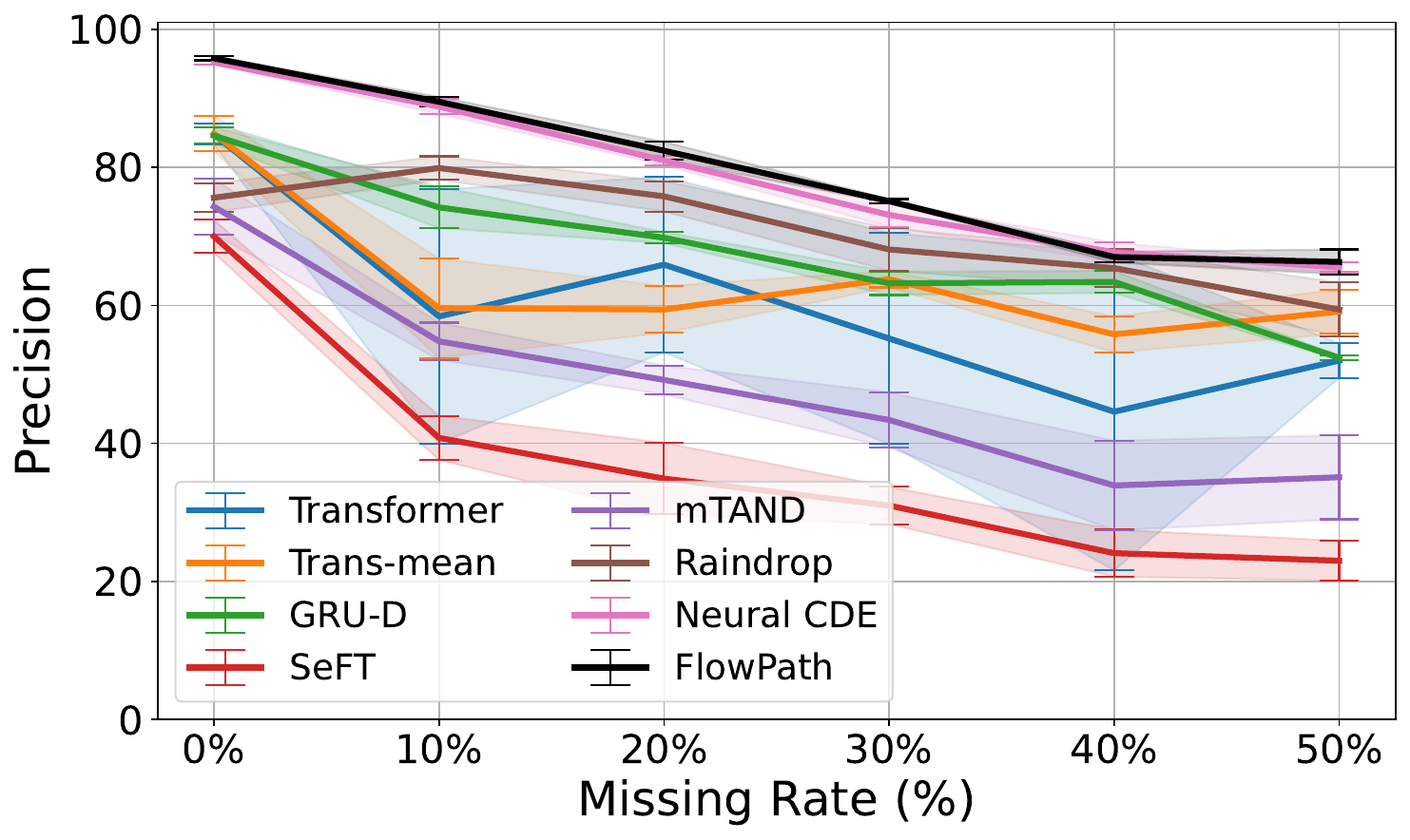}} \\
    \subfloat[Recall]{
      \includegraphics[width=0.48\linewidth]{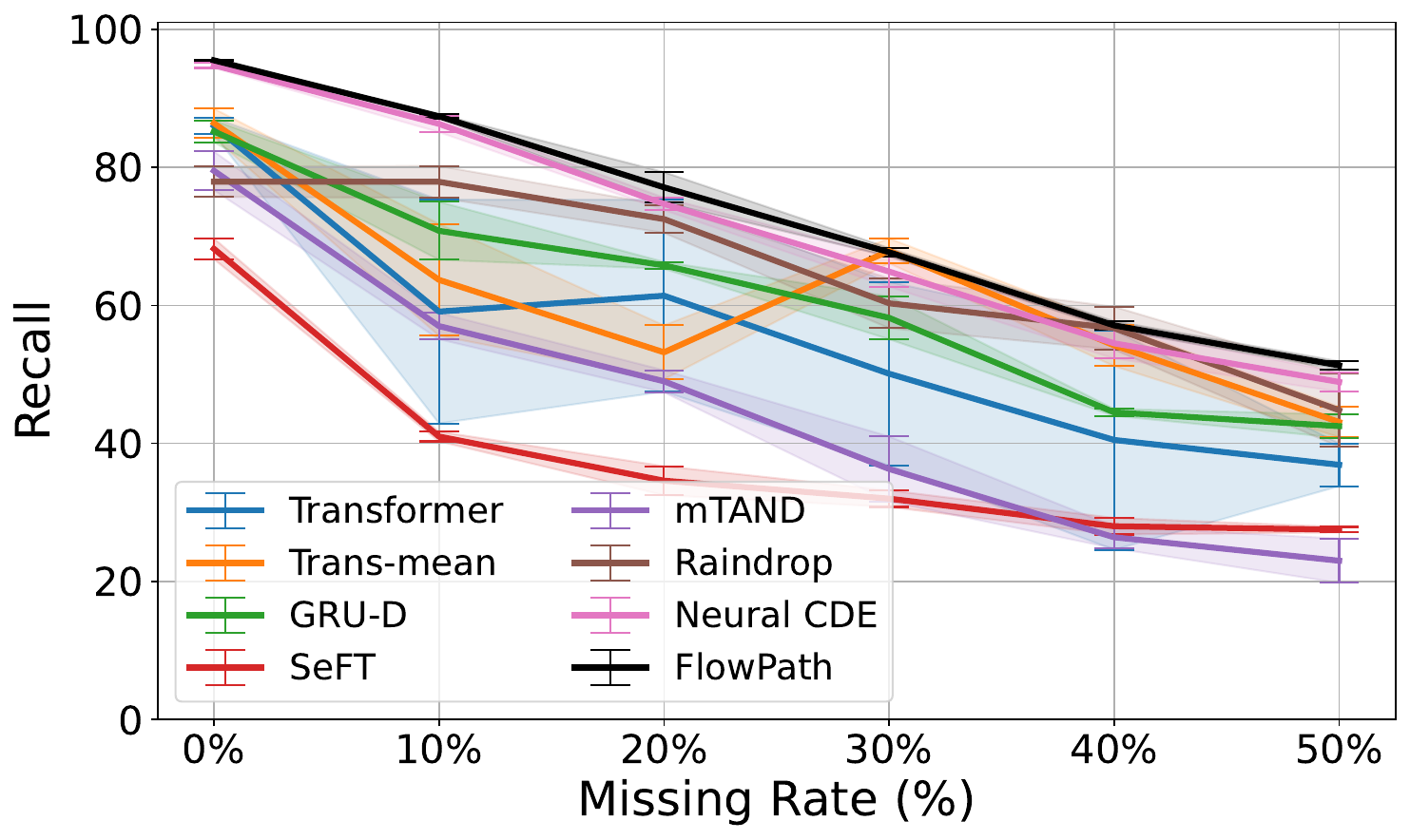}} \hfil
    \subfloat[F1 Score]{
      \includegraphics[width=0.48\linewidth]{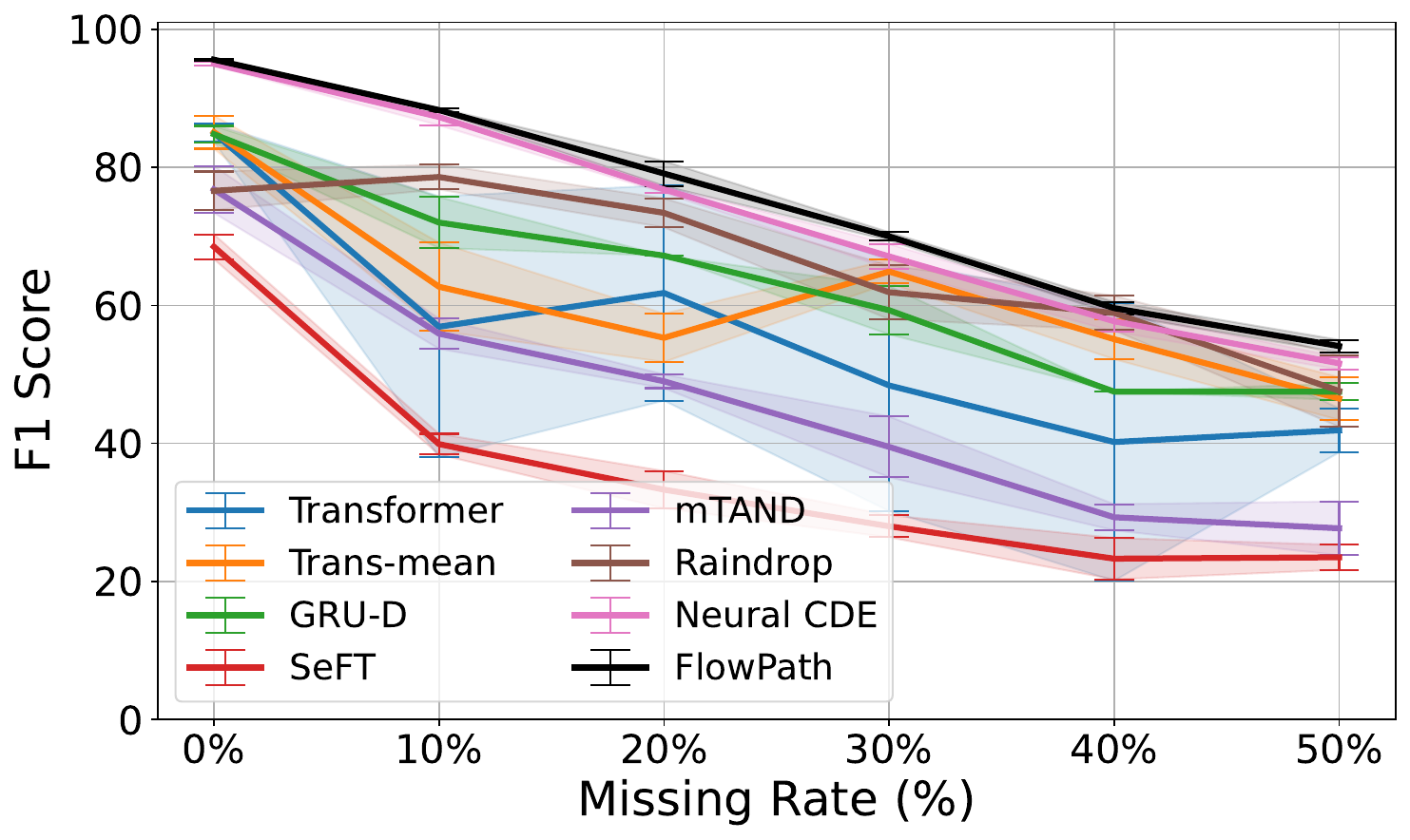}}
\caption{Classification performance of FlowPath compared to baseline models on the PAMAP2 dataset as the percentage of sensor dropout increases. Solid lines indicate the mean performance, and shaded regions represent one standard deviation.
FlowPath exhibits superior and more stable performance across all metrics, demonstrating robustness in handling ISTS.
}
\label{fig:result_pam_all}
\end{figure}

Figure~\ref{fig:result_pam_all} and Table~\ref{tab:result_pam_split} present the performance comparison between the proposed method and baseline models under varying sensor missing rates. FlowPath consistently achieves higher classification accuracy and robustness across all levels of missingness. The performance gap widens as the missing rate increases, indicating that learning a structurally-constrained control path is particularly beneficial when observations are sparse. 

%%%
\begin{table}[!htb]
\scriptsize\centering\captionsetup{skip=5pt}
\vspace{-1.0em}
% \caption{Classification performance on PAMAP2 dataset with original and sensor dropout from 10\% to 50\%}
% \label{tab:result_pam_split}
\subfloat[Accuracy]{%
\begin{tabular}{@{}lC{0.7cm}C{0.7cm}C{0.7cm}C{0.7cm}C{0.7cm}C{0.7cm}@{}}
\toprule
\textbf{Methods}   & \textbf{0\%}   & \textbf{10\%}   & \textbf{20\%}   & \textbf{30\%}   & \textbf{40\%}   & \textbf{50\%}  \\ \midrule
Transformer       & 83.5 \std{1.5} & 60.9 \std{12.8} & 62.3 \std{11.5} & 52.0 \std{11.9} & 43.8 \std{14.0} & 43.2 \std{2.5} \\
Trans-mean        & 83.7 \std{2.3} & 62.4 \std{3.5}  & 56.8 \std{4.1}  & 65.1 \std{1.9}  & 48.7 \std{2.7}  & 46.4 \std{1.4} \\
GRU-D             & 83.3 \std{1.6} & 68.4 \std{3.7}  & 64.8 \std{0.4}  & 58.0 \std{2.0}  & 47.7 \std{1.4}  & 49.7 \std{1.2} \\
SeFT              & 67.1 \std{2.2} & 40.0 \std{1.9}  & 34.2 \std{2.8}  & 31.7 \std{1.5}  & 26.8 \std{2.6}  & 26.4 \std{1.4} \\
mTAND             & 74.6 \std{4.3} & 53.4 \std{2.0}  & 45.6 \std{1.6}  & 34.7 \std{5.5}  & 23.7 \std{1.0}  & 20.9 \std{3.1} \\
Raindrop          & 74.3 \std{3.8} & 76.7 \std{1.8}  & 71.3 \std{2.5}  & 60.3 \std{3.5}  & 57.0 \std{3.1}  & 47.2 \std{4.4} \\
Neural CDE        & 94.2 \std{0.5} & 85.8 \std{1.4}  & 74.4 \std{0.5}  & 65.5 \std{1.8}  & 55.8 \std{0.5}  & 49.9 \std{1.7} \\ \midrule
\textbf{FlowPath} & 94.8 \std{0.2} & 86.6 \std{1.1}  & 77.0 \std{1.7}  & 67.3 \std{0.2}  & 58.1 \std{0.1}  & 52.0 \std{0.6} \\ \bottomrule
\end{tabular}}\hfil
\subfloat[Precision]{%
\begin{tabular}{@{}lC{0.7cm}C{0.7cm}C{0.7cm}C{0.7cm}C{0.7cm}C{0.7cm}@{}}
\toprule
\textbf{Methods}   & \textbf{0\%}   & \textbf{10\%}   & \textbf{20\%}   & \textbf{30\%}   & \textbf{40\%}   & \textbf{50\%}  \\ \midrule
Transformer       & 84.8 \std{1.5} & 58.4 \std{18.4} & 65.9 \std{12.7} & 55.2 \std{15.3} & 44.6 \std{23.0} & 52.0 \std{2.5} \\
Trans-mean        & 84.9 \std{2.6} & 59.6 \std{7.2}  & 59.4 \std{3.4}  & 63.8 \std{1.2}  & 55.8 \std{2.6}  & 59.1 \std{3.2} \\
GRU-D             & 84.6 \std{1.2} & 74.2 \std{3.0}  & 69.8 \std{0.8}  & 63.2 \std{1.7}  & 63.4 \std{1.6}  & 52.4 \std{0.3} \\
SeFT              & 70.0 \std{2.4} & 40.8 \std{3.2}  & 34.9 \std{5.2}  & 31.0 \std{2.7}  & 24.1 \std{3.4}  & 23.0 \std{2.9} \\
mTAND             & 74.3 \std{4.0} & 54.8 \std{2.7}  & 49.2 \std{2.1}  & 43.4 \std{4.0}  & 33.9 \std{6.5}  & 35.1 \std{6.1} \\
Raindrop          & 75.6 \std{2.1} & 79.9 \std{1.7}  & 75.8 \std{2.2}  & 68.1 \std{3.1}  & 65.4 \std{2.7}  & 59.4 \std{3.9} \\
Neural CDE        & 95.2 \std{0.4} & 88.8 \std{1.2}  & 81.0 \std{0.8}  & 73.1 \std{1.8}  & 67.7 \std{1.5}  & 65.5 \std{0.8} \\ \midrule
\textbf{FlowPath} & 95.8 \std{0.4} & 89.5 \std{0.8}  & 82.4 \std{1.4}  & 75.1 \std{0.4}  & 67.0 \std{0.9}  & 66.3 \std{1.9} \\ \bottomrule
\end{tabular}}\\
\subfloat[Recall]{%
\begin{tabular}{@{}lC{0.7cm}C{0.7cm}C{0.7cm}C{0.7cm}C{0.7cm}C{0.7cm}@{}}
\toprule
\textbf{Methods}   & \textbf{0\%}   & \textbf{10\%}   & \textbf{20\%}   & \textbf{30\%}   & \textbf{40\%}   & \textbf{50\%}  \\ \midrule
Transformer       & 86.0 \std{1.2} & 59.1 \std{16.2} & 61.4 \std{13.9} & 50.1 \std{13.3} & 40.5 \std{15.9} & 36.9 \std{3.1} \\
Trans-mean        & 86.4 \std{2.1} & 63.7 \std{8.1}  & 53.2 \std{3.9}  & 67.9 \std{1.8}  & 54.2 \std{3.0}  & 43.1 \std{2.2} \\
GRU-D             & 85.2 \std{1.6} & 70.8 \std{4.2}  & 65.8 \std{0.5}  & 58.2 \std{3.1}  & 44.5 \std{0.5}  & 42.5 \std{1.7} \\
SeFT              & 68.2 \std{1.5} & 41.0 \std{0.7}  & 34.6 \std{2.1}  & 32.0 \std{1.2}  & 28.0 \std{1.2}  & 27.5 \std{0.4} \\
mTAND             & 79.5 \std{2.8} & 57.0 \std{1.9}  & 49.0 \std{1.6}  & 36.3 \std{4.7}  & 26.4 \std{1.6}  & 23.0 \std{3.2} \\
Raindrop          & 77.9 \std{2.2} & 77.9 \std{2.3}  & 72.5 \std{2.0}  & 60.3 \std{3.6}  & 56.7 \std{3.1}  & 44.8 \std{5.3} \\
Neural CDE        & 94.8 \std{0.5} & 86.3 \std{1.3}  & 74.7 \std{1.0}  & 64.9 \std{2.3}  & 54.5 \std{2.3}  & 48.9 \std{1.5} \\ \midrule
\textbf{FlowPath} & 95.5 \std{0.2} & 87.4 \std{0.4}  & 77.1 \std{2.3}  & 67.7 \std{0.7}  & 57.1 \std{0.7}  & 51.3 \std{0.7} \\ \bottomrule
\end{tabular}}\hfil
\subfloat[F1 Score]{%
\begin{tabular}{@{}lC{0.7cm}C{0.7cm}C{0.7cm}C{0.7cm}C{0.7cm}C{0.7cm}@{}}
\toprule
\textbf{Methods}   & \textbf{0\%}   & \textbf{10\%}   & \textbf{20\%}   & \textbf{30\%}   & \textbf{40\%}   & \textbf{50\%}  \\ \midrule
Transformer       & 85.0 \std{1.3} & 56.9 \std{18.9} & 61.8 \std{15.6} & 48.4 \std{18.2} & 40.2 \std{20.1} & 41.9 \std{3.2} \\
Trans-mean        & 85.1 \std{2.4} & 62.7 \std{6.4}  & 55.3 \std{3.5}  & 64.9 \std{1.7}  & 55.1 \std{2.9}  & 46.5 \std{3.1} \\
GRU-D             & 84.8 \std{1.2} & 72.0 \std{3.7}  & 67.2 \std{0.0}  & 59.3 \std{3.5}  & 47.5 \std{0.0}  & 47.5 \std{1.2} \\
SeFT              & 68.5 \std{1.8} & 39.9 \std{1.5}  & 33.3 \std{2.7}  & 28.0 \std{1.6}  & 23.3 \std{3.0}  & 23.5 \std{1.8} \\
mTAND             & 76.8 \std{3.4} & 55.9 \std{2.2}  & 49.0 \std{1.0}  & 39.5 \std{4.4}  & 29.3 \std{1.9}  & 27.7 \std{3.9} \\
Raindrop          & 76.6 \std{2.8} & 78.6 \std{1.8}  & 73.4 \std{2.1}  & 61.9 \std{3.9}  & 58.9 \std{2.5}  & 47.6 \std{5.2} \\
Neural CDE        & 95.0 \std{0.4} & 87.3 \std{1.3}  & 76.8 \std{0.6}  & 67.1 \std{1.9}  & 57.7 \std{1.6}  & 51.6 \std{1.0} \\ \midrule
\textbf{FlowPath} & 95.6 \std{0.2} & 88.3 \std{0.4}  & 79.1 \std{1.9}  & 70.0 \std{0.7}  & 59.6 \std{1.0}  & 54.1 \std{1.0} \\ \bottomrule
\end{tabular}}
% \vspace{-1.0em}
\caption{Classification performance on PAMAP2 dataset with original and sensor dropout from 10\% to 50\%}
\label{tab:result_pam_split}
\end{table}
%%%

\subsection{Results of hyperparameter tuning.}
Following \citet{zhang_graph-guided_2022}, training was performed with a batch size of 128, a learning rate of 0.0001, and 20 epochs.
Similar to the previous experiment, we employed grid search for hyperparameter tuning: number of layers $n_l \in \{1, 2, 3, 4\}$, and hidden vector dimensions $n_h \in \{16, 32, 64, 128\}$.
The results of this tuning are presented in Table~\ref{tab:hyperparameter_pam}, where we discovered that a configuration with \(n_l=4\) layers and \(n_h=128\) hidden vector dimensions yields the optimal performance of F1 Score.

%%%
\begin{table}[htbp]
\scriptsize\centering\captionsetup{skip=5pt}
\vspace{-1.0em}
% \caption{Performance comparison on `PAMAP2' with different flow configuration % on different scenarios
% }\label{tab:all_results_pam}
\subfloat[Accuracy]{%
\begin{tabular}{@{}lccccc@{}}
\toprule
\multirow{2.5}{*}{\textbf{Settings}} & \multicolumn{2}{c}{\textbf{Neural CDE}} & \multicolumn{3}{c}{\textbf{FlowPath}}                \\ \cmidrule(lr){2-3} \cmidrule(lr){4-6} 
                                & \textbf{Spline}     & \textbf{MLP}      & \textbf{ResNet} & \textbf{GRU}   & \textbf{Coupling} \\ \midrule
\textbf{Regular}                & 94.2 \std{0.5}      & 93.3 \std{0.3}    & 94.1 \std{0.4}  & 94.8 \std{0.2} & 94.4 \std{0.1}    \\
\textbf{10\% Missing}           & 85.8 \std{1.4}      & 84.8 \std{0.4}    & 86.0 \std{0.3}  & 86.6 \std{1.1} & 86.0 \std{0.4}    \\
\textbf{20\% Missing}           & 74.4 \std{0.5}      & 73.2 \std{0.4}    & 76.7 \std{1.2}  & 77.0 \std{1.7} & 75.0 \std{0.4}    \\
\textbf{30\% Missing}           & 65.5 \std{1.8}      & 65.5 \std{0.2}    & 69.6 \std{0.5}  & 67.3 \std{0.2} & 67.4 \std{0.3}    \\
\textbf{40\% Missing}           & 55.8 \std{0.5}      & 57.5 \std{1.8}    & 63.9 \std{1.7}  & 58.1 \std{0.1} & 59.0 \std{0.3}    \\
\textbf{50\% Missing}           & 49.9 \std{1.7}      & 51.5 \std{2.7}    & 55.1 \std{0.3}  & 52.0 \std{0.6} & 53.1 \std{2.0}    \\ \bottomrule
\end{tabular}} \hfil
\subfloat[Precision]{%
\begin{tabular}{@{}lccccc@{}}
\toprule
\multirow{2.5}{*}{\textbf{Settings}} & \multicolumn{2}{c}{\textbf{Neural CDE}} & \multicolumn{3}{c}{\textbf{FlowPath}}                \\ \cmidrule(lr){2-3} \cmidrule(lr){4-6} 
                                & \textbf{Spline}     & \textbf{MLP}      & \textbf{ResNet} & \textbf{GRU}   & \textbf{Coupling} \\ \midrule
\textbf{Regular}                & 95.2 \std{0.4}      & 94.2 \std{0.2}    & 94.8 \std{0.4}  & 95.8 \std{0.4} & 95.1 \std{0.3}    \\
\textbf{10\% Missing}           & 88.8 \std{1.2}      & 87.6 \std{1.5}    & 88.5 \std{0.5}  & 89.5 \std{0.8} & 88.1 \std{0.6}    \\
\textbf{20\% Missing}           & 81.0 \std{0.8}      & 79.0 \std{1.8}    & 82.3 \std{1.4}  & 82.4 \std{1.4} & 80.2 \std{1.0}    \\
\textbf{30\% Missing}           & 73.1 \std{1.8}      & 74.1 \std{3.8}    & 76.8 \std{1.5}  & 75.1 \std{0.4} & 74.4 \std{1.7}    \\
\textbf{40\% Missing}           & 67.7 \std{1.5}      & 65.9 \std{3.7}    & 70.2 \std{2.1}  & 67.0 \std{0.9} & 68.0 \std{1.4}    \\
\textbf{50\% Missing}           & 65.5 \std{0.8}      & 60.6 \std{2.5}    & 67.4 \std{2.4}  & 66.3 \std{1.9} & 65.4 \std{2.7}    \\ \bottomrule
\end{tabular}} \\ 
\subfloat[Recall]{%
\begin{tabular}{@{}lccccc@{}}
\toprule
\multirow{2.5}{*}{\textbf{Settings}} & \multicolumn{2}{c}{\textbf{Neural CDE}} & \multicolumn{3}{c}{\textbf{FlowPath}}                \\ \cmidrule(lr){2-3} \cmidrule(lr){4-6} 
                                & \textbf{Spline}     & \textbf{MLP}      & \textbf{ResNet} & \textbf{GRU}   & \textbf{Coupling} \\ \midrule
\textbf{Regular}                & 94.8 \std{0.5}      & 93.6 \std{0.6}    & 94.4 \std{0.7}  & 95.5 \std{0.2} & 94.6 \std{0.3}    \\
\textbf{10\% Missing}           & 86.3 \std{1.3}      & 85.5 \std{0.8}    & 86.4 \std{0.4}  & 87.4 \std{0.4} & 87.0 \std{0.2}    \\
\textbf{20\% Missing}           & 74.7 \std{1.0}      & 73.3 \std{2.1}    & 77.7 \std{1.1}  & 77.1 \std{2.3} & 76.4 \std{0.5}    \\
\textbf{30\% Missing}           & 64.9 \std{2.3}      & 65.7 \std{1.8}    & 68.8 \std{1.4}  & 67.7 \std{0.7} & 67.9 \std{1.4}    \\
\textbf{40\% Missing}           & 54.5 \std{2.3}      & 56.9 \std{2.5}    & 64.2 \std{1.7}  & 57.1 \std{0.7} & 58.4 \std{1.5}    \\
\textbf{50\% Missing}           & 48.9 \std{1.5}      & 48.9 \std{2.4}    & 55.1 \std{0.3}  & 51.3 \std{0.7} & 51.4 \std{2.7}    \\ \bottomrule
\end{tabular}} \hfil
\subfloat[F1 Score]{%
\begin{tabular}{@{}lccccc@{}}
\toprule
\multirow{2.5}{*}{\textbf{Settings}} & \multicolumn{2}{c}{\textbf{Neural CDE}} & \multicolumn{3}{c}{\textbf{FlowPath}}                \\ \cmidrule(lr){2-3} \cmidrule(lr){4-6} 
                                & \textbf{Spline}     & \textbf{MLP}      & \textbf{ResNet} & \textbf{GRU}   & \textbf{Coupling} \\ \midrule
\textbf{Regular}                & 95.0 \std{0.4}      & 93.9 \std{0.4}    & 94.6 \std{0.5}  & 95.6 \std{0.2} & 94.8 \std{0.2}    \\
\textbf{10\% Missing}           & 87.3 \std{1.3}      & 86.4 \std{1.0}    & 87.2 \std{0.3}  & 88.3 \std{0.4} & 87.4 \std{0.3}    \\
\textbf{20\% Missing}           & 76.8 \std{0.6}      & 75.1 \std{0.8}    & 79.3 \std{1.0}  & 79.1 \std{1.9} & 77.8 \std{0.6}    \\
\textbf{30\% Missing}           & 67.1 \std{1.9}      & 68.2 \std{0.8}    & 71.4 \std{0.7}  & 70.0 \std{0.7} & 70.2 \std{1.0}    \\
\textbf{40\% Missing}           & 57.7 \std{1.6}      & 58.8 \std{2.8}    & 65.9 \std{1.9}  & 59.6 \std{1.0} & 60.8 \std{1.1}    \\
\textbf{50\% Missing}           & 51.6 \std{1.0}      & 50.7 \std{2.0}    & 57.9 \std{0.2}  & 54.1 \std{1.0} & 54.2 \std{3.1}    \\ \bottomrule
\end{tabular}}
% \vspace{-1.0em}
\caption{Performance comparison on `PAMAP2' with different flow configuration % on different scenarios
}\label{tab:all_results_pam}
\end{table}
%%%

Table~\ref{tab:all_results_pam} presents a performance comparison between the standard Neural CDE, incorporating an MLP-based path, and the proposed FlowPath model in scenarios with missing sensor data. Baseline methods that rely on fixed or unconstrained paths show greater performance degradation, underscoring the advantages of the proposed approach.

\end{document}